%% file: main.tex
\documentclass[11pt]{article}

\usepackage[utf8]{inputenc}
\usepackage[T1]{fontenc}
\usepackage{amsmath,amssymb,amsfonts}
\usepackage{graphicx}
\usepackage{booktabs}
\usepackage{hyperref}
\usepackage{xcolor}
\usepackage{multirow}
\usepackage{caption}
\usepackage[margin=1in]{geometry}
\usepackage{natbib}
\usepackage{enumitem}
\usepackage{subcaption}
\usepackage{array}
\usepackage{longtable}
\usepackage{tabularx}
\usepackage{longtable}
\usepackage{lscape}
\usepackage{pdflscape}
\usepackage{float}

\hypersetup{colorlinks=true, linkcolor=blue, citecolor=blue, urlcolor=blue}

\title{\textbf{Incompressible Knowledge Probes: Estimating Black-Box LLM Parameter Counts via Factual Capacity}}

\author{
  Bojie Li \\[6pt]
  Pine AI
}

\date{}

\begin{document}
\maketitle

\begin{abstract}
Closed-source frontier labs do not disclose parameter counts, and the standard alternative---inference economics---carries $2\times$+ uncertainty from hardware, batching, and serving-stack assumptions external to the model. We ask a different question: how much does a model \emph{know}? Storing $F$ facts requires at least $F/$(bits per parameter) weights, so factual recall lower-bounds parameter count---an \emph{intrinsic}, serving-independent signal, though (as we show) a coarse one. We introduce \textbf{Incompressible Knowledge Probes (IKPs)}, a benchmark of 1{,}400 factual questions spanning 7 tiers of obscurity, designed to isolate knowledge that cannot be derived by reasoning or compressed by architectural improvements.

We score with no hallucination penalty ($\lambda = 0$: IKP accuracy is simply the fraction of probed facts answered correctly), which removes both the penalty hyperparameter and the per-tier flooring choice; a full $\lambda \times$ flooring ablation (Appendix~\ref{app:lambda}) shows the calibration is robust across scoring choices while individual estimates are not, motivating the no-penalty default. We calibrate a log-linear mapping from IKP accuracy to parameter count on 93 open-weight models (135M--1{,}600B) spanning 19 vendors, achieving $R^2 = 0.910$; leave-one-out cross-validation confirms generalization (median fold error $1.48\times$, $72\%$ within $2\times$, $86\%$ within $3\times$). The instrument is deliberately coarse---its 90\% prediction interval spans ${\sim}3\times$ in either direction, wider than inference economics---so IKP recovers order-of-magnitude effective capacity and relative rankings, not precise parameter counts. For Mixture-of-Experts models, total parameters predict knowledge ($R^2 = 0.67$) better than active parameters ($R^2 = 0.41$). We evaluate 201 models from 27 vendors on a curated probe set (1{,}311 of 1{,}400 probes surviving name-collision and label-ambiguity filters) and report effective knowledge capacity for all major proprietary frontier models as prediction bands rather than point estimates; for heavily safety-tuned models these are lower bounds, since refusal policy can suppress tens of percentage points of otherwise-answerable capacity. A black-box hallucination-similarity test---rare-fact agreement on \emph{wrong} answers---separates weight-sharing siblings, post-training lineages, and independent retrains without model weights.

The widely-reported saturation of reasoning benchmarks does not imply the end of scaling. Procedural capability compresses under the ``Densing Law,'' but across 100 dated open-weight models the IKP time coefficient is $+0.0013$/month (95\% CI $[-0.0004, +0.0033]$, $p = 0.19$)---indistinguishable from zero, and rejecting the Densing prediction of $+0.0129$/month at $p < 10^{-15}$. Factual capacity continues to scale log-linearly with parameters across generations and across vendors.
\end{abstract}

\section{Introduction}
\label{sec:intro}

For the past three years, the author and a few peers have stress-tested newly released models against a standing probe---``What do you know about USTC Hackergame?''---an annual Capture-the-Flag contest, run by the author's university Linux User Group, with idiosyncratic Chinese-language challenge titles. The same prompt, asked at each model, traced a single fact arriving in parameters over time. In May 2024, GPT-4o knew the contest existed but invented fake challenge names when asked. Just nine months later, in February 2025, Claude 3.7 Sonnet listed nineteen verified Hackergame 2023 titles, verifiable against the official writeup repository. By April 2026, Kimi K2.6, Claude Opus 4.7 and Gemini 3.1 Pro list specific challenges from multiple consecutive years (Appendix~\ref{app:hackergame-case}).

This is not a quirk of one event. Transformer feed-forward layers function as key-value memories for factual associations~\citep{geva2021transformer, meng2022locating, dai2022knowledge}, with an empirical storage capacity of ${\sim}2$--$4$ bits of factual knowledge per parameter~\citep{allenzhu2025, morris2025memorize}.

A systematic probe across fellow researchers revealed an unexpected pattern: \emph{citation count and $h$-index alone do not determine whether a frontier model recognizes a researcher.} Two researchers with similar metrics can elicit entirely different responses---those whose work has broad impact appear reliably, while those whose citations spread across many incremental papers often do not.

A second pattern appears across vendors: as of April 2026, Gemini 3.1 Pro recognizes researchers and systems that GPT-5.5 hesitates on, and GPT-5.5 in turn recognizes more than Claude Opus 4.7. This ranking shifts across release cycles, so an instrument re-run at each release is itself useful.

The generalization: frontier models have become compressed images of the expert discourse of their era. Human experts work in the open---papers, code, talks, documentation---and each training run distills more of that output into parameters, so the knowledge of entire research communities flows from the long tail of the open web into a handful of proprietary models. How much a model has absorbed is bounded below by the Shannon entropy~\citep{shannon1948mathematical} of the stored facts and above by its parameter budget, and empirically, accuracy on rare facts scales log-linearly with model size across three orders of magnitude~\citep{petroni2019language, roberts2020much, kandpal2023, lu2024factmemorization}. The fraction of an expert's contributions a model has internalized is thus \emph{empirically measurable}, via probes spanning (i)~rare-but-real researchers and their subfields and (ii)~specific attributes of specific entities, scored through the public API.

The same instrument, inverted, addresses a problem practitioners care about for independent reasons. Because closed-source labs do not disclose parameter counts, the field relies on inference economics~\citep{epochai2024, cai2025substitution}---size inferred from API throughput, pricing, and hardware cost models---which carries acknowledged $2\times$+ uncertainty from factors (hardware generation, batching, quantization, serving stack) external to the model itself. A knowledge-based estimator is \emph{intrinsic}: because storing $F$ facts requires at least $F/$(bits per parameter) weights, the furthest tier of the frequency long tail a model has internalized directly constrains its parameter count from below. Measuring how much a model \emph{knows} therefore \emph{estimates}---coarsely---how many parameters it \emph{has}: an inverse problem not previously exploited for parameter estimation, using only black-box API access. As we show, the estimate carries a ${\sim}3\times$ prediction interval; its value is order-of-magnitude sizing and relative ranking, not precise counts.

This reframing matters because of a now-widespread misreading of the ``Densing Law''~\citep{densing2025}: capability per parameter doubles every ${\sim}3.5$ months, a 2026 7B model matches a 2023 70B on MMLU~\citep{hendrycks2021measuring} or HELM~\citep{liang2023holistic}, therefore scaling has stopped mattering. We argue this conflates two distinct resources stored in the same parameter budget. \emph{Procedural} capability---reasoning, parsing, instruction following---is a compressible function over its inputs; better architectures and training recipes genuinely do pack more of it into fewer parameters, and the Densing Law is real for it. \emph{Factual} knowledge is not. ``USTC Hackergame started in 2014'' cannot be derived, computed, or inferred from general knowledge; it must be stored explicitly, and Shannon entropy provides a hard lower bound on the bits required. IKPs are designed to isolate this incompressible component, and empirically the isolation holds (Section~\ref{sec:densing-falsification}): the time coefficient on IKP accuracy is statistically indistinguishable from zero and rejects the Densing prediction at $p < 10^{-15}$. Benchmark saturation is therefore not evidence that scaling has ended, but that benchmarks have stopped measuring the part of scaling that cannot compress.

\begin{figure}[H]
    \centering
    \includegraphics[width=\textwidth]{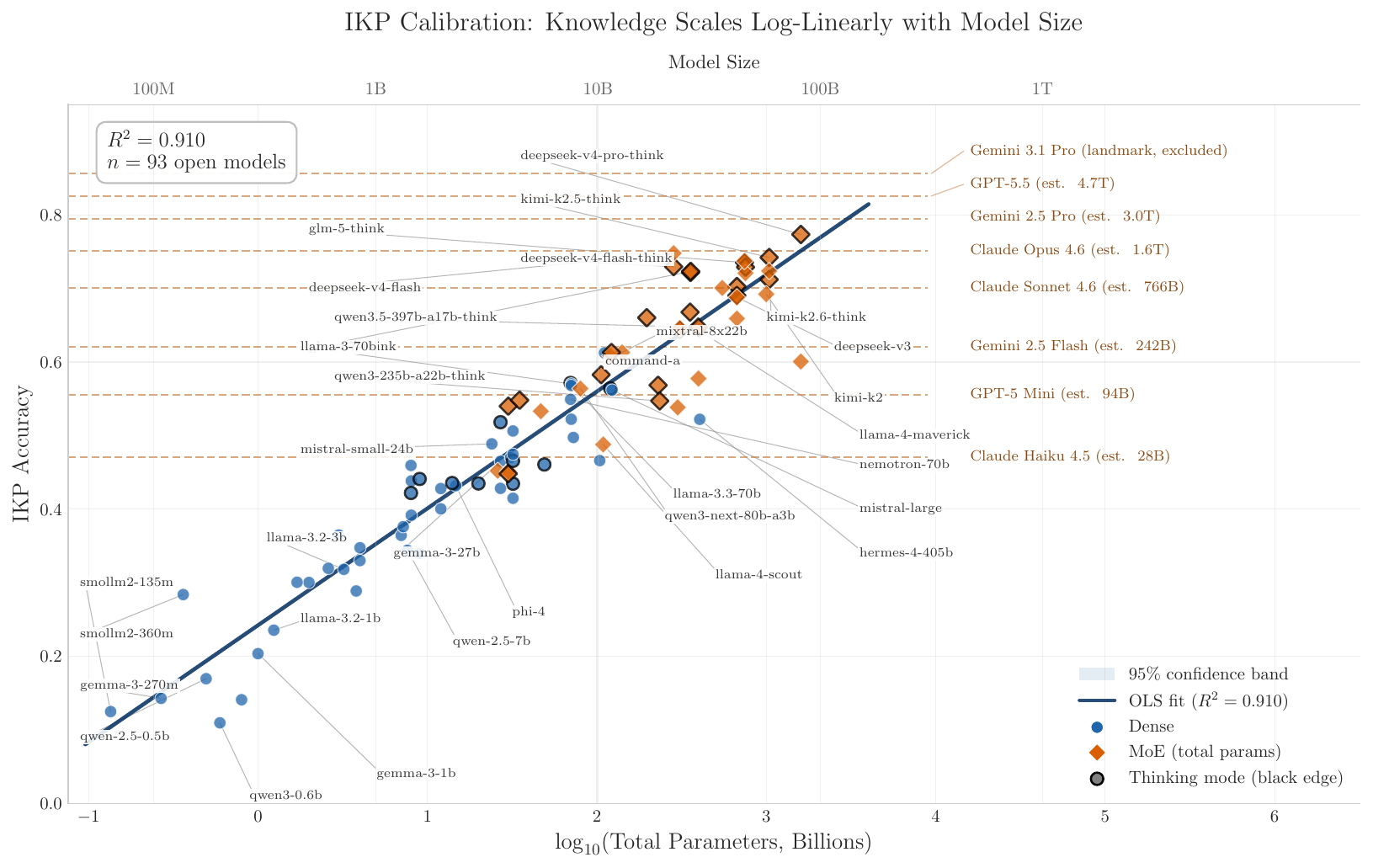}
    \caption{IKP calibration curve. Each point is an open-weight model with known parameter count. Blue circles: dense models; orange diamonds: MoE models (plotted at total parameter count). The regression line ($R^2 = 0.910$, 93 models from 19 vendors, no-penalty scoring) enables parameter estimation for proprietary models (shown as horizontal dashed lines on the right). Gemini 3.1 Pro is excluded from estimation (landmark, T6 score inflated by construction).}
    \label{fig:calibration}
\end{figure}

Figure~\ref{fig:calibration} summarises the picture: projecting proprietary frontier scores onto the open-weight calibration line places every major closed model at a coarse effective size (a ${\sim}3\times$ band), several plausibly above the largest open-weight release.

\paragraph{Contributions.}
\begin{enumerate}[leftmargin=*]
    \item \textbf{Incompressibility framework.} We distinguish \emph{compressible} procedural capability (subject to the Densing Law) from \emph{incompressible} factual capacity (bounded by Shannon entropy), and introduce Incompressible Knowledge Probes (IKPs), a tiered benchmark designed to measure only the latter from black-box API access.
    \item \textbf{Calibration and validation.} We calibrate IKP on 93 open-weight models across 19 vendors ($R^2 = 0.910$, no-penalty scoring), with leave-one-out cross-validation placing 72\% of models within $2\times$ and 86\% within $3\times$ of their known size---a coarse but consistent estimator, not a precise one.
    \item \textbf{Densing Law falsification.} Across 100 dated open-weight models, the time coefficient on IKP is $+0.0013$/month (95\% CI $[-0.0004, +0.0033]$)---indistinguishable from zero and rejecting the Densing prediction of $+0.0129$/month at $p < 10^{-15}$. Factual capacity does not compress.
    \item \textbf{Frontier estimates and the MoE total-params rule.} We give coarse effective-capacity estimates (${\sim}3\times$ prediction interval) for 97 proprietary frontier models, and show that total---not active---parameters predict MoE knowledge capacity ($R^2 = 0.67$ vs.\ $0.41$).
    \item \textbf{Knowledge fingerprinting.} Combining rare-fact Jaccard overlap with \emph{hallucination similarity} (the rate at which two models produce the same wrong answer on rare facts) yields a training-free test that distinguishes weight-sharing siblings, post-training lineages, and full retrains---including across closed-vendor version families---without requiring model weights.
    \item \textbf{Open-source release.} We release the IKP evaluation toolkit, the full probe set, and the results so that any researcher can estimate a model's effective knowledge capacity from OpenRouter's standard API endpoint. Code: \url{https://github.com/19PINE-AI/ikp}. Companion website: \url{https://01.me/research/ikp}.
\end{enumerate}

\section{Background and Related Work}
\label{sec:related}

\subsection{Knowledge Capacity of Language Models}

\citet{allenzhu2025} establish that transformer models store approximately 2 bits of knowledge per parameter, derived from synthetic factual tuples. \citet{morris2025memorize} refine this estimate to ${\sim}3.6$ bits per parameter for GPT-style models, distinguishing unintended memorization from generalization and showing that models ``memorize until their capacity fills, at which point grokking begins.'' Together, these provide a theoretical link between parameter count and factual storage capacity with a $2$--$4$ bits/param range. \citet{lu2024factmemorization} show that fact capacity scales linearly with model size and exponentially with training epochs, estimating that memorizing all Wikidata facts requires ${\sim}1{,}000$B parameters for 100 epochs. \citet{pan2025compression} formalize knowledge acquisition through the lens of Kolmogorov complexity, proposing a Syntax-Knowledge model grounded in Heap's and Zipf's laws that explains scaling laws and hallucination patterns from first principles. \citet{chang2024} show that 7B models have significantly greater factual knowledge acquisition effectivity than 1B models, demonstrating that model size qualitatively changes knowledge acquisition. \citet{petroni2019language} first demonstrated that pretrained language models serve as implicit knowledge bases, and \citet{roberts2020much} showed that model size directly determines how much knowledge can be packed into parameters. At the mechanistic level, transformer feed-forward layers function as key-value memories~\citep{geva2021transformer}, with specific ``knowledge neurons'' responsible for individual facts~\citep{dai2022knowledge} that can be causally localized and edited~\citep{meng2022locating}.

\subsection{Long-Tail Knowledge and Scaling}

Neural scaling laws~\citep{kaplan2020scaling, hoffmann2022training} establish power-law relationships between model size, compute, data, and loss. \citet{kandpal2023} demonstrate that accuracy on rare facts scales log-linearly with model size ($R^2 = 0.98$ within the BLOOM family). Larger models extend further down the power-law frequency tail of web knowledge~\citep{tirumala2022}, consistent with the Zipfian distribution of entity popularity~\citep{zipf1949human, piantadosi2014zipf}. \citet{badhe2026longtail} provide a structured taxonomy of long-tail knowledge loss mechanisms---gradient dilution, representational interference, tokenization effects, and post-training compression---formalizing the challenges that IKP tiers exploit for discrimination. \citet{mallen2023trust} confirm that LLMs are unreliable on less popular facts, with parametric memory failing where retrieval succeeds. The ``Law of Knowledge Overshadowing''~\citep{overshadow2025} shows that popular knowledge suppresses less popular knowledge, with hallucination rate increasing log-linearly with knowledge popularity---implying that obscure probes are more discriminating. \citet{carlini2021extracting, carlini2023quantifying} demonstrate that memorization scales with model size, and \citet{chen2025factoids} show that factoid retention degrades across fine-tuning stages, explaining why post-training can shift IKP scores between generations.

\subsection{Existing Model Size Estimation}

Epoch AI's inference economics approach~\citep{epochai2024} estimates parameter counts from token throughput and API pricing, with acknowledged $2\times$+ uncertainty. \citet{gao2025modelequality} formalize API model verification as a two-sample testing problem, finding that 11 of 31 Llama API endpoints serve distributions different from Meta's reference weights with just ${\sim}10$ samples per prompt. LLMmap~\citep{llmmap2025} fingerprints identify \emph{which} model is behind an API ($>95\%$ accuracy across 42 versions) but do not estimate size of unknown models. More recent fingerprinting work includes RoFL~\citep{tsai2025rofl}, which identifies models from statistical output patterns without model modification, and \citet{nasery2025fingerprinting}, who embed $24{,}576$ fingerprints into a single model that persist after fine-tuning. Both are complementary to IKP: they identify \emph{which} model is served but do not estimate size of unknown models. \citet{cai2025substitution} show that text-output statistical tests fail at ${\sim}50\%$ for detecting quantization, motivating knowledge-based approaches. The Densing Law~\citep{densing2025} demonstrates why standard benchmarks~\citep{hendrycks2021measuring, joshi2017triviaqa, kwiatkowski2019natural, liang2023holistic} fail as size proxies: capability density doubles every ${\sim}3.5$ months, making a 7B model from 2026 match a 70B model from 2023 on reasoning tasks.

\subsection{Parameter Specialization}

\citet{hong2025} show that stronger models develop more specialized parameter vectors for factual knowledge storage, with the ratio of parameters dedicated to factual knowledge trending predictably with model quality. For MoE models, joint scaling laws~\citep{ludziejewski2025moe} show MoE can be more memory-efficient than dense, while \citet{zhao2025moescaling} identify five factors (data size, total and active model size, active expert count, shared expert ratio) governing MoE performance---informing our effective-parameter analysis (Section~\ref{sec:results}).

\subsection{Distillation Detection}

Knowledge distillation detection is an emerging problem. \citet{shi2025kddistillation} formalize the task for open-weight models, using data-free input synthesis and statistical scoring when only the student's weights and the teacher's API are available. \citet{li2025experts} detect distillation via MoE expert routing patterns that persist through distillation, achieving $>94\%$ detection accuracy with a Shadow-MoE approach for black-box settings. IKP offers a complementary signal: knowledge fingerprints (Section~\ref{sec:results}) measure rare-fact overlap and shared-error rates between model pairs through black-box API access alone, without requiring model weights or MoE structure.

\section{Theoretical Framework}
\label{sec:theory}

\subsection{Factual Knowledge as Incompressible Information}

We define \emph{incompressible factual knowledge} as a factual association $(e, a, v)$---entity $e$, attribute $a$, value $v$---where $v$ cannot be derived, computed, or inferred from other known facts or from structural regularities in language.

We acknowledge that factual knowledge is not \emph{perfectly} incompressible: prior distributions over values provide some compression. For example, a founding year can be narrowed to a plausible era (say, 1500--2025), reducing the search space from $2^{32}$ possibilities to ${\sim}500$, or ${\sim}9$ bits. But the exact year within that range still requires ${\sim}9$ bits per fact, which cannot be compressed further. Similarly, while statistical regularities exist across entities (e.g., European capitals tend to be large cities), IKP probes are specifically designed to minimize such compressibility: researcher probes exclude ML/AI researchers (whose field is trivially predictable from venue co-occurrence), founding-year probes target the specific year (not the century), and name-collision-prone entities are filtered out (Section~\ref{sec:method}). The residual compressibility of each probe is small relative to the total information content of the 1{,}400-probe set, preserving the log-linear relationship between stored facts and parameter count.

\subsection{The Capacity Bound}

From \citet{allenzhu2025}, a model with $N$ parameters stores at most ${\sim}2N$ bits of factual knowledge under ideal training (with empirical estimates ranging up to ${\sim}3.6$ bits/param~\citep{morris2025memorize}). From \citet{kandpal2023}, observed accuracy scales as $\text{acc} \approx a \cdot \log(N) + b$, confirmed by \citet{lu2024factmemorization} who show linear scaling of fact capacity with model size. Combining these: for a probe set $P$ with known information content, observed accuracy constrains $N$ from below.

\subsection{Why Procedural Improvements Do Not Help}

We decompose model parameters into three functional roles: $N = N_{\text{fact}} + N_{\text{proc}} + N_{\text{ling}}$, where $N_{\text{fact}}$ is the share devoted to storing specific factual associations (entity attributes, dates, names), $N_{\text{proc}}$ is the share devoted to procedural skills (reasoning, parsing, instruction following, tool use), and $N_{\text{ling}}$ is the share devoted to linguistic competence (syntax, morphology, vocabulary, register). The Densing Law improves the efficiency of $N_{\text{proc}}$ and $N_{\text{ling}}$, freeing capacity for $N_{\text{fact}}$ or allowing smaller total $N$. But $N_{\text{fact}}$ is bounded below by the information content of stored facts. IKPs are designed to probe primarily $N_{\text{fact}}$ (answering still requires basic parsing, i.e.\ some $N_{\text{ling}}$/$N_{\text{proc}}$), and to the extent they do, aggregate accuracy provides a lower bound on $N$.

\subsection{The Frequency-Capacity Relationship}

Web knowledge follows a power-law frequency distribution (Zipf's law~\citep{zipf1949human}). Models with more capacity memorize facts further down the long tail. This creates a natural frequency cutoff per model: facts above the cutoff are memorized, those below are not. Since the number of facts above a power-law cutoff grows logarithmically with the cutoff position, aggregate accuracy scales as $\log(N)$---yielding the log-linear relationship we observe empirically.

Per-tier accuracy follows a logistic sigmoid: $T_i(N) = L_i / (1 + \exp(-k_i \cdot (\log N - m_i)))$, where $m_i$ is the midpoint parameter count for tier $i$. The aggregate (mean of 7 shifted sigmoids) approximates a straight line over a wide range of $\log N$.

\section{Methodology}
\label{sec:method}

\subsection{Probe Generation}
\label{sec:probe-generation}

IKP probes are generated through a two-phase pipeline that targets seven \emph{difficulty tiers} (T1--T7) defined by entity-popularity proxies (Common Crawl document frequency) and validated empirically against the landmark ladder (Section~\ref{sec:tiers}). The full pipeline produces 1{,}400 probes (200 per tier); the design and per-tier composition are summarized here, with full reproducibility details in Appendix~\ref{app:probes}.

\textbf{Phase A: LLM-generated candidates (T1--T2 primary, T3--T4 supplementary).} A capable LLM (GPT-5) generates candidate questions with answers, prompted with tier-specific frequency targets and rotating region/domain emphasis to enforce coverage across people, places, organizations, events, publications, and measurements. Of the 1{,}400 final probes, 401 originate from this phase, dominantly at T1 ($n=166$) and T2 ($n=152$), with smaller supplementary contributions at T3 ($n=51$) and T4 ($n=32$). Empirically, ${\sim}82\%$ of LLM-generated candidates land in T1--T2 regardless of difficulty prompting, indicating that an LLM cannot reliably generate factual probes beyond its own knowledge horizon.

\textbf{Phase B: Corpus-grounded probes (T3--T7).} For harder tiers, LLM recall is circular by construction (the generator can only produce facts it knows). We instead sample entities from external corpora and verify answers against the source. Two corpora supply the bulk of T3--T7:

\begin{itemize}[leftmargin=*,topsep=2pt,itemsep=2pt,parsep=0pt]
    \item \textbf{Wikidata}~\citep{vrandecic2014wikidata} via SPARQL: 557 probes targeting founding years and capital-style attributes for institutions (universities, journals, museums), civic landmarks (bridges, sports clubs), and geographic places. Wikidata supplies ground truth via published property values; we sample entities by Wikipedia view-count quartiles to populate T3 ($n=94$), T4 ($n=111$), T5 ($n=100$), T6 ($n=141$), and T7 ($n=100$).
    \item \textbf{DBLP / arXiv researcher records}: 345 probes asking the model to identify a CS researcher's primary subfield \emph{and} name one verifiable artifact---a paper title, named system, institutional affiliation, or co-author---associated with that researcher's work. Researchers are sampled from DBLP and OpenAlex by citation-count buckets and assigned a primary subfield via venue tags (e.g., SOSP/OSDI $\to$ ``operating systems''). The evidence requirement separates models that genuinely know the researcher from those that emit a plausible but unattested CS subfield label; the judge accepts a response only if the subfield matches \emph{and} the response cites a verifiable artifact from a per-researcher OpenAlex-derived evidence bundle (Appendix~\ref{app:probes}). This source dominates T5--T7 ($n=100, 59, 100$) and supplies the steepest discrimination at the frontier: even a 1.6T model cannot reliably name the subfield and a real artifact for a researcher with 10--50 citations.
\end{itemize}

The remaining 97 probes are manual or supplementary additions used to balance T1--T4 coverage.

\textbf{Probe quality filters.} We exclude: (1)~\emph{computable} probes whose answers can be derived by rule (e.g., IUPAC element naming, arithmetic on years, alphabetical ordering), since these test reasoning rather than memorization; (2)~\emph{ambiguous} probes that larger models legitimately disagree on, identified by the monotonicity filter (Appendix~\ref{app:probes}): if a more-capable landmark gets a probe wrong while a less-capable one gets it right, the probe is dropped; (3)~\emph{collision-prone} researcher probes with common names (two-character Chinese names, single-initial given names) where multiple distinct researchers share an identifier; (4)~\emph{contamination-prone} probes drawn from machine-learning or AI subfields, where researchers' own work tends to dominate frontier-model training corpora and inflates accuracy beyond what generic factual recall would predict; (5)~\emph{grounding embedded in the question template} for Wikidata-sourced probes, following the researcher-probe pattern: every entity-naming question carries a discriminating attribute (year, country, admin region, genre, or publisher) extracted from a Wikidata field other than the answer field. This prevents title-collision failures (``Madonna and Child''---which one?, ``Putnam''---in which state?) without leaking the answer. We arrived at this template after a 10-round audit (Section~\ref{sec:wikidata-audit}) revealed that ${\sim}12\%$ of T5--T7 Wikidata probes carried a bare-name ambiguity defect and ${\sim}33$--$46\%$ of diverse-fact-type Wikidata candidates failed independent web cross-check at the obscure tier.

\textbf{Automated collision/ambiguity audit and robustness.} Beyond the template filters, we run an automated post-hoc audit of the released set. For every researcher probe we query OpenAlex for same-name authors and flag a probe when a distinct researcher (different institution, ${\geq}100$ citations) shares the exact name, so the model cannot know which person is meant; for every Wikidata probe we fetch the entity's label and flag it when ${\geq}2$ items of the same type share that label and the question carries no disambiguating qualifier. This flags $89/1400$ probes ($6.4\%$: $59$ researcher name-collisions, $29$ Wikidata label-ambiguities, one known-wrong manual answer), concentrated in the hard tiers (T3--T7) as expected. Re-scoring every model on the cleaned $1{,}311$-probe subset \emph{improves} the calibration---$R^2$ rises from $0.900$ to $0.910$ and the 90\% prediction-interval factor tightens from $3.45\times$ to $3.20\times$---and individual frontier estimates move by ${<}5\%$. The ambiguous probes added noise, not signal; estimates are robust to their removal, and all results use the cleaned set.

The complete generation pipeline, per-tier composition table, sample probes, verification procedure for each source type, and the audit log are detailed in Appendix~\ref{app:probes}.

\subsection{Landmark-Based Tier Assignment}
\label{sec:tiers}

Tiers are assigned empirically, not by proxy metrics. Six \textbf{landmark models} spanning from 0.5B to frontier define the tier boundaries:

\begin{center}
\begin{tabular}{lllr}
\toprule
Tier landmark & Model & Params & Tier boundary \\
\midrule
T1 & Qwen 2.5 0.5B & 0.5B & T1/T2 \\
T2 & Qwen 2.5 7B & 7.6B & T2/T3 \\
T3 & Qwen 3 32B & 32B & T3/T4 \\
T4 & Qwen 3 235B & 235B & T4/T5 \\
T5 & Kimi K2.5 & 1T & T5/T6 \\
T6 & Gemini 3.1 Pro & Frontier & T6/T7 \\
\bottomrule
\end{tabular}
\end{center}

A probe is assigned to tier $k$ if the T$_k$ landmark answers correctly but the T$_{k-1}$ landmark does not. Probes with non-monotonic correctness across the ladder are dropped (${\sim}15\%$), as non-monotonicity indicates ambiguous answers that larger models legitimately contest.

\subsection{Scoring}

Each model receives the question with a system prompt instructing direct, concise answers. All queries use temperature${}=0$.

\textbf{Judge.} Gemini 3 Flash Preview classifies each response. Non-researcher probes (Wikidata, LLM-generated, manual) use a 3-way judge: CORRECT, REFUSAL (``I don't know''), or WRONG (confident incorrect answer). Researcher subfield probes use a 4-way \emph{evidence-aware} judge that consumes the per-researcher evidence bundle (Section~\ref{sec:setup}, Appendix~\ref{app:probes}): CORRECT\_STRONG (right subfield \emph{and} cites a verifiable evidence item---paper title fragment, named system, venue, affiliation, or co-author), CORRECT\_WEAK (right subfield only, no specific evidence; or correct evidence with adjacent-but-not-listed subfield), WRONG (subfield outside \{primary, secondary\} or fabricated specifics), or REFUSAL.

\textbf{Per-probe score.} Each probe scores in $\{+1.0,\; +0.5,\; 0,\; \lambda\}$:
\begin{center}
\small
\begin{tabular}{lcc}
\toprule
Verdict & Non-researcher & Researcher \\
\midrule
CORRECT (= STRONG)   & $+1.0$  & $+1.0$ \\
CORRECT\_WEAK        & ---     & $+0.5$ \\
REFUSAL              & $0$     & $0$ \\
WRONG                & $\lambda$ & $\lambda$ \\
\bottomrule
\end{tabular}
\end{center}
We use $\lambda = 0$ (no penalty) throughout: IKP accuracy is then simply the fraction of probed facts a model answers correctly, with \emph{no} scoring hyperparameter to tune and \emph{no} per-tier flooring decision (per-tier scores are $\text{correct}/\text{total} \geq 0$ by construction, so flooring is a no-op). This is a \emph{parsimony principle}, not a convenience: we take the parameter-free operating point unless a nonzero penalty materially improves the calibration---which it does not (Appendix~\ref{app:lambda}: across $\lambda \in [0,-2]$ the fit varies by ${\leq}0.03$ in $R^2$ while estimates swing $2$--$3\times$ by vendor, and the flooring choice under a negative penalty can itself inflate frontier estimates several-fold). Readers who prefer a hallucination-aware score can read the calibration and per-model estimates under any $\lambda$ from Table~\ref{tab:lambda-sweep}. At $\lambda = 0$ the researcher judge's four verdicts collapse to STRONG ($1.0$) $>$ WEAK ($0.5$) $>$ REFUSAL $=$ WRONG ($0$): the evidence-aware STRONG/WEAK distinction is retained, and a wrong subfield guess scores the same as an honest ``I don't know.''

\textbf{Aggregate accuracy.} Per-tier score is the mean per-probe score in that tier; overall accuracy is the unweighted mean of the seven per-tier scores, so hard-tier behavior cannot erase easy-tier knowledge. Per-tier scores are not floored at zero (under a negative penalty a strong bluffer can score negative), preserving the bluff signal; at the default $\lambda=0$ this is moot, as scores are $\geq 0$.

\subsection{Calibration Curve}

We fit $A = \alpha \cdot \log_{10}(N) + \beta$ where $A$ is aggregate accuracy (no-penalty, $\lambda=0$) and $N$ is parameter count in billions, using ordinary least squares on open-weight models with known parameter counts. To estimate the parameter count of a target model, we invert the regression: $\hat{N} = 10^{(A - \beta)/\alpha}$. We report the forward-direction $R^2$ (predicting accuracy from size), which equals the squared Pearson correlation and is invariant to inversion.

\section{Experimental Setup}
\label{sec:setup}

\textbf{Probe set.} 1{,}400 probes (200 per tier, T1--T7) across 16 domains: researcher subfields (345 probes), founding years, geography, history, science, culture, and others.

\textbf{Models.} 201 models from 27 vendors, accessed via OpenRouter API and local Ollama server. Open-weight models with known parameter counts (93 models after excluding pathological-refuser and post-training-degraded outliers, 135M to 1{,}600B including DeepSeek V4 Pro) serve as calibration data, including models from the Llama~\citep{touvron2023llama, touvron2023llama2}, Qwen~\citep{yang2024qwen2}, Gemma~\citep{gemma2024}, DeepSeek~\citep{deepseek2024v2, deepseek2025r1}, and OLMo~\citep{groeneveld2024olmo} families. Proprietary models (97 models), including GPT-4~\citep{openai2023gpt4}, are estimation targets.

\textbf{Judge.} Gemini 3 Flash Preview with temperature${}=0$ and low reasoning effort. Estimated judge error rate: $0.1$--$0.2\%$ (verified by manual audit).

\textbf{Researcher probes.} All 345 researcher probes use a two-part format: ``In computer science, what is the research subfield of [Name], and name one paper, system, institution, or co-author associated with their work? If you don't know who this person is, say so.'' The CS scoping reduces cross-field name collisions; the artifact requirement forces models to produce verifiable evidence rather than emit a plausible-sounding but unattested subfield label. Each researcher carries a manually- or OpenAlex-derived evidence bundle (subfield, top venues, named systems, affiliations, top co-authors), and the four-way judge classifies a response as CORRECT-STRONG only when the response names the right subfield \emph{and} cites at least one matching evidence item; subfield-only responses without specific evidence get CORRECT-WEAK; subfield mismatch with fabricated specifics gets WRONG; ``I don't know'' is REFUSAL. See Appendix~\ref{app:probes} for the full evidence schema, sample probes, and judge rubric.

\section{Results}
\label{sec:results}

\subsection{Calibration Curve Quality}

\begin{table}[h]
\centering
\caption{Scaling law fits: $\text{acc} = \text{slope} \cdot \log_{10}(N_B) + \text{intercept}$}
\label{tab:scaling}
\begin{tabular}{lrrrr}
\toprule
Subset & $n$ & Slope & $R^2$ (tier-mean) & $R^2$ (overall) \\
\midrule
All open models & 93 & 0.159 & \textbf{0.910} & 0.910 \\
Dense only & 52 & 0.151 & 0.875 & 0.855 \\
Dense non-thinking & 42 & 0.150 & 0.877 & 0.866 \\
MoE (total params) & 41 & 0.145 & 0.667 & 0.712 \\
MoE (active params) & 41 & 0.144 & 0.412 & 0.483 \\
\bottomrule
\end{tabular}
\end{table}

\begin{figure}[H]
    \centering
    \includegraphics[width=\columnwidth]{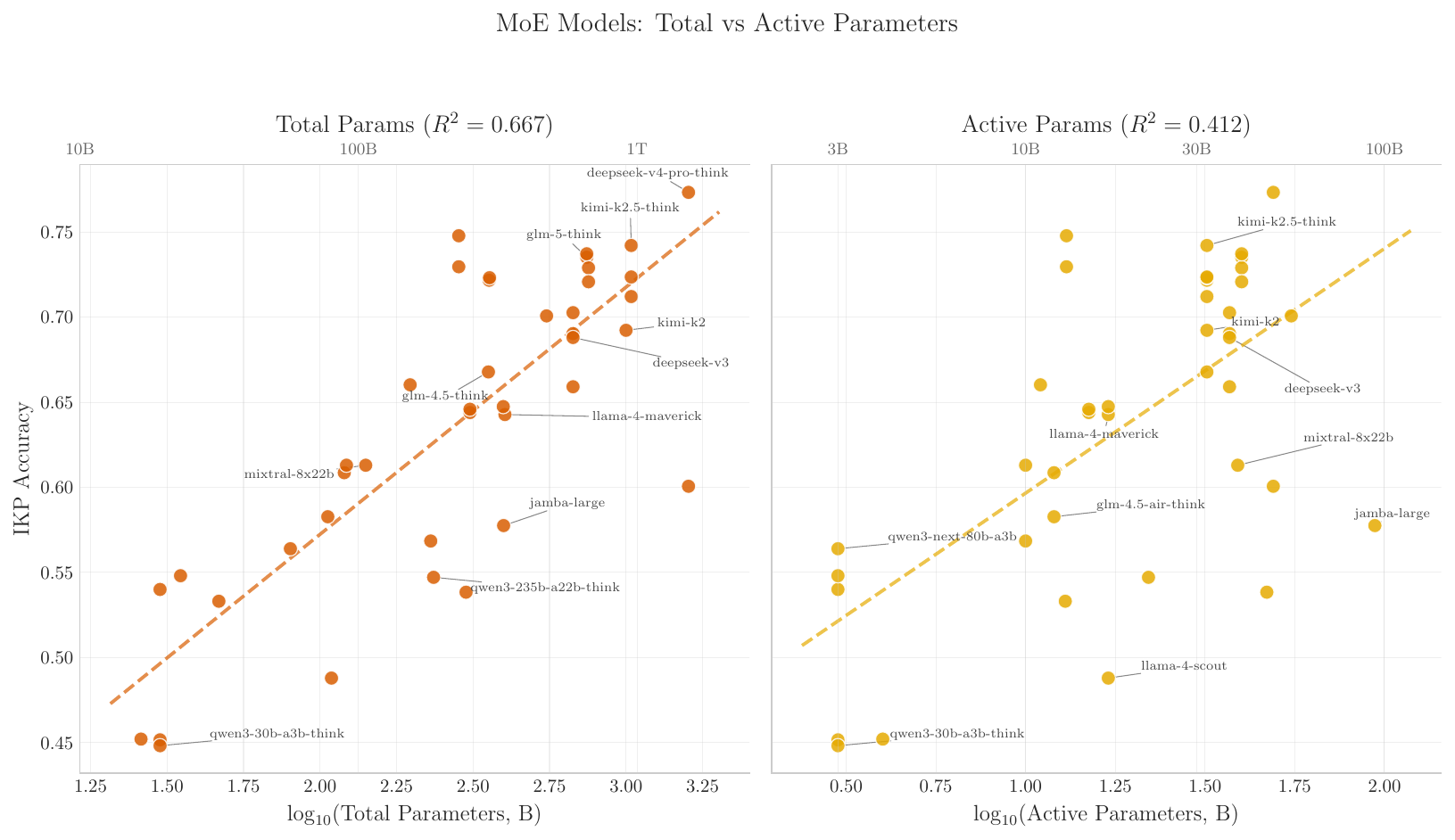}
    \caption{MoE models: accuracy vs total parameters (left, $R^2 = 0.67$) and vs active parameters (right, $R^2 = 0.41$). Total parameters are clearly the better predictor of factual knowledge capacity.}
    \label{fig:moe}
\end{figure}

The all-open fit achieves $R^2 = 0.910$ under no-penalty scoring ($\lambda = 0$), with each $10\times$ increase in parameters adding ${\sim}15.9$ percentage points (Figure~\ref{fig:calibration}). The tier-mean and overall-accuracy fits agree ($R^2 = 0.910$ both). For Mixture-of-Experts models~\citep{shazeer2017outrageously, fedus2022switch}, total parameters ($R^2 = 0.67$) outperform active parameters ($R^2 = 0.41$), consistent with factual knowledge being stored across all expert weights rather than only those activated per token (Figure~\ref{fig:moe}). Dense and MoE share nearly the same slope; we calibrate on the combined set because fitting MoE anchors alone does not improve recovery of known MoE sizes in leave-one-out (Appendix~\ref{app:moe}), though it would raise the highest frontier estimates by ${\sim}18\%$. This matches MoE scaling laws in which total capacity governs knowledge storage~\citep{clark2022unified, ludziejewski2025moe, zhao2025moescaling}.

\subsection{Leave-One-Out Cross-Validation}
\label{sec:loo}

To validate that the calibration generalizes rather than overfitting to the 93-model training set, we perform leave-one-out cross-validation (LOO-CV): for each open model, we refit the regression on the remaining 92 models and predict the held-out model's parameter count from its accuracy.

\begin{figure}[H]
    \centering
    \includegraphics[width=\columnwidth]{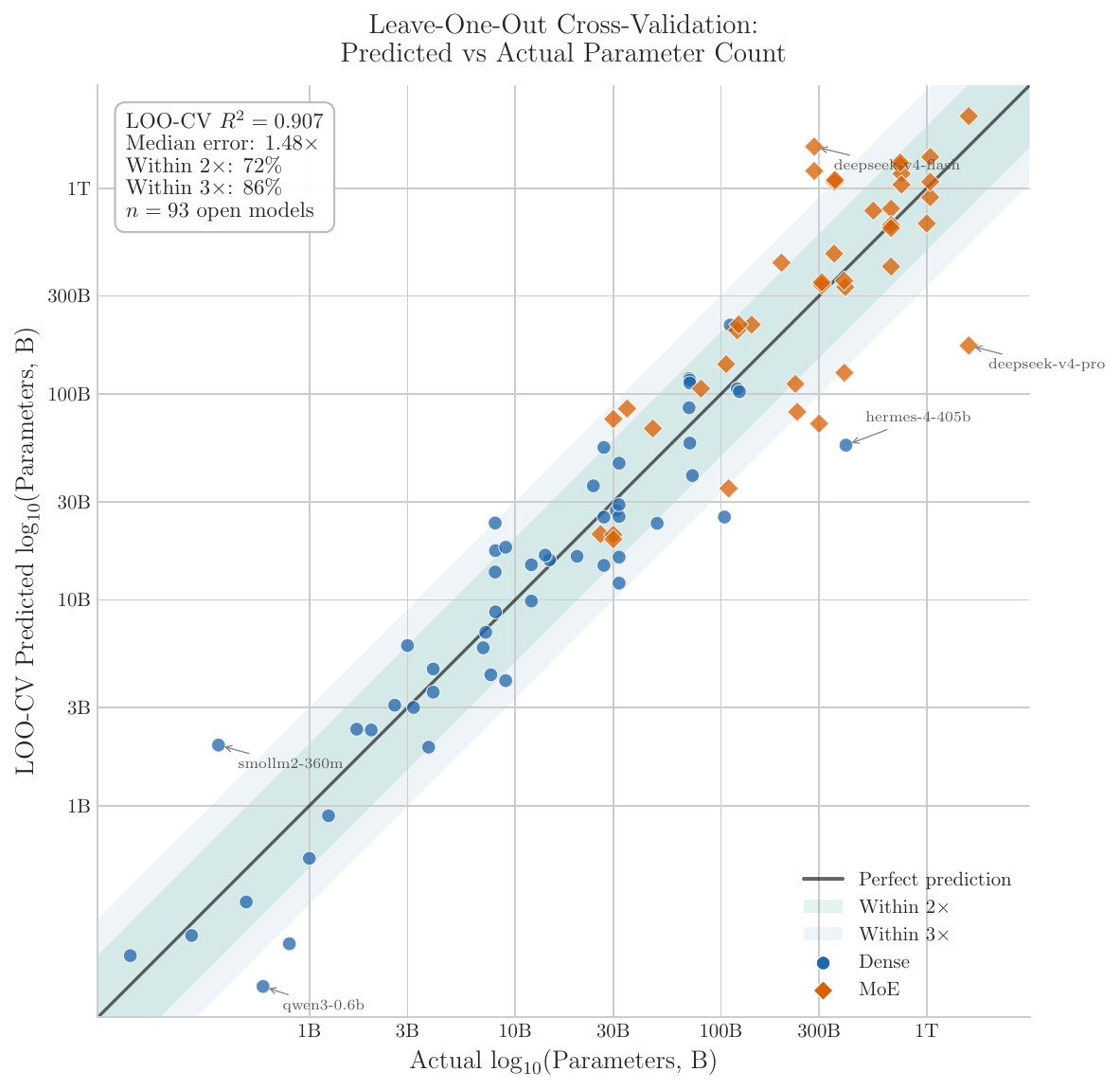}
    \caption{Leave-one-out cross-validation: predicted vs actual parameter count for 93 open models. Green band: within $2\times$; blue band: within $3\times$. Median fold error is $1.48\times$; $72\%$ within $2\times$, $86\%$ within $3\times$. The 90\% prediction interval factor is $3.20\times$.}
    \label{fig:loo}
\end{figure}

Figure~\ref{fig:loo} shows predicted versus actual parameter counts. The LOO median multiplicative fold error is $1.48\times$: $72\%$ of models are predicted within $2\times$ of their true size and $86\%$ within $3\times$, with a 90\% prediction interval factor of $3.20\times$. The worst outliers in either direction are Nemotron-70B (predicted ${\sim}$490B, ECR $7.0\times$, NVIDIA's heavy RLHF post-training pass), Llama~3.1-70B and 3.3-70B (ECRs $\sim 3.5\text{--}4.2\times$), Llama~4~Scout (109B predicted as ${\sim}$18B, ECR $0.17\times$, suggesting heavy refusal calibration), and a cluster of small Qwen variants under-predicted around the $0.3\text{--}0.4\times$ band.

\subsection{Frontier Model Estimates}
\label{sec:frontier}

Table~\ref{tab:frontier} shows parameter estimates for the major proprietary models, all derived by inverting the calibration curve fit on 93 open-weight models. None of the entries below were used to fit the curve; the full ranking is therefore an out-of-sample readout of where each closed-source frontier model sits along the same log-linear scaling axis that produced the open-weight regression.

\begin{table}[H]
\centering
\caption{Effective knowledge capacity of proprietary frontier models, in open-model-equivalent parameters, as 90\% prediction bands (calibration: 93 open models, $R^2 = 0.910$, $\lambda = 0$; PI factor ${\sim}3.2\times$). Higher-scoring variant shown per model. Estimates above ${\sim}1$T rest on only two calibration anchors and carry wider uncertainty than the global PI factor (Section~\ref{sec:limitations}). The Gemini 3.x family is excluded (Gemini 3.1 Pro is the T6 landmark; the Flash line inherits its inflation).}
\label{tab:frontier}
{\small\input{tables/frontier_estimates}}
\end{table}

\textbf{The shape of the frontier.} Estimates are reported as 90\% prediction intervals, not point values. At the top, GPT-5.5 Pro and GPT-5.5 lead at ${\sim}5.3$T and ${\sim}4.7$T, followed by Claude Fable 5 (${\sim}3.5$T) and Gemini 2.5 Pro (${\sim}3.0$T). A broad ${\sim}2$T cluster gathers GPT-4.1, GPT-5.4 Pro, GPT-5 Pro, o3, Grok-3, and Grok-4 (all within $[0.6, 2.2]$T), indicating that competitive frontier development has converged to within ${\sim}1.5\times$ effective capacity. The ratio between the top of the proprietary fleet and the smallest estimated model (${\sim}9$B) is ${\sim}580\times$, spanning the same dynamic range as the open-weight calibration set.

\textbf{Pro tiers add little factual capacity.} OpenAI's ``Pro'' variants sit only $+1.2$ (GPT-5), $+0.8$ (GPT-5.2), and $+0.9$ (GPT-5.5) pp above their non-Pro siblings---effective-capacity premiums of $1.1$--$1.2\times$---consistent with the Pro tier targeting reasoning, agentic, and long-context capability through stronger post-training and longer inference-time budgets rather than added \emph{stored facts}. The lone exception is GPT-5.4 Pro ($+5.5$ pp, $2.2\times$), which we do not over-interpret given the ${\sim}3.2\times$ prediction-interval width. IKP measures factual capacity specifically, so a near-flat Pro-vs-base gap is the expected outcome and does not contradict the Pro tier's purpose (Section~\ref{sec:theory}).

\textbf{Point releases reshuffle; major versions scale up.} Within OpenAI's GPT-5.x series the non-thinking scores span $68$--$83\%$: GPT-5 at $75.8\%$, the 5.1--5.4 point releases clustered at $68$--$72\%$, and GPT-5.5 stepping clear to $82.6\%$. Read literally, the point releases share a similar parameter budget while GPT-5.5 is a genuine scale-up; the fingerprint analysis (Section~\ref{sec:fingerprint}) agrees, placing every consecutive GPT-5$\to$5.\texttt{x} transition in the \emph{retrained} regime (HSS $<0.10$) rather than the lineage regime. The Claude Opus non-thinking line, by contrast, is \emph{non-monotonic} under no-penalty scoring (4.5 at ${\sim}892$B, 4.6 at ${\sim}1.3$T, 4.7 at ${\sim}538$B, 4.8 at ${\sim}236$B). This is not evidence of shrinking models: the later Opus releases refuse far more often (Section~\ref{sec:silent-tax}), which depresses raw accuracy and hence the no-penalty estimate. Under a mild penalty (Table~\ref{tab:lambda-sweep}) the ordering largely reverts; like all heavily refusal-trained models, these estimates are lower bounds (Section~\ref{sec:silent-tax}).

\subsection{Per-Tier Discrimination}

\begin{figure}[!htbp]
    \centering
    \includegraphics[width=\textwidth]{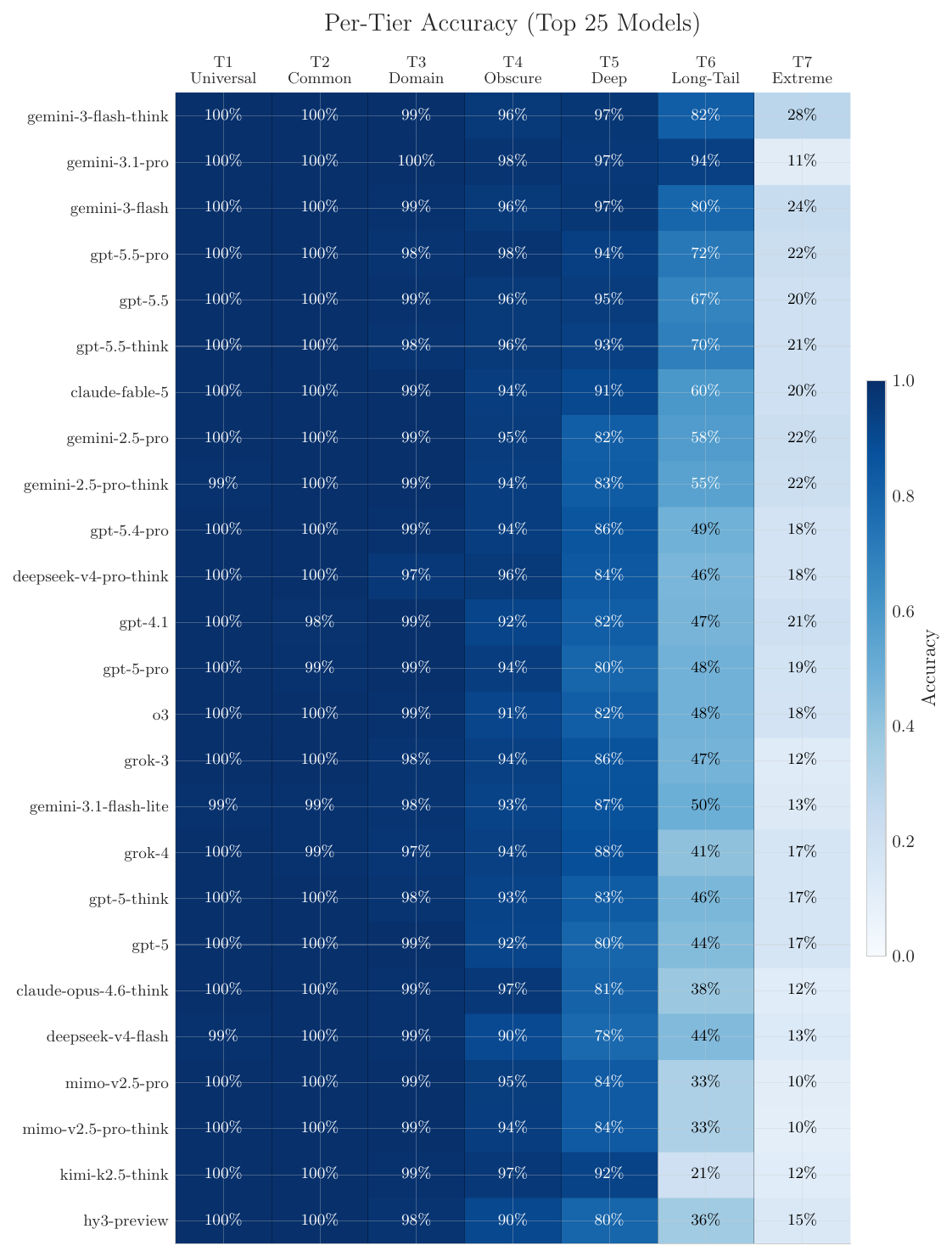}
    \caption{Per-tier accuracy for the top 25 models. Each row is a model (sorted by overall accuracy); each column is a tier (T1--T7). The step-function pattern is clear: T1--T2 are saturated (dark blue), T3--T5 provide the main discrimination, and T6--T7 separate only the strongest frontier models.}
    \label{fig:heatmap}
\end{figure}

Each obscurity tier acts as a different rung of the parameter ladder. Fitting a log-linear regression of per-tier accuracy against $\log_{10}(N)$ on the 93-model calibration set reveals a striking stratification (Table~\ref{tab:scaling}, no-penalty scoring): T1's slope is only $0.072$ per decade; T2 $0.144$; T3 $0.275$; T4 has the steepest slope ($0.297$ per decade) and does not saturate within the calibration range; T5 $0.224$; T6 drops to $0.077$ (nearly flat in log-parameters within the open-weight set); and T7 $\approx 0.022$. Each tier is a step function whose midpoint sits at a different parameter scale, and the seven midpoints together span the ladder. Figure~\ref{fig:heatmap} shows the resulting step-function pattern.

\textbf{T3 is the most informative single tier.} It combines a steep slope ($0.275$ per decade), the highest Spearman correlation with overall accuracy ($\rho = 0.965$), and the widest within-set spread (near-zero on sub-1B models up to $97\%$ on the largest open-weights), so a model's T3 score alone is a strong proxy for its overall ranking when compute precludes a full evaluation. T2 is less informative ($\rho = 0.742$) and saturates earlier; T1 is approximately a binary check (is this even a real LM?) and contributes essentially nothing to the discrimination of frontier models. Practically, a 200-probe sub-evaluation drawing primarily from T3 (with a thin T1/T2 sanity check and a thin T5/T6 ceiling check) recovers most of the signal of the full benchmark.

\textbf{T6 discriminates the frontier.} Under no-penalty scoring the open-weight T6 slope is a modest $0.077$/decade, and large open models do register some T6 accuracy (DeepSeek V4 Pro at $16.5\%$ non-thinking, $46.4\%$ thinking), so T6 is a softer separator than under penalized scoring. Even so, the strongest proprietary models lead clearly: GPT-5.5 at $67.4\%$, GPT-5.5-Pro at $72.2\%$, GPT-5.5-Think at $69.5\%$, GPT-5 at $43.9\%$, Claude Opus 4.6-Think at $38.0\%$, o1 at $32.1\%$, and Opus 4.7 at $20.3\%$. The Gemini 3.x family posts the highest T6 scores (up to $94\%$), inflated by construction because Gemini 3.1 Pro is the T6 calibration landmark---hence their exclusion from the frontier table. The T6 gap between the leading proprietary models and the open-weight cluster remains the clearest signal of an effective-capacity step beyond $\sim$$2$T.

\textbf{T7 is not a hard ceiling.} We do not treat any tier as a permanent ceiling. An earlier version of IKP, scored with a penalty, reported near-zero T7 accuracy for every model and used T7 as a saturation ``anchor''; that was a scoring artifact---the penalty netted correct answers against bluffs. At $\lambda=0$, T7 accuracy is nonzero for most models (median ${\sim}8\%$) and reaches ${\sim}22$--$28\%$ at the frontier. We do not read this as a clean knowledge measurement: part of it reflects lucky or partially-informed guesses, and ${\sim}9$ T7 probes are miscategorized (too easy; Limitations). T7 is simply the least-saturated tier and the main headroom for future obscurity scaling.

\textbf{Saturation thresholds give a back-of-envelope size estimator.} Reading the table of slopes and intercepts in reverse, the parameter scale at which each tier crosses a given accuracy threshold is informative. High T2/T3 accuracy implies frontier-scale counts, $50\%$ T4/T5 implies several-hundred-billion parameters, and meaningful T6 accuracy implies effective capacity above the calibration range. A practitioner can thus read the bands directly from per-tier accuracies without inverting the joint regression.

\subsection{Thinking Mode Analysis}

\begin{figure}[H]
    \centering
    \includegraphics[width=\columnwidth]{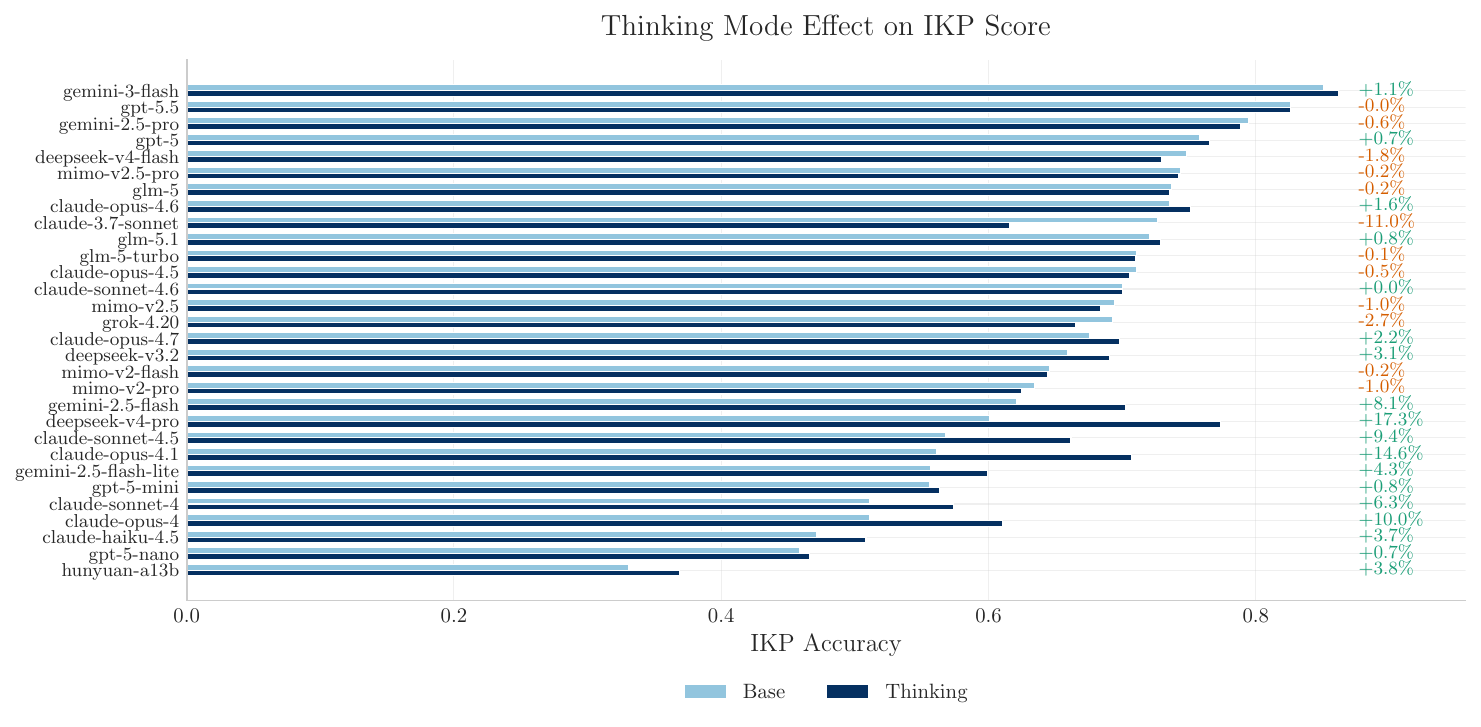}
    \caption{Thinking mode effect across 30 base/think model pairs. Light bars: base model accuracy; dark bars: thinking variant. Thinking improves accuracy in 18 of 30 cases (mean $+2.3$ pp). Regressions on the remaining pairs (up to $-11$ pp for Claude 3.7 Sonnet) are largely due to increased refusal conservatism in thinking mode.}
    \label{fig:thinking}
\end{figure}

Across 30 base/think pairs, thinking mode~\citep{wei2022chain} improves accuracy in 18 cases (mean $+2.3$ pp, range $-11.0$ to $+17.3$ pp; Figure~\ref{fig:thinking}). The largest gains are DeepSeek V4 Pro ($+17.3$ pp) and Claude Opus 4.1 ($+14.6$ pp); the largest regression is Claude 3.7 Sonnet ($-11.0$ pp), with smaller regressions on Grok-4.20 and MiMo variants plausibly reflecting increased refusal conservatism in thinking mode. The benefit peaks at T3--T4 (medium-hard tiers) and vanishes at T7, supporting the interpretation that chain-of-thought helps with knowledge \emph{retrieval} but does not create new \emph{stored} knowledge.

\subsection{Cross-Generation Improvement}

Within families, newer generations \emph{often} know more, but not monotonically. Claude Opus rises $4\to4.6$ ($51.1\to73.5\%$) then declines at $4.7$ ($67.6\%$) and $4.8$ ($61.9\%$); GLM jumps $4$-32B$\to$5 ($50.6\to73.7\%$) then plateaus; GPT-4$\to$4o \emph{drops} ($68.6\to65.4\%$); and Claude 3.5 Haiku$\to$Haiku 4.5 loses $7.2$ pp ($54.3\to47.1\%$). These late-generation declines coincide with higher refusal rates (Section~\ref{sec:silent-tax}); since under $\lambda=0$ a more cautious model scores lower at equal knowledge, they likely reflect refusal policy more than knowledge loss, and within-family ordering near the frontier tracks refusal policy as much as capacity.

\subsection{Densing Law Falsification}
\label{sec:densing-falsification}

A central empirical implication of our framework (Section~\ref{sec:theory}) is that the Densing Law~\citep{densing2025} should \emph{not} hold on IKP. If factual capacity is incompressible, then at fixed $\log_{10}(N)$ there should be no per-month improvement in IKP accuracy---in contrast with reasoning benchmarks, on which capability-per-parameter doubles every ${\sim}3.5$ months.

\textbf{Design.} We assembled release dates (YYYY-MM-DD, verified against vendor announcements and HuggingFace model cards) for all 201 models in our evaluation. For the 100 open-weight models with published parameter counts and IKP results, spanning 2023-09-27 to 2026-06-12, we fit
\[
  \text{pen\_acc} = \beta_0 + \beta_1 \log_{10}(N_B) + \beta_2 \cdot \text{months},
\]
where \texttt{months} is release date measured relative to 2024-01-01. The baseline $\beta_1 = 0.149$ (R$^2 = 0.815$) implies, under the Densing Law, a monthly accuracy gain of $\beta_2^{\mathrm{Densing}} = \beta_1 \log_{10}(2)/3.5 \approx +0.0129$/month ($\approx +15.4$ pp/year).

\begin{figure}[H]
    \centering
    \includegraphics[width=\columnwidth]{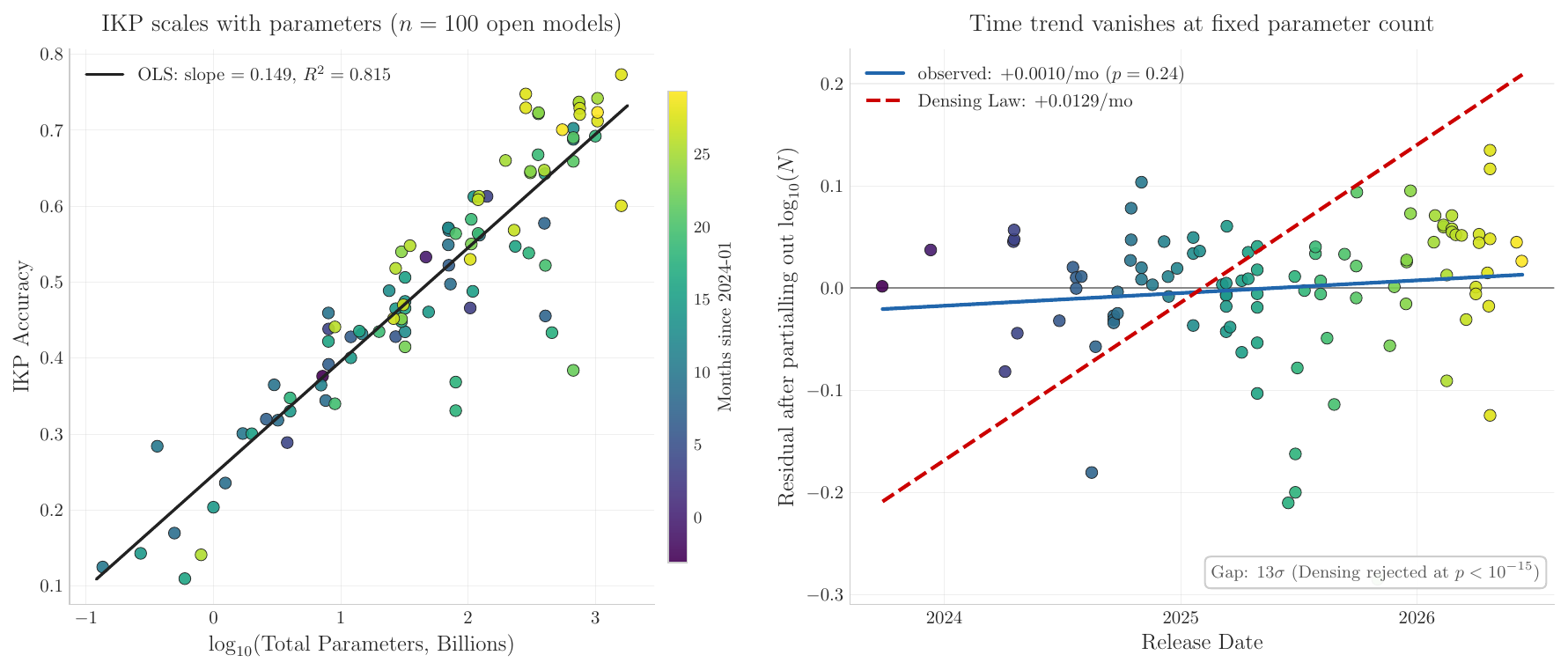}
    \caption{Densing Law falsification on IKP across 100 open-weight models (2023-09 to 2026-06). \textbf{Left:} IKP accuracy vs $\log_{10}$(params); points colored by release month. \textbf{Right:} residuals after partialling out $\log_{10}$(params), vs release date. The observed trend (solid blue, $+0.0013$/month) is indistinguishable from zero, while the Densing-Law prediction (dashed red, $+0.0129$/month) would imply ${\sim}15$ pp/year of per-parameter improvement. The gap exceeds $11\times$ the bootstrap standard error.}
    \label{fig:densing}
\end{figure}

\textbf{Result.} The fitted coefficient is $\hat\beta_2 = +0.0013$/month [95\% bootstrap CI: $-0.0004$, $+0.0033$]. The point estimate is indistinguishable from zero ($p = 0.19$) and rejects the Densing prediction at $p < 10^{-15}$. Adding release date to the scaling regression increases $R^2$ by only $+0.0033$. Controls for thinking mode and MoE architecture do not change this conclusion (Appendix~\ref{app:densing}, Table~\ref{tab:densing-full}): across specifications the time coefficient stays within $\pm 0.002$/month, while the Densing prediction is never within the confidence interval. Figure~\ref{fig:densing} partials out $\log_{10}(N)$ and contrasts the flat observed residual-vs-date relationship with the steep Densing-Law prediction line.

\textbf{Interpretation.} Capability density has indeed increased on reasoning benchmarks over 2024--2026, but the improvement has come from more efficient use of procedural and linguistic parameters ($N_\mathrm{proc}, N_\mathrm{ling}$), not from more efficient storage of facts. Smaller recent models know no more than older models of the same size---exactly the prediction of incompressibility (Section~\ref{sec:theory}). The positive marginal $R^2$ of time alone (Table~\ref{tab:densing-full}) is fully accounted for by the selection effect that newer releases skew larger.

\subsection{Why not just use MMLU? Standard knowledge benchmarks as size proxies}
\label{sec:benchmark-comparison}

\begin{figure}[H]
    \centering
    \includegraphics[width=\textwidth]{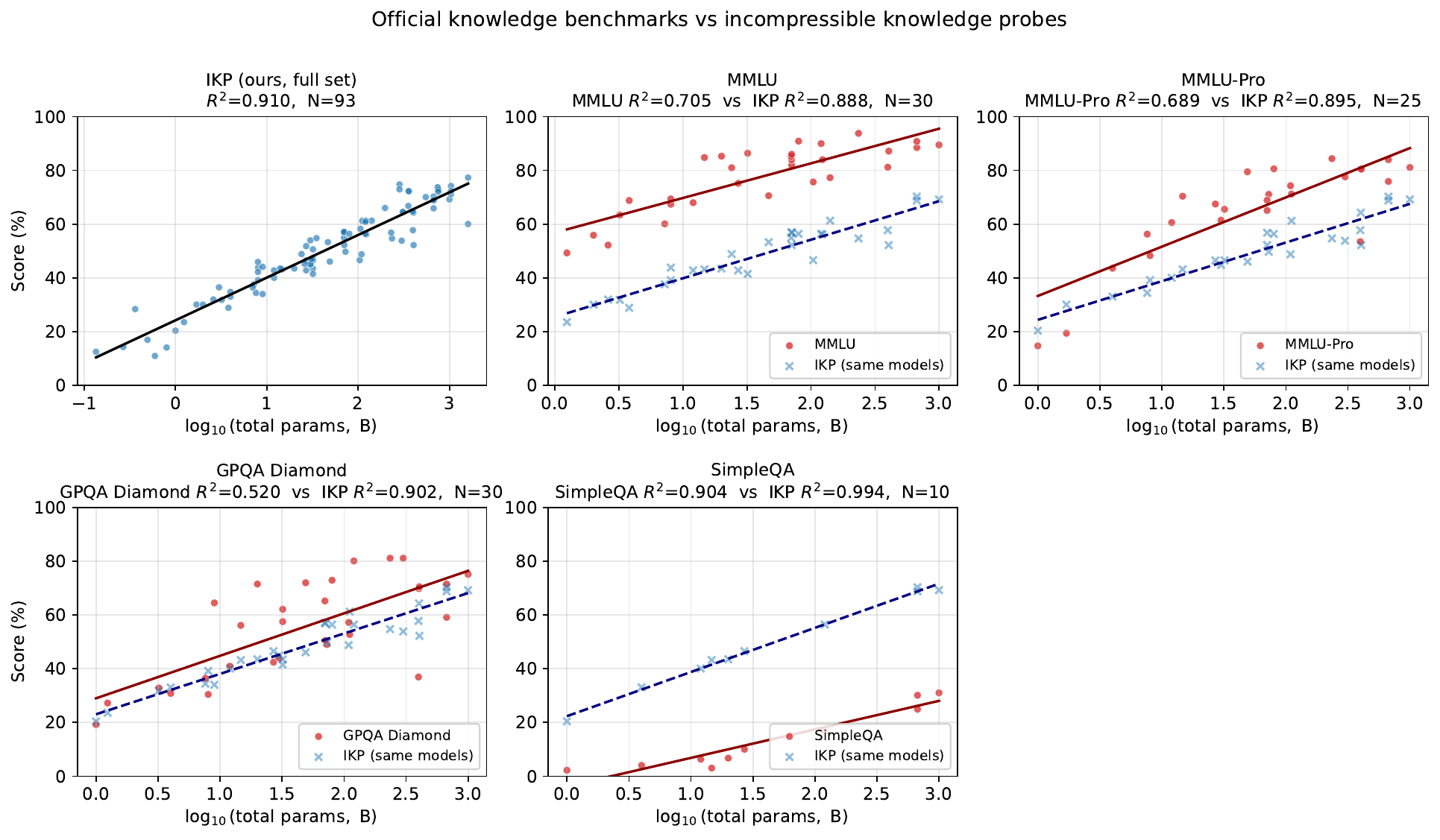}
    \caption{Vendor-reported benchmark scores vs $\log_{10}$(total params, B). Top-left: IKP across all 93 calibration models ($R^2 = 0.910$). Other panels: each standard benchmark (red) plotted against the same models with IKP (blue $\times$, dashed line) overlaid for direct comparison. On every matched subset, IKP yields a tighter fit; the gap is largest for GPQA Diamond (reasoning-heavy) and smallest for SimpleQA (factual).}
    \label{fig:benchmark-comparison}
\end{figure}

A natural question is whether existing knowledge benchmarks could substitute for IKP. We collected vendor-published official scores from primary sources (model cards, system cards, technical reports, vendor blog posts) for four widely-reported benchmarks: MMLU~\citep{hendrycks2021measuring}, MMLU-Pro~\citep{wang2024mmlupro}, GPQA Diamond~\citep{rein2023gpqa}, and SimpleQA~\citep{wei2024simpleqa}. We then fit each benchmark as a parameter-count proxy ($\text{score} \sim \log_{10} N$) and as a Densing-style joint fit ($\text{score} \sim \log_{10} N + \text{months}$), restricting to the same calibration set used for IKP. Crucially, we re-fit IKP on each matched subset so the comparison is apples-to-apples (Table~\ref{tab:benchmark-comparison}, Figure~\ref{fig:benchmark-comparison}).

\begin{table}[H]
\centering
\small
\begin{tabular}{lrrrr}
\toprule
Metric & $N$ & $R^2$ vs $\log_{10} N$ & IKP $R^2$ (same subset) & Time slope (pp/month) \\
\midrule
\textbf{IKP (full set)}      & \textbf{93} & \textbf{0.910} & ---   & \textbf{$+0.09$} \\
SimpleQA                     & 10          & 0.904          & 0.991 & $+0.03$ \\
MMLU                         & 30          & 0.705          & 0.886 & $+0.58$ \\
MMLU-Pro                     & 25          & 0.689          & 0.900 & $+0.82$ \\
GPQA Diamond                 & 30          & 0.520          & 0.903 & $+1.99$ \\
\bottomrule
\end{tabular}
\caption{Standard knowledge benchmarks vs IKP as parameter-count proxies. On every matched subset of dated open-weight models, IKP explains substantially more variance in $\log_{10}(N)$ than the standard benchmark does. The right-most column reports the joint fit's time slope (percentage points per month) at fixed $\log_{10}(N)$: reasoning-heavy benchmarks (GPQA Diamond, MMLU-Pro) drift fastest, exactly as the Densing Law predicts; the pure-factual benchmark (SimpleQA) and IKP drift near zero. Sample sizes differ because vendors do not report all benchmarks for all models, and we exclude scores from with-search or tools-augmented configurations.}
\label{tab:benchmark-comparison}
\end{table}

\textbf{Three findings.} (i) On every matched subset IKP wins on $R^2$, with the gap largest for the most reasoning-heavy benchmark (GPQA Diamond: $0.52$ vs $0.90$) and smallest for the only purely-factual benchmark (SimpleQA: $0.90$ vs $0.99$). (ii) Reasoning benchmarks drift sharply over time at fixed $\log_{10}(N)$: GPQA Diamond gains $\approx 2$ pp/month, meaning a $33$B model improves by $\approx 24$ points across one year of releases without growing. This is exactly the Densing prediction, and it directly invalidates these benchmarks as parameter proxies. (iii) Factual benchmarks behave like IKP: SimpleQA's time slope is $+0.03$ pp/month, statistically indistinguishable from zero, supporting the broader claim that the incompressibility property holds for the factual subspace specifically rather than for ``benchmarks'' as a category.

\textbf{Coverage matters.} Vendors publish MMLU-Pro, GPQA, and SimpleQA scores predominantly for upper-mid and frontier models. Tiny pretrained models (smollm2-135m, gemma-3-270m) and many older mid-range models do not appear on these benchmark tables at all because they score near random. IKP, evaluated identically on all $93$ calibration models from $135$M to $1.6$T parameters ($> 4$ orders of magnitude), avoids this sample-selection asymmetry. Even setting aside the Densing-Law time confound, this coverage difference is itself a methodological argument: a parameter-estimation instrument needs to work across the full size range, including the small models that anchor the bottom of the calibration curve.

\subsection{Knowledge Fingerprinting: Lineage vs.\ Retraining}
\label{sec:fingerprint}

\begin{figure}[H]
    \centering
    \includegraphics[width=\columnwidth]{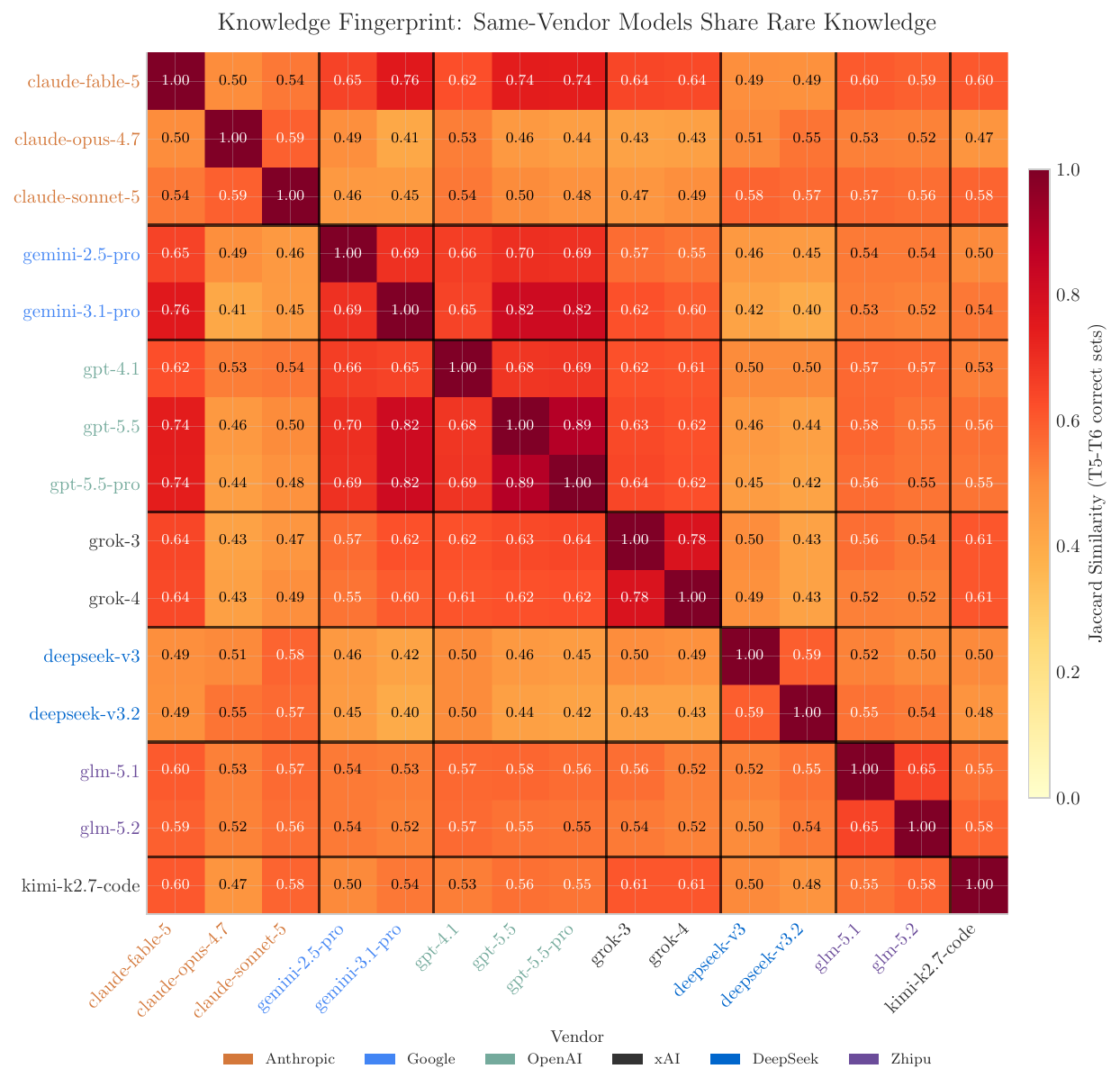}
    \caption{Jaccard similarity on T5--T6 correct-answer sets for 15 frontier models, clustered by vendor. Within-vendor similarity (diagonal blocks) is consistently higher than cross-vendor (off-diagonal). Jaccard alone does not separate shared-base from independent-but-competitive models; see Figure~\ref{fig:lineage} for the sharper HSS view.}
    \label{fig:fingerprint}
\end{figure}

The specific set of rare facts (T5--T6) a model knows constitutes a \emph{knowledge fingerprint} (Figure~\ref{fig:fingerprint}), complementing output-distribution fingerprints~\citep{tsai2025rofl} and embedded watermarks~\citep{nasery2025fingerprinting} with a training-free, knowledge-based signal. For each pair of models we compute three metrics on the 400 T5--T6 probes:

\begin{itemize}[leftmargin=*]
    \item \textbf{Jaccard similarity} $J$ on correct-answer sets. Inflated by probes that almost every frontier model knows.
    \item \textbf{Lift} = observed intersection / expected-under-independence. Controls for common knowledge.
    \item \textbf{Hallucination similarity} $\mathrm{HSS}$: among probes where both models are wrong (non-refusal), the fraction on which they produce the \emph{same} normalized wrong answer. $\mathrm{HSS}$ is the most diagnostic of the three because independently trained models essentially never converge on an identical wrong rare fact, whereas weight-sharing siblings do so on 30--55\% of shared-wrong probes.
\end{itemize}

\begin{figure}[H]
    \centering
    \includegraphics[width=\textwidth]{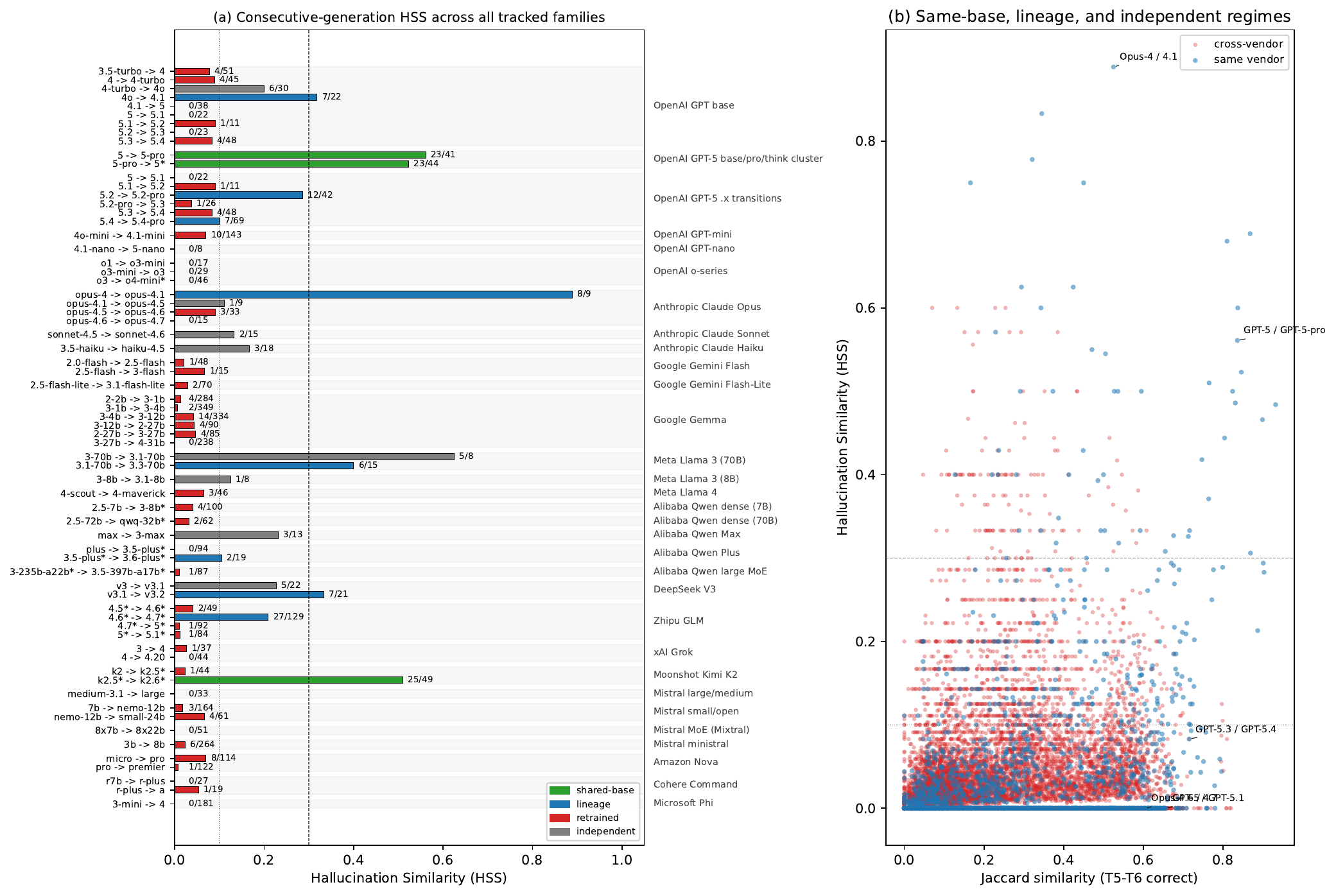}
    \caption{(a)~Hallucination-similarity ($\mathrm{HSS}$) for consecutive-generation pairs within each family. Annotation is $\text{same\_wrong}/\text{both\_wrong}$. Dashed line: shared-base threshold (0.30); dotted line: lineage threshold (0.10). GPT-5 through 5.4 transitions all fall in the retrained regime. (b)~All pairs with $\geq 5$ joint-wrong probes in the $(J,\mathrm{HSS})$ plane. Same-vendor (blue) and cross-vendor (red) pairs largely overlap on $J$ but separate on $\mathrm{HSS}$.}
    \label{fig:lineage}
\end{figure}

\paragraph{Three regimes.} Applied pairwise across 201 models, the three metrics cleanly separate into three regimes (Figure~\ref{fig:lineage}a).\footnote{The regime boundaries are empirical: weight-sharing siblings (e.g., GPT-5 / GPT-5-pro at $\mathrm{HSS} = 0.56$, GPT-5-pro / GPT-5-think at $0.52$) concentrate at $\mathrm{HSS} \geq 0.30$, whereas cross-vendor pairs with $\geq 10$ joint-wrong probes have median $\mathrm{HSS} = 0.00$ (mean $0.02$).}
\emph{Shared base} pairs ($\mathrm{HSS} \geq 0.30$, $J \geq 0.60$) represent the same weights served with different inference or light alignment; \emph{lineage} pairs ($0.10 \leq \mathrm{HSS} < 0.30$) are consistent with post-training, continued pretraining, or distillation on top of a shared ancestor; \emph{retrained} pairs ($\mathrm{HSS} < 0.10$ on $\geq 10$ joint-wrong probes) fall in the same statistical regime as cross-vendor independent pairs even when the models share a version-family label.

\paragraph{Within-family lineage (selected).}
Applied to consecutive-generation pairs within each vendor's family, this test gives a varied picture of release practices---though the joint-wrong counts are small (7--71 probes), so individual provenance calls are suggestive, not conclusive, and cannot be independently verified for closed models:
\begin{itemize}[leftmargin=*]
    \item \textbf{OpenAI GPT-5 family.} GPT-5, GPT-5-pro and GPT-5-think form a tight shared-base cluster ($\mathrm{HSS} = 0.56, 0.52$). But every ``\texttt{.x}'' transition---GPT-5 $\to$ 5.1, 5.1 $\to$ 5.2, 5.2-pro $\to$ 5.3, and 5.3 $\to$ 5.4---has $\mathrm{HSS} < 0.10$ on 11--48 joint-wrong probes. This is the retrained regime, consistent with GPT-5.3 and GPT-5.4 being independent training runs rather than post-trains of GPT-5---though the fingerprint cannot distinguish a from-scratch run from a substantially changed base. GPT-5.4 vs. GPT-5.4-pro ($\mathrm{HSS} = 0.10$, 69 joint-wrong) is instead a lineage pair, consistent with a shared base differentiated only at the alignment stage.
    \item \textbf{Anthropic Claude Opus.} Opus 4 $\to$ 4.1 is clear lineage ($\mathrm{HSS} = 0.89$, 9 joint-wrong; also supported by Anthropic's own description of 4.1 as a post-training update). Opus 4.5 $\to$ 4.6 ($\mathrm{HSS} = 0.09$, 33 joint-wrong) and Opus 4.6 $\to$ 4.7 ($\mathrm{HSS} = 0.00$, 15 joint-wrong) are both in the retrained regime. Sonnet 4.5 $\to$ 4.6 sits at the lineage boundary ($\mathrm{HSS} = 0.13$, 15 joint-wrong).
    \item \textbf{Google Gemini.} Every cross-generation Flash and Flash-Lite pair is retrained (Flash 2.0 $\to$ 2.5: $\mathrm{HSS} = 0.02$, 48 joint-wrong; Flash 2.5 $\to$ 3: $0.07$, 15; Flash-Lite 2.5 $\to$ 3.1: $0.03$, 70). Gemini generations appear to be full re-trains, not post-training increments.
    \item \textbf{DeepSeek V3.} V3 $\to$ V3.1 ($\mathrm{HSS} = 0.23$, lineage) and V3.1 $\to$ V3.2 ($0.33$, at the shared-base boundary) are consistent with incremental continued pretraining on a shared base.
    \item \textbf{Zhipu GLM.} GLM 4.5 $\to$ 4.6 (retrained, $0.04$), 4.6 $\to$ 4.7 (lineage, $0.21$), 4.7 $\to$ 5 (retrained, $0.01$) alternate, consistent with a pattern of minor point releases interleaved with full retrains.
\end{itemize}

\paragraph{Cross-family outliers.} Applying the same test to the ${\sim}14{,}000$ cross-vendor pairs flags a small number as $\mathrm{HSS} \geq 0.20$ with $\geq 10$ joint-wrong probes---the regime occupied by shared weights within a vendor. The strongest signal (across multiple tests) is Baidu ERNIE 4.5, which achieves $\mathrm{HSS} = 0.36$--$0.50$ against GPT-4o, Llama-3-70B, Mistral and Qwen-Max simultaneously---a pattern consistent with heavy training on mixed distilled outputs rather than a single teacher. Llama 3.1 70B shows up as an apparent ``teacher'' in a striking number of pairings ($\mathrm{HSS} \geq 0.30$ against grok-3, gemini-2.0-flash, qwen3-max, GPT-4.1-nano, and several other models), which is most likely an artifact of Llama 3.1 being the most widely used open base for synthetic-data generation in 2024. GPT-5 vs.\ Grok-4 ($\mathrm{HSS} = 0.38$, 21 joint-wrong) and GPT-5-pro vs.\ Kimi-K2.6 ($\mathrm{HSS} = 0.32$, 22 joint-wrong) are individually above threshold and merit follow-up, though neither is large enough, at our probe count, to reject an innocent-explanation null with confidence. Full rankings and raw counts are in Appendix~\ref{app:fingerprint}.

\subsection{What Determines Whether an LLM Knows Something?}
\label{sec:recognition}

This is the question that motivated the IKP instrument in the first place (Section~\ref{sec:intro}): a researcher or an entity is ``inside'' a model to the extent that the model reliably produces the right answers to probes about it. Here we turn the instrument on itself and ask what observable properties of a researcher, or of a fact, predict whether a model has absorbed it. We cross-reference IKP recognition rates with external metrics for researcher probes (OpenAlex citations and h-index, $n = 345$) and for factual probes about entities (Wikipedia sitelinks and pageviews, $n = 557$). Both analyses point to the same interpretation: recognition tracks not raw prominence but how often the specific fact appears in retrievable form in text---a latent \emph{effective mention frequency} that citations and sitelinks only partially proxy and that we do not observe directly.

\subsubsection{Researcher Probes}
\label{sec:recognition-researchers}

\begin{figure}[H]
    \centering
    \includegraphics[width=\columnwidth]{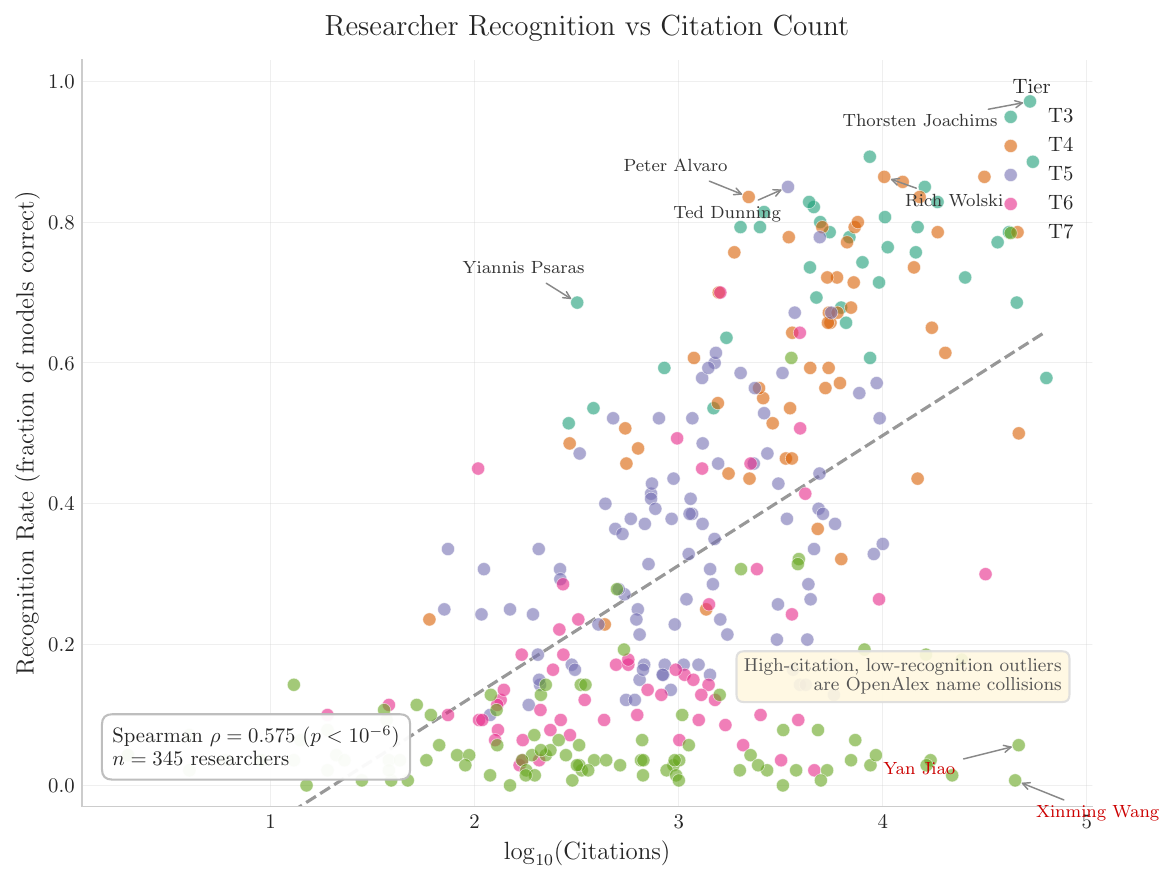}
    \caption{Researcher recognition rate vs log(citations), colored by tier. The moderate correlation ($\rho = 0.575$) is consistent with name uniqueness and subfield-ecosystem effects beyond raw citation count.}
    \label{fig:researcher}
\end{figure}

\textbf{Bibliometric signals explain roughly a third of the variance.} Across 345 researcher probes scored by 140 models, log-transformed OpenAlex citations correlate with recognition rate at Spearman $\rho = 0.575$ (Pearson $r = 0.590$; Figure~\ref{fig:researcher}); log-transformed h-index correlates at $\rho = 0.561$. Tier assignment is well-predicted by citations (median citations drop from 6{,}859 at T3 to 325 at T7, Spearman $\rho = -0.51$), indicating that tiers track bibliometric mass at the population level. But at the individual-probe level, citations and h-index together explain only ${\sim}35\%$ of variance in recognition. Table~\ref{tab:citation-hindex-buckets} makes the shallow slope explicit: recognition grows monotonically with either metric, yet no threshold cleanly separates recognized from unrecognized researchers. Even at h-index $\geq 50$, mean recognition is $0.55$ and only $65\%$ are recognized by a majority of models---nearly half the variance sits elsewhere.

\textbf{High citation is approximately necessary but not sufficient for high recognition.} The asymmetry is visible directly in Figure~\ref{fig:researcher}: the upper-left region is essentially empty. No researcher with $<50$ citations exceeds $15\%$ recognition, and no researcher with $<500$ citations exceeds $75\%$. The lower-right region, by contrast, is densely populated---many researchers with $10$K+ citations sit below $25\%$ recognition. Citation count therefore acts as a probabilistic upper bound on recognition (a low-cited researcher rarely becomes a household name to a frontier model), but does not lift recognition on its own; the gap between the bound and the realized recognition rate is filled by the artifact-and-derivative-content mechanisms detailed in the audit below. The exceptions to the necessary-condition reading---low-cited researchers with high recognition, e.g.\ Tri Dao (3K, 100\%), Eyuboglu (267, 57\%), Psaras (318, 69\%)---are each attached to a named, widely-distributed artifact (FlashAttention, ColBERT-style work, IPFS), consistent with artifact attachment---rather than citation mass alone---being the route to escaping the citation ceiling.

\begin{table}[H]
\centering
\small
\caption{Recognition rate by bibliometric bucket ($n = 345$ researchers, 140 models). Recognition grows sub-logarithmically in both citations and h-index, and the top bucket does not saturate---a substantial fraction of even the most-cited researchers are unknown to most models.}
\label{tab:citation-hindex-buckets}
\begin{tabular}{lrrr@{\hskip 18pt}lrrr}
\toprule
\multicolumn{4}{c}{\textbf{Citations}} & \multicolumn{4}{c}{\textbf{h-index}} \\
Bucket & $n$ & Mean recog. & $>\!50\%$ recog. & Bucket & $n$ & Mean recog. & $>\!50\%$ recog. \\
\midrule
0--99        & 36  & 0.07 & 0\%   & 0--4   & 43  & 0.09 & 2\%   \\
100--499     & 78  & 0.16 & 10\%  & 5--9   & 61  & 0.14 & 7\%   \\
500--999     & 47  & 0.24 & 19\%  & 10--19 & 99  & 0.28 & 25\%  \\
1{,}000--4{,}999 & 111 & 0.38 & 37\% & 20--29 & 48  & 0.40 & 42\%  \\
5{,}000--9{,}999 & 41  & 0.52 & 61\% & 30--49 & 63  & 0.51 & 62\%  \\
10{,}000$+$  & 32  & 0.59 & 69\%  & 50$+$  & 31  & 0.55 & 65\%  \\
\bottomrule
\end{tabular}
\end{table}

\textbf{Subfield predicts recognition beyond bibliometric mass.} Table~\ref{tab:recog-by-field} breaks recognition down by CS subfield. Information retrieval (IR) and programming languages (PL) researchers are recognized $1.5$--$2\times$ more often than computer-architecture researchers at comparable or higher citation levels; HCI and theoretical CS sit near the bottom despite high mean bibliometric density. The subfields at the top tend to be those with a dense derivative-content layer---tutorials, blog posts, course materials, library documentation (which we do not measure directly)---suggesting that what a model memorizes may be less the research itself than the web text that surrounds it. Subfield means rest on small samples (e.g., IR $n=7$) and should be read as indicative.

\begin{table}[H]
\centering
\small
\caption{Recognition rate by CS subfield, sorted by mean recognition ($n \geq 6$). IR and PL lead despite moderate bibliometric mass; HCI, theoretical CS and computer architecture lag despite comparable or higher citation averages.}
\label{tab:recog-by-field}
\begin{tabular}{lrrrr}
\toprule
Subfield & $n$ & Avg.\ citations & Avg.\ h-index & Mean recog. \\
\midrule
Information retrieval         & 7  & 10{,}497 & 29.6 & 0.571 \\
Programming languages         & 43 & 2{,}596  & 19.3 & 0.437 \\
Database systems              & 7  & 3{,}670  & 26.7 & 0.429 \\
Data mining                   & 6  & 7{,}039  & 34.3 & 0.370 \\
Computer networking           & 60 & 7{,}840  & 29.8 & 0.354 \\
Distributed systems           & 42 & 2{,}206  & 16.1 & 0.307 \\
Natural language processing   & 7  & 1{,}569  & 13.6 & 0.302 \\
Computer security             & 62 & 2{,}597  & 18.5 & 0.283 \\
Computer vision               & 9  & 10{,}347 & 32.8 & 0.283 \\
Operating systems             & 38 & 3{,}718  & 20.6 & 0.277 \\
Embedded systems              & 11 & 1{,}414  & 14.0 & 0.277 \\
Human-computer interaction    & 7  & 10{,}051 & 35.6 & 0.260 \\
Theoretical CS                & 11 & 7{,}563  & 23.3 & 0.204 \\
Computer architecture         & 32 & 3{,}151  & 19.5 & 0.179 \\
\bottomrule
\end{tabular}
\end{table}

\textbf{A controlled audit isolates three mechanisms.} To characterize the residual variance, we ran a web-presence audit on a $2\times2\times2$ matrix of 20 researchers: \{ML, Systems\} $\times$ \{high citations, low citations\} $\times$ \{high recognition, low recognition\} (``ML'' $=$ core ML/AI subfields; ``Systems'' $=$ networking, OS, architecture, security). For each researcher we recorded personal site, Wikipedia page, GitHub-artifact stars, secondary-content presence (tutorials, blog posts, course materials), and name uniqueness. Table~\ref{tab:8cell} reports the cell-level outcomes; three patterns emerge.

\begin{table}[H]
\centering
\small
\caption{Controlled $2\times2\times2$ audit of 20 researchers. Each cell lists sampled researchers with (OpenAlex citations, recognition rate). $\dagger$ marks OpenAlex name-collision errors, where the citation count belongs to a same-named researcher in a different field. The ``ML / low-recognition'' cell is empty in our sample---no ML researcher falls below 43\%.}
\label{tab:8cell}
\renewcommand{\arraystretch}{1.15}
\begin{tabular}{p{3.2cm}|p{5.1cm}|p{5.1cm}}
\toprule
 & \textbf{ML / AI} & \textbf{Systems (net / OS / arch / sec)} \\
\midrule
\textbf{High citation /\newline High recognition} & Tri Dao (3K, 100\%)\newline Raghunathan (4.7K, 100\%)\newline A.\ Gu (3K, 86\%) & Joachims (53K, 97\%)\newline Gerla (54K, 89\%) \\
\midrule
\textbf{High citation /\newline Low recognition}  & \emph{Empty: ML floor $\geq 43\%$.}\newline Mid-recognition analogs:\newline Mitchell (15K, 57\%)\newline Goel (3K, 57\%), Xie (5K, 57\%) & Yan Jiao (46K$^\dagger$, 6\%)\newline Xinming Wang (45K$^\dagger$, 1\%)\newline Dan Suciu (24K, 18\%) \\
\midrule
\textbf{Low citation /\newline High recognition}  & M.\ Chen (215, 57\%)\newline Eyuboglu (267, 57\%)\newline Dunlap (303, 43\%) & Psaras (318, 69\%)\newline Micinski (850, 59\%)\newline Lysecky (382, 54\%) \\
\midrule
\textbf{Low citation /\newline Low recognition}   & \emph{Empty in sample.} & Q.\ Rao (15, 0\%) \\
\bottomrule
\end{tabular}
\end{table}

\begin{enumerate}[leftmargin=*, itemsep=2pt]
    \item \textbf{Named artifacts dominate bibliometric mass.} The ``low-citation / high-recognition'' cells are populated by researchers attached to widely-adopted artifacts (FlashAttention, IPFS, zyBooks, a free YouTube PL course) whose names travel in derivative documents. Within our full 345-researcher sample, researchers attached to a tool with $\geq 10$K GitHub stars or a dedicated Wikipedia page are recognized at rates $\geq 86\%$ regardless of citation count; researchers without either average $0.34$ recognition at matched citation buckets. A single high-visibility artifact is worth more than an order of magnitude in additional citations.
    \item \textbf{Name uniqueness is multiplicative.} Controlling for citation count, researchers with common East Asian surnames (two-character names and single-initial given names) are recognized at $22.6\%$ in our dataset versus $44.6\%$ for uniquely spelled names---a factor-of-two attenuation associated with the name type (citation count controlled, though other correlates may remain). The same effect is consistent with why Eric Mitchell (DPO lead, ${\sim}15$K Google Scholar citations) sits at 57\%: ``Eric Mitchell'' is dominated in web indices by a film director and unrelated professionals, so even a named method cannot fully anchor the researcher name.
    \item \textbf{The ML--Systems floor gap.} The ``ML / low-recognition'' cell in Table~\ref{tab:8cell} is empty: no ML researcher in our sample falls below 43\%, including PhD students at ${<}300$ citations. The equivalent Systems cell is non-empty at arbitrarily low citation counts. The asymmetry is not individual merit; it is subfield ecosystem density---ML labs generate orders of magnitude more Twitter, blog, and podcast content per paper than Systems labs, and this derivative content is what foundation-model training pipelines scrape.
\end{enumerate}

\textbf{Name-collision data quality caveat.} The audit surfaced an under-appreciated artifact in bibliometric studies of LLM recognition: three of the four ``Systems / high-citation / low-recognition'' cell occupants were OpenAlex disambiguation errors---the citations belonged to same-named researchers in chemistry, atmospheric science, or medicine. Any future work correlating bibliometric metrics with LLM memorization should treat common-name high-citation profiles as provisional until verified against field-consistent publication lists.

\textbf{Implication for impact measurement.} The mapping from academic output to LLM-internalized knowledge is not monotonic in citation count. It is closer to \emph{citations $\times$ name uniqueness $\times$ named-artifact amplification $\times$ subfield-ecosystem density}, with each multiplicative factor spanning roughly a $2$--$5\times$ range in the regime we observe. For a working researcher, the marginal effect of one additional widely-used open-source tool with clean name attribution is larger than the marginal effect of one additional well-cited paper: the tool generates derivative documents that each carry the author's name, while the paper typically carries it only in its own bibliography record and those of its direct citers.

\subsubsection{Factual Probes About Entities}
\label{sec:recognition-wikidata}

For the 557 Wikidata-grounded factual probes---with founding-year questions the single largest category among universities, journals, museums, bridges, sports clubs, and places---the picture is sharper and more mechanistic.

\textbf{Pageviews dominate sitelinks.} Sitelink count (how many Wikipedia language editions link to the entity) correlates with recognition at Pearson $r = 0.502$ (log-transformed). For a 78-probe subset with English Wikipedia pageview data, pageviews reach $r = 0.774$ and \emph{completely subsume} sitelinks in joint regression (sitelink coefficient falls to $-0.003$ when both are included). Sitelinks measure breadth of multilingual coverage; pageviews approximate English-language discourse volume, which is what actually appears in the training corpora.

\textbf{The name--fact gap widens with prominence.} Identity facts (``what is the name of X?'') are generally easier than temporal facts (``when was X founded?''), but the gap is a function of entity prominence:

\begin{center}
\begin{tabular}{lrrrr}
\toprule
Entity prominence (sitelinks) & $n$ & Name recog. & Year recog. & Gap \\
\midrule
1--5 (obscure)     & 126 & 0.299 & 0.259 & $+0.040$ \\
6--15 (moderate)   & 127 & 0.614 & 0.533 & $+0.081$ \\
16+ (prominent)    & \phantom{0}39 & 0.728 & 0.473 & $+0.255$ \\
\bottomrule
\end{tabular}
\end{center}

A famous bridge is mentioned thousands of times, reinforcing its name; its opening year appears in only a small fraction of those mentions. Prominence amplifies name knowledge proportionally but amplifies temporal knowledge only weakly---consistent with the hypothesis that each \emph{specific} fact must clear its own mention-frequency bar rather than being carried along by its entity's overall visibility. Founding years are also uniquely hallucination-prone: wrong-answer rates reach 37--47\% on year probes (versus ${<}1\%$ on capitals), because any plausible 4-digit number is an attractor for a model that does not refuse.

\textbf{Domain-specific mention multipliers.} Controlling for sitelinks, some domains are systematically easier than entity prominence would predict, and others are systematically harder:

\begin{center}
\begin{tabular}{lrr}
\toprule
Domain & $n$ & Residual (easier $+$, harder $-$) \\
\midrule
Journal founding years        & 40 & $+0.201$ \\
Journals (names)              & 39 & $+0.215$ \\
University founding years     & 60 & $+0.117$ \\
Universities (names)          & 36 & $+0.120$ \\
Museum founding years         & 37 & $-0.102$ \\
Museums (names)               & 26 & $-0.118$ \\
Sports-club founding years    & 37 & $-0.144$ \\
Bridge opening years          & 20 & $-0.219$ \\
Place founding years          & 44 & $-0.310$ \\
\bottomrule
\end{tabular}
\end{center}

Journal founding years are +0.22 easier than sitelinks predict because every citation and every bibliography entry for an article in that journal implicitly states the journal's name; publication-year metadata is correspondingly dense. Place founding years are $-0.31$ harder because a municipality's founding date is typically buried in a single ``History'' section of its Wikipedia page, while the municipality is otherwise mentioned in contexts (weather, sports, transit) that do not carry the year. The \emph{structure of web discourse around each fact type} determines mention frequency far more than the prominence of the underlying entity.

\textbf{A documentation sweet spot.} Entities founded in 1900--1950 are best-known overall (mean recognition 0.40), outperforming both older entities ($<$1800: 0.25) that have fewer sitelinks and digital records, and younger entities (2000+: 0.26) that have not yet accumulated comprehensive historical coverage. This is consistent with a training-corpus curve that peaks where sustained institutional documentation (Wikipedia, newspaper archives, bibliographic databases) and historical notability both apply.

\subsubsection{Synthesis: Effective Mention Frequency}

Researcher probes and factual probes are two paths to the same conclusion. LLM knowledge is determined not by abstract prominence but by \emph{effective mention frequency}: the number of training-corpus documents that state the specific target fact in retrievable form, attributed to the specific name or entity being queried. Citation count, sitelink count, and h-index are partial proxies; pageviews, practitioner adoption, named-artifact count, and the density of derivative-document mentions are better ones.

This has two practical corollaries. For \emph{researchers}, the fraction of one's work that foundational models have internalized is not a rank-ordered function of citations. It is closer to a function of how much downstream text---tutorials, course materials, model cards, blog posts, GitHub documentation, news coverage---mentions the work by name. The strongest correlate of being memorized by frontier models is having named artifacts whose names travel widely. For \emph{facts about entities}, the query-design lesson is that any structured, bibliographically-enforced metadata (publication years, DOIs, institutional affiliations encoded in citation strings) is far better memorized than the same metadata when it lives only on the entity's own page. IKP difficulty tiers implicitly exploit this: the hard-tier probes are precisely those whose specific facts clear the mention-frequency bar only for the largest models.

\subsection{Key Empirical Findings}
\label{sec:stylized}

The evaluation yields a handful of empirical regularities we state crisply for reference:

\begin{itemize}[leftmargin=*]
    \item \textbf{Knowledge scales log-linearly.} Aggregate accuracy grows by ${\sim}15.9$ percentage points per $10\times$ increase in parameters, across four orders of magnitude (135M--1{,}600B).
    \item \textbf{Total parameters, not active, predict MoE knowledge.} $R^2 = 0.67$ versus $0.41$, consistent with factual storage being distributed across experts rather than localized to those activated per token.
    \item \textbf{Thinking mode is a flat ${\sim}2$ pp bonus.} The benefit peaks at T3--T4 and vanishes at T7: chain-of-thought aids retrieval, not storage.
    \item \textbf{No tier is a permanent ceiling.} T7 is the least-saturated tier (median ${\sim}8\%$, frontier ${\sim}22$--$28\%$); the penalized-era claim that all models score $0\%$ on T7 was a scoring artifact.
    \item \textbf{Safety tuning hides measurable knowledge.} Within the Claude Sonnet line, Sonnet 4 reads $21.5$ pp lower than Claude 3.7 Sonnet on IKP, while its T5 refusal rate jumps from $15\%$ to $85\%$---an artifact of refusal policy, not capacity.
    \item \textbf{Hallucination rate is a vendor fingerprint.} Google's smaller Gemma models hallucinate on 89--97\% of unknown probes at T5--T7; Anthropic Claude on 3--28\%. This wrong-versus-refuse split is a stable vendor signature.
    \item \textbf{Hallucination similarity exposes lineage.} Comparing models on \emph{which wrong answer they give} on rare facts cleanly separates shared-base pairs ($\mathrm{HSS} \geq 0.30$) from independent retrains ($\mathrm{HSS} < 0.10$), even when raw Jaccard overlap is high. Several nominally ``\texttt{.x}'' point releases (GPT-5 $\to$ 5.1, Opus 4.6 $\to$ 4.7, every cross-generation Gemini pair) fall in the retrained regime, while others (Opus 4 $\to$ 4.1, DeepSeek V3 $\to$ V3.1) are clear lineages.
\end{itemize}

\section{Discussion}
\label{sec:discussion}

\subsection{The Silent Tax of Safety Tuning}
\label{sec:silent-tax}

Heavily safety-tuned models~\citep{ouyang2022training, christiano2017deep, bai2022constitutional} refuse probes that earlier, comparable models answer, which can produce systematic capacity underestimates. Within the Claude Sonnet line, Sonnet 4 scores 51.1\% while Claude 3.7 Sonnet scores 72.6\%. On T5, Sonnet 4 refuses 85\% of probes; Sonnet 3.7 refuses only 15\%, bluffing instead. If Sonnet 4 knows at least as much, most of the $21.5$ pp gap is refusal policy, not capacity.

This has two implications beyond IKP methodology. First, refusal is now a dominant failure mode at the frontier: the same prompts that elicit correct answers from earlier generations produce refusals in later ones, with between-generation refusal gaps reaching tens of percentage points. Whether the refusing model still \emph{knows} these facts is not directly measurable; the earlier generation's answers make it plausible but do not prove it. Second, under no-penalty scoring ($\lambda=0$: a refusal and a confidently wrong answer both score $0$), a refusal is scored identically to genuine ignorance, so IKP \emph{may} under-count refusal-heavy models. Their estimates should therefore be read as lower bounds to the extent that refusal masks retained knowledge; a mild penalty (Appendix~\ref{app:lambda}) raises them.

\subsection{Hallucination Rate as a Vendor Signature}

When models encounter probes beyond their knowledge (T5--T7), the ratio of wrong answers to non-correct responses (wrong / (wrong + refusal)) is a characteristic vendor fingerprint. Google's smaller Gemma models hallucinate at 89--97\%---they almost never refuse---while larger Gemma and Gemini frontier models span 10--90\% (Gemini 3.1 Pro as low as 10\%, Gemini 3 Flash at 35\%). Anthropic Claude averages 11\%, with thinking variants as low as 3\% and non-thinking frontier reaching 28\%. The most striking within-vendor variation appears at OpenAI: GPT-4.1 and GPT-4.1-mini hallucinate at 98--99\%, while GPT-5 nano/mini models drop to 3--4\%---a generational shift in safety calibration that makes the wrong-versus-refuse split a clean vendor-generational signature (Section~\ref{sec:silent-tax}).

\subsection{Knowledge Regression Between Generations}

Multiple model families show newer models knowing \emph{less} than their predecessors on IKP. Two mechanisms account for this: (1)~RLHF conservatism~\citep{ouyang2022training} increasing between generations, so the newer model likely knows the facts but refuses to state them under the IKP prompt (Section~\ref{sec:silent-tax}); (2)~apparent knowledge loss when a smaller model succeeds a larger one in the same naming family (GPT-3.5 Turbo~\citep{brown2020language} outperforms GPT-4o-mini, $62.2\%$ vs $55.4\%$), which we cannot cleanly separate from the size difference itself. Later generations partially reverse these regressions (Claude 4.5/4.6 recover, DeepSeek v3.2 recovers from v3.1's drop).

\subsection{Robustness to Benchmark Saturation}
\label{sec:robustness-saturation}

A common pattern in LLM evaluation is that any benchmark, once influential, eventually saturates: MMLU~\citep{hendrycks2021measuring}, GSM8K~\citep{cobbe2021gsm8k}, SimpleQA~\citep{wei2024simpleqa}, and most recently Humanity's Last Exam~\citep{phan2025hle} are all approaching ceiling on frontier models, in part because the underlying tasks are compressible---once a model has internalized modular arithmetic or the scientific-reasoning template, it solves every instance of that task---and in part because high-profile benchmarks inevitably enter the training pipeline. IKP is structurally resistant to both failure modes. The probes measure stored facts, not procedures, so a model that ``saturates'' a tier has by construction memorized the long-tail facts that tier was sampling---which is precisely the quantity IKP exists to estimate. Direct contamination of the released probe file is a separate concern, but the pipeline (Section~\ref{sec:method}) is itself the benchmark: Wikidata sitelink-stratified sampling, DBLP citation-stratified sampling, and tier calibration against the landmark ladder regenerate a fresh probe set from disjoint entities in a few hours of compute.

\subsection{Limitations}
\label{sec:limitations}

\textbf{Training-data and post-training variance.} The calibration assumes approximately similar training signals across models. Vendor-specific choices in pretraining-data curation, RLHF refusal calibration, and post-training fine-tuning produce within-family scatter that the calibration absorbs as residual rather than slope. Heavily safety-tuned proprietary models (e.g., Anthropic's Haiku line, the GPT-5 nano/mini variants) systematically read smaller than their underlying weights: under $\lambda=0$ a refused-but-known fact scores as ignorance, so these estimates are lower bounds (Section~\ref{sec:silent-tax}).

\textbf{Prediction interval width.} The 90\% prediction interval factor is ${\sim}3.20\times$ in either direction, with $72\%$ of LOO folds within $2\times$ and $86\%$ within $3\times$ (Section~\ref{sec:loo}). This width is inherent to the single-variable approach and comparable to inference-economics estimates~\citep{epochai2024}. Combining IKP with inference-economics constraints could narrow the interval.

\textbf{Calibration sparsity above 1T parameters.} The largest open-weight calibration anchors are DeepSeek V4 Pro (1{,}600B) and Kimi K2.5/K2.6 ($\sim$1{,}000B). With only two calibration points above 1T, the calibration curve is effectively extrapolated beyond this range, and the slope at the high end is determined by very few data points. Consequently, parameter estimates for proprietary frontier models whose effective capacity falls above ${\sim}1$T should be read with wider uncertainty than the global $3.20\times$ prediction interval factor suggests. We expect this limitation to be substantially mitigated as additional open-weight models above 1T parameters are released.

\textbf{Retrieval augmentation confound.} A model using retrieval augmentation~\citep{lewis2020retrieval} could achieve high scores without parametric storage. That no model exceeds ${\sim}28\%$ on T7 suggests RAG is not currently deployed for factual queries of this type.

\textbf{Landmark circularity.} All six landmark models have scores inflated at their defining tier boundary by construction, since probe assignment is conditioned on landmark correctness. The effect is strongest for L6 (Gemini 3.1 Pro), whose 90\% T6 accuracy is an artifact. The five open-weight landmarks (L1--L5) also participate in the calibration set, potentially biasing the regression. The LOO-CV analysis (Section~\ref{sec:loo}) partially addresses this: when each landmark is held out, its prediction uses a regression fit without it, and the resulting LOO-CV $R^2 = 0.907$ confirms the fit is not driven by landmarks alone. Nevertheless, \emph{landmark models should be excluded from estimation targets}.

\textbf{Non-deterministic API responses.} We query all models at temperature${}=0$, but many API providers do not guarantee deterministic outputs at this setting (due to batching, quantization, or speculative decoding). We did not perform repeated evaluations to quantify this noise source, though the large probe count (1{,}400) mitigates its effect on aggregate scores.

\textbf{Probe contamination.} If the probe set leaks into training data, accuracy inflates. The probe set should remain private; we release only the methodology and evaluation toolkit. This creates a tension with reproducibility: independent researchers cannot replicate the calibration without the probes. We mitigate this by describing the probe generation procedure in sufficient detail to construct equivalent probe sets, and by releasing the scoring and estimation code.

\textbf{Probe allocation.} The uniform allocation of 200 probes per tier is suboptimal: T1--T2 saturate early and T7 discriminates only the top few models, while T3--T5 carry most of the discriminative power. Future versions could allocate more probes to the informative tiers.

\textbf{T7 probe quality.} Nine T7 probes have $>25\%$ correct rate across 201 models, indicating they are miscategorized and should be reassigned to T5 or T6 in future versions. The most extreme case (Ivan Beschastnikh, 64.0\% correct) is clearly not T7-difficulty knowledge.

\subsection{Wikidata Long-Tail Data Quality}
\label{sec:wikidata-audit}

After an initial evaluation pass, we audited every T5--T7 Wikidata-sourced probe (557 probes) for ground-truth quality, web-verifiable correctness, and question-form ambiguity. The audit revealed three classes of issues that motivated the released probe set: (i)~Wikidata-itself-wrong facts (e.g., Makran Medical College's founding year listed as 2015 but Wikipedia says 2017; ``L'Homme qui marche I'' attributed to the bronze foundry Susse Fr\`{e}res rather than Giacometti; Roku still listed as headquartered in Los Gatos despite a 2019 move to San Jose); (ii)~probe-construction ambiguity, where Wikidata's specific entity is correct but the question is unanswerable without disambiguation (``In what year was Putnam founded?''---which Putnam?); and (iii)~politically contested attributions where Wikidata picks one answer that reasonable models will dispute (Cape Plaka in Crimea, Loaita Cay in the Spratlys).

\textbf{Per-fact-type Wikidata reliability at the long tail} (verified empirically across 10 sourcing rounds and ${\sim}120$ web cross-checks):

\begin{center}
\small
\begin{tabular}{lr}
\toprule
\textbf{Fact type (Wikidata property)} & \textbf{Web pass rate} \\
\midrule
\texttt{P403} river mouth                          & 100\% \\
\texttt{P57} director (film)                       & 67--100\% \\
\texttt{P17} country (cape/island/lake/mountain)   & 75--100\% \\
\texttt{P571} inception (founding year)            & ${\sim}95\%$ \\
\texttt{P112} founder                              & 75\% (sometimes returns the founding \emph{organization} not person) \\
\texttt{P170} creator (sculpture)                  & ${\sim}70\%$ (often returns the bronze \emph{foundry} not sculptor) \\
\texttt{P58} screenwriter                          & 50\% (conflates story with screenplay credit) \\
\texttt{P159} headquarters                         & 25\% (frequently stale: Roku, Vista Outdoor) \\
\texttt{P170} creator (painting)                   & ${\sim}70\%$ on attribution, but ${\sim}100\%$ title-collision \\
                                                   & at obscurity (``Madonna and Child'') \\
\texttt{P138} named after                          & mixed (self-referential errors observed) \\
\bottomrule
\end{tabular}
\end{center}

The audit applied 119 changes to the probe set: 4 fact corrections, 64 question rewrites adding country/admin grounding, and 51 drop-and-replace pairs sourced from new diverse fact types. T6/T7 Wikidata-source fact-type diversity went from a 100\% founding-year monoculture before the audit to 16 fact types after (river mouths, film directors, capes, sculptors, founders, mountains, lakes, painters, screenwriters, etc.). \emph{All probes with corrected answers were re-calibrated through the landmark ladder}; all 5 corrected probes shifted from T7 down to T5 after correction (models that previously gave the ``wrong'' answer were now judged correct), illustrating that gold-truth correctness drives tier assignment, so any answer-field edit invalidates the prior calibration. Full audit telemetry, the failure-mode taxonomy, and the changeset are in Appendix~\ref{app:wikidata-audit}.

\subsection{Open Questions}
\label{sec:open-questions}

Our results suggest a set of falsifiable questions that incompressibility-grounded probes can answer as more models and more ground-truth anchors become available:

\begin{enumerate}[leftmargin=*]
    \item \textbf{When does T7 saturate?} The best models answer only ${\sim}25$--$28\%$ of T7. If factual capacity keeps scaling log-linearly, T7 accuracy should continue to climb; tracking it as new models release would locate the next plateau and test the incompressibility argument.
    \item \textbf{How much within-family scatter is post-training rather than parameter count?} Proprietary ``Flash''-class variants typically read close to their Pro-class siblings on IKP despite presumed parameter-count differences. Extending the analysis to additional same-vendor pairs with disclosed sizes would partition the scatter into pretraining-data, RLHF, and architecture contributions.
    \item \textbf{Can RLHF-hidden knowledge be recovered?} The ``Opus 4.1 knows but refuses'' pattern implies an alignment tax measurable in percentage points of factual recall. Prompting strategies, deprobing, or activation-steering methods that recover refused-but-known answers would place an upper bound on how much capacity safety tuning obscures.
    \item \textbf{Do knowledge fingerprints persist through continued pretraining?} Our fingerprints survive fine-tuning and distillation. If they also survive substantial continued pretraining on fresh data, they become a practical training-free provenance tool for open-weight licensing enforcement.
\end{enumerate}

\section{Conclusion}
\label{sec:conclusion}

We began from the observation that frontier language models have quietly absorbed a large fraction of the open expert discourse of the past two decades---individual researchers, their named artifacts, founding years of institutions, and thousands of facts whose ``incompressible'' nature makes them impossible to reconstruct from language alone. Incompressible Knowledge Probes turn this observation into a measurement instrument: a principled method for estimating LLM effective knowledge capacity from black-box API access, and for characterizing what kinds of people, artifacts, and facts a model has actually internalized. The approach achieves $R^2 = 0.910$ across 93 open models from 19 vendors spanning four orders of magnitude (135M to 1{,}600B), under no-penalty scoring and validated by leave-one-out cross-validation (median fold $1.48\times$; $72\%$ of models predicted within $2\times$, $86\%$ within $3\times$). We demonstrate that total parameters---not active parameters---predict MoE knowledge capacity ($R^2 = 0.67$ vs $0.41$), provide effective-capacity estimates for 97 proprietary models, and show that within-family scatter (vendor-specific RLHF, training-data curation, post-training format) is sufficient to explain the distance between same-vendor variants without invoking a separate scaling regime.

The broader claim is conceptual. The widely-reported saturation of reasoning benchmarks has been read as evidence that scaling is winding down. Our results suggest this reading confuses two distinct resources: procedural capability, which compresses under the Densing Law, and factual storage, which does not compress under pretraining. Measuring only the incompressible component, we find that parameter scaling continues across three orders of magnitude with no detectable time effect at fixed parameter count. How much of any given researcher's contributions will have flowed into the next generation of frontier models is now an empirical question with a measurement instrument.

We release the IKP evaluation toolkit, probe set, raw per-model responses, fingerprinting scripts, and all data needed to reproduce every figure and table in this paper. Code: \url{https://github.com/19PINE-AI/ikp}. Interactive companion site (calibration curves, per-model knowledge profiles, per-probe responses, fingerprint heatmaps, densing-law analysis): \url{https://01.me/research/ikp}.

\section*{Acknowledgements}

The author thanks Jiyan He and Zihan Zheng for early discussions and observations that seeded this work, including the inspiration drawn from three years of factual probes against newly released models---most notably the USTC Hackergame stress test that opens this paper, where the same prompt asked at each new release traced a single fact arriving in parameters over time. The framing of factual capacity as the incompressible component of model knowledge owes much to those conversations. 

We thank Sturb and LawrenceC, whose independent sanity-check of an earlier version of this work~\citep{sturb2026sanity} identified a discrepancy between the described and implemented per-tier scoring and the sensitivity of frontier estimates to the hallucination penalty. This revision responds directly to those points: it adopts no-penalty scoring ($\lambda = 0$), which removes both the penalty and the flooring degree of freedom, reports the full $\lambda \times$ flooring ablation and prediction bands, and cleans the probe set of name-collision and label-ambiguity defects (Appendix~\ref{app:lambda}).

This paper was produced using Pine Copilot's voice-directed \emph{whisper coding} workflow~\citep{pineai2026whispercoding}, in which the author specifies, discusses, and reviews the work by voice while a coding agent---Claude Code with Claude Opus 4.8---carries out the planning, coding, experiments, and paper writing.

\bibliographystyle{plainnat}
\bibliography{references}

\clearpage
\appendix
\input{appendix}

\end{document}

%% file: tables/frontier_estimates.tex
\begin{tabular}{llrrr}
\toprule
Model & Vendor & Accuracy & Est.\ Size & 90\% PI \\
\midrule
GPT-5.5 Pro (think) & OpenAI & 83.4\% & ${\sim}5.3$T & [1.7T--17.1T] \\
GPT-5.5 & OpenAI & 82.6\% & ${\sim}4.7$T & [1.5T--15.1T] \\
Claude Fable 5 & Anthropic & 80.5\% & ${\sim}3.5$T & [1.1T--11.2T] \\
Gemini 2.5 Pro & Google & 79.5\% & ${\sim}3.0$T & [943B--9.6T] \\
GPT-5.4 Pro & OpenAI & 77.8\% & ${\sim}2.4$T & [744B--7.6T] \\
GPT-4.1 & OpenAI & 77.2\% & ${\sim}2.2$T & [677B--6.9T] \\
GPT-5 Pro & OpenAI & 77.0\% & ${\sim}2.1$T & [657B--6.7T] \\
o3 & OpenAI & 76.9\% & ${\sim}2.1$T & [650B--6.7T] \\
Grok-3 & xAI & 76.8\% & ${\sim}2.1$T & [642B--6.6T] \\
Grok-4 & xAI & 76.6\% & ${\sim}2.0$T & [619B--6.3T] \\
GPT-5 (think) & OpenAI & 76.5\% & ${\sim}2.0$T & [614B--6.3T] \\
Claude Opus 4.6 (think) & Anthropic & 75.1\% & ${\sim}1.6$T & [503B--5.1T] \\
MiMo v2.5 Pro & Xiaomi & 74.4\% & ${\sim}1.4$T & [451B--4.6T] \\
Hy3 Preview & Tencent & 74.1\% & ${\sim}1.4$T & [433B--4.4T] \\
GPT-5.4 & OpenAI & 72.3\% & ${\sim}1.1$T & [334B--3.4T] \\
GPT-5.3 & OpenAI & 72.2\% & ${\sim}1.1$T & [330B--3.4T] \\
GLM-5.2 & Z.ai & 72.2\% & ${\sim}1.0$T & [326B--3.3T] \\
GLM-5 Turbo & Z.ai & 71.1\% & ${\sim}896$B & [280B--2.9T] \\
Claude Opus 4.5 & Anthropic & 71.1\% & ${\sim}892$B & [279B--2.9T] \\
Qwen3.7 Max & Alibaba & 71.0\% & ${\sim}887$B & [277B--2.8T] \\
Claude Opus 4.1 (think) & Anthropic & 70.7\% & ${\sim}847$B & [265B--2.7T] \\
Claude Sonnet 4.6 (think) & Anthropic & 70.0\% & ${\sim}766$B & [240B--2.5T] \\
Claude Opus 4.7 (think) & Anthropic & 69.8\% & ${\sim}738$B & [231B--2.4T] \\
MiMo v2.5 & Xiaomi & 69.5\% & ${\sim}705$B & [220B--2.3T] \\
Grok-4.20 & xAI & 69.3\% & ${\sim}689$B & [215B--2.2T] \\
GPT-5.2 Pro & OpenAI & 68.8\% & ${\sim}639$B & [200B--2.0T] \\
GPT-5.1 & OpenAI & 68.7\% & ${\sim}634$B & [198B--2.0T] \\
Qwen3 Max & Alibaba & 68.4\% & ${\sim}606$B & [189B--1.9T] \\
Claude Sonnet 5 & Anthropic & 68.2\% & ${\sim}584$B & [183B--1.9T] \\
Qwen3.7 Plus & Alibaba & 68.1\% & ${\sim}579$B & [181B--1.9T] \\
Step-3.7 Flash & StepFun & 66.7\% & ${\sim}471$B & [147B--1.5T] \\
MiniMax M3 & MiniMax & 63.9\% & ${\sim}313$B & [98B--1.0T] \\
Claude Opus 4.8 & Anthropic & 61.9\% & ${\sim}236$B & [74B--753B] \\
\bottomrule
\end{tabular}

%% file: appendix.tex

\section{Probe Generation Methodology}
\label{app:probes}

This appendix expands Section~\ref{sec:probe-generation} with the full reproducibility detail of the IKP probe pipeline: tier definitions, per-corpus sampling procedures, quality filters with their drop rates, the final per-tier composition, verification procedure, and sample probes.

\subsection{Tier Definitions}
\label{app:tier-defs}

Each of the seven IKP tiers is defined by a calibrated parameter range: the model size at which a model with typical training-data coverage starts answering reliably. Tier targets are then \emph{validated} against the landmark ladder (Section~\ref{sec:tiers}); probes whose landmark behavior contradicts the assigned tier are reassigned or dropped.

\begin{table}[H]
\centering
\small
\setlength{\tabcolsep}{4pt}
\begin{tabularx}{\linewidth}{@{}l l l X@{}}
\toprule
\textbf{Tier} & \textbf{Name} & \textbf{Param range} & \textbf{Example} \\
\midrule
T1 & Universal Knowledge & 0.1B--0.5B & ``What is the capital of Norway?'' (Oslo) \\
T2 & Common Reference Knowledge & 0.5B--7B & ``Who composed the Enigma Variations?'' (Edward Elgar) \\
T3 & Domain-Specific Knowledge & 7B--32B & ``In what year was the Battle of Hastings fought?'' (1066) \\
T4 & Obscure Knowledge & 32B--235B & ``In computer science, what is the research subfield of Peter Druschel?'' (computer networking) \\
T5 & Deep Knowledge & 235B--1T & ``In what year was the town of Eliot in York County, Maine founded?'' (1810) \\
T6 & Long-Tail Knowledge & 1T--10T & ``In computer science, what is the research subfield of Jeffrey Helt?'' (distributed systems) \\
T7 & Extreme Long-Tail & $>$10T & ``In what country is the mountain Tadekho Hill located?'' (Canada) \\
\bottomrule
\end{tabularx}
\caption{Tier definitions used during probe generation. Parameter range is the model size at which a typically-trained model begins answering reliably; T7 is by construction beyond all current models and used as a ceiling probe. Examples here match those in Appendix~\ref{app:sample-probes}.}
\label{tab:tier-defs}
\end{table}

\subsection{Phase A: LLM-Generated Candidates}

Phase A targets T1--T2 saturation and supplementary fill at T3--T4. We use GPT-5 as the candidate generator with a structured prompt that:
(i)~specifies the target tier's parameter range and exemplar Q/A pairs,
(ii)~rotates a primary region (8 regions: North America, Europe, East Asia, South Asia, Middle East, Africa, Latin America, Oceania) and primary domain (6 domains: people, places, publications, measurements, events, organizations) per generation batch,
(iii)~requests 50 candidates per batch with $\geq 3$ paraphrases each, and
(iv)~requires verifiable, single-token-or-short-string gold answers.

\textbf{Yield.} 401 of 1{,}400 final probes ($28.6\%$) come from Phase A: T1 $n{=}166$, T2 $n{=}152$, T3 $n{=}51$, T4 $n{=}32$. The empirical tier distribution of LLM-generated candidates is heavily skewed toward T1--T2 (${\sim}82\%$) regardless of difficulty prompting, confirming that an LLM is structurally unable to generate probes beyond its own knowledge horizon---the central reason Phase B exists.

\subsection{Phase B: Corpus-Grounded Probes}

Phase B grounds probes in two external corpora whose ground-truth answers can be verified independently of any LLM.

\paragraph{Wikidata-grounded probes (557 probes).} We use Wikidata~\citep{vrandecic2014wikidata} SPARQL queries to sample entities from six categories: universities, journals, museums, bridges, sports clubs, and geographic places. Per-category SPARQL queries filter on entity type (P31 / instance of), require a published founding year (P571) or equivalent property, and stratify by Wikipedia article monthly-views-as-of-2025 to target a specific tier. Gold answers are read directly from the queried Wikidata property value. T3--T7 receive 94, 111, 100, 141, 100 Wikidata probes respectively.

\paragraph{Researcher subfield probes (345 probes).} Computer-science researchers are sampled from DBLP and OpenAlex, stratified by citation-count buckets (10--50 citations for T7, 50--500 for T6, 500--5{,}000 for T5, 5{,}000--50{,}000 for T3/T4). Each researcher's primary subfield is initially assigned from the dominant DBLP venue tag via a venue $\to$ subfield mapping table (e.g., SOSP/OSDI $\to$ ``operating systems''; SIGCOMM $\to$ ``computer networking''; CRYPTO $\to$ ``computer security''; SIGMOD/VLDB $\to$ ``database systems''; STOC/FOCS $\to$ ``theoretical computer science''). T5--T7 each receive 100, 59, 100 researcher probes; T3--T4 receive 35 and 51.

\textbf{Probe template.} The released format is two-part:
\begin{quote}
\itshape In computer science, what is the research subfield of \emph{Name}, and name one paper, system, institution, or co-author associated with their work? If you don't know who this person is, say so.
\end{quote}
The CS scoping (``In computer science'') reduces cross-field name collisions for short / common names. The artifact requirement (``name one paper, system, institution, or co-author'') is the load-bearing change: it forces models to produce a verifiable token that can be cross-checked against ground truth, rather than emitting a plausible-sounding but unattested CS subfield label. Without this requirement, a model that has never encountered ``Dan Schatzberg'' can still get full credit by guessing ``computer architecture'' (a popular subfield), since OpenAlex tags him as such; with the requirement, the model must additionally name TMO, EbbRT, ASPLOS, ISCA, Boston University, Meta, or some other token in his actual research footprint.

\textbf{Per-researcher evidence bundle.} Each researcher carries a structured gold record built from their OpenAlex profile and, for the tail of name-collision-flagged researchers, manually web-verified:
\begin{itemize}[leftmargin=*,topsep=2pt,itemsep=1pt]
    \item \texttt{primary\_subfield}: one of a controlled set of CS subfield labels.
    \item \texttt{secondary\_subfields}: 0--3 acceptable adjacent labels for researchers whose work spans multiple subfields.
    \item \texttt{affiliations}: top 1--2 last-known institutions.
    \item \texttt{named\_systems}: short, capitalized system / artifact tokens extracted from paper titles by a regex pass over the top 10 most-cited works (e.g., ``TMO'', ``Zoltan'', ``Wukong'').
    \item \texttt{venues}: top 5--6 publication venues from the same set.
    \item \texttt{co\_authors}: top 5 collaborators ranked by joint-paper count.
    \item \texttt{top\_works}: top 5 paper titles with publication years and citation counts.
\end{itemize}

\textbf{Four-way evidence-aware judge.} Responses are classified into four classes, scored at $\lambda=-1$ for WRONG:
\begin{itemize}[leftmargin=*,topsep=2pt,itemsep=1pt]
    \item \textbf{CORRECT-STRONG} (score $+1.0$): response names \texttt{primary\_subfield} (or a direct synonym, or a listed \texttt{secondary\_subfield}) \emph{and} cites at least one matching evidence item---a system from \texttt{named\_systems}, a venue from \texttt{venues}, an affiliation from \texttt{affiliations}, a co-author from \texttt{co\_authors}, or a paper-title fragment overlapping with \texttt{top\_works}.
    \item \textbf{CORRECT-WEAK} (score $+0.5$): response names a primary or secondary subfield but cites no specific evidence (e.g., ``computer architecture, focusing on memory systems and cache coherence'' for a researcher with no actual cache work).
    \item \textbf{REFUSAL} (score $0$): response says ``I don't know'', expresses uncertainty, or declines.
    \item \textbf{WRONG} (score $\lambda=-1$): response names a CS subfield outside \{primary, secondary\} \emph{or} confidently fabricates specifics that contradict / are absent from the evidence bundle.
\end{itemize}
The score formula $\lambda \cdot \mathbb{1}[\text{WRONG}] + 0.5 \cdot \mathbb{1}[\text{WEAK}] + 1.0 \cdot \mathbb{1}[\text{STRONG}]$ makes confident bluffing strictly worse than refusing, while still rewarding partial knowledge (WEAK) above refusal.

\textbf{Name-collision sub-class.} Sixteen names (all in T7) had no real CS researcher despite OpenAlex listing a person under that name (the OpenAlex profile pointed at, e.g., a chemist, virologist, or radiologist with the same name). These were replaced with web-verified CS researchers in the same subfield, re-calibrated through the 6-landmark ladder, and re-confirmed as T7 (no landmark answers correctly).

\paragraph{Manual / supplementary probes (97 probes).} A small set of high-quality probes were manually authored or carried over from earlier dataset versions to balance per-tier coverage in T1--T4 where automated sampling was insufficient.

\subsection{Quality Filters and Drop Rates}

Four filters are applied to every candidate before inclusion. Each filter is conservative (false positives drop legitimate probes; false negatives admit problematic ones, which we accept as long as the rate is bounded):

\begin{enumerate}[leftmargin=*,topsep=2pt,itemsep=1pt]
    \item \textbf{Computable knowledge filter.} Drops probes whose answer can be derived by rule rather than memorization: arithmetic on years (``what year was X 10 years ago''), alphabetical/ordinal answers, IUPAC chemical naming, etc. Implementation: regex + LLM classification, manual review of edge cases.
    \item \textbf{Monotonicity filter.} A probe is admitted only if accuracy across the landmark ladder (Section~\ref{sec:tiers}) is monotonic non-decreasing in landmark capability. Probes where a smaller landmark answers correctly while a larger one does not---typically caused by ambiguous phrasing, multi-valued ground truth, or judge errors---are dropped. Drops $\sim 15\%$ of post-Phase-B candidates.
    \item \textbf{Name-collision filter (researcher probes).} Drops two-character Chinese names and single-initial given names where multiple distinct researchers share the identifier; a 50-citation Liu Yang and a 5{,}000-citation Liu Yang are not the same person. Implementation: cross-check unique \emph{(name, affiliation)} tuples against DBLP author IDs.
    \item \textbf{Contamination filter (researcher probes).} Drops researchers whose primary subfield is machine learning, deep learning, or reinforcement learning. These authors' own work disproportionately appears in frontier-model training corpora, inflating accuracy in a way that conflates IKP-style long-tail knowledge with training-set memorization of one's own field. The 14 retained CS subfields (systems, databases, security, networking, theory, IR, NLP, vision, etc.) are those in Table~\ref{tab:recog-by-field}.
    \item \textbf{Question-template grounding filter.} For Wikidata-sourced probes, every question template embeds a discriminating attribute---year, country, administrative region, genre, language, or publisher---extracted from a Wikidata field that is \emph{not} the answer field. This is the same design used by researcher probes (which prefix every question with ``In computer science''). The filter exists because an audit of the bare-name templates (``In what year was X founded?'') revealed silent failures under title collision: ``Putnam'' could refer to Putnam, Connecticut (Wikidata's intended entity) or Putnam County, Florida; ``Madonna and Child'' refers to dozens of Renaissance paintings. Grounding raises the within-probe identifiability bar without leaking the answer (e.g., ``Who painted the 1522 painting `The Virgin of Carmel'?'' grounds via year without naming the artist). The grounded template is generated programmatically from Wikidata claims at SPARQL time, then independently re-calibrated through the landmark ladder---a non-trivial post-condition, since grounding tends to make probes easier and shift them out of T6/T7. In our audit, $\sim 30\%$ of grounded T7 candidates re-calibrated to T5 (became too easy with grounding) and ${\sim}5\%$ became non-monotonic; the surviving $\sim 65\%$ form the released T6/T7 Wikidata pool.
\end{enumerate}

\subsection{Final Composition}
\label{app:probe-composition}

\begin{table}[H]
\centering
\small
\begin{tabular}{l rrrrr}
\toprule
\textbf{Tier} & \textbf{LLM} & \textbf{Wikidata} & \textbf{Researcher} & \textbf{Manual} & \textbf{Total} \\
\midrule
T1 & 166 & 6   & 0   & 28 & 200 \\
T2 & 152 & 5   & 0   & 43 & 200 \\
T3 & 51  & 94  & 35  & 20 & 200 \\
T4 & 32  & 111 & 51  & 6  & 200 \\
T5 & 0   & 100 & 100 & 0  & 200 \\
T6 & 0   & 141 & 59  & 0  & 200 \\
T7 & 0   & 100 & 100 & 0  & 200 \\
\midrule
\textbf{Total} & 401 & 557 & 345 & 97 & \textbf{1{,}400} \\
\bottomrule
\end{tabular}
\caption{Per-tier probe composition by source. ``Manual'' aggregates manually-authored and supplementary probes from earlier dataset iterations. The dominance of researcher and Wikidata probes at T5--T7 reflects the unavoidability of corpus grounding at frontier difficulty.}
\label{tab:probe-composition}
\end{table}

For T5--T7 specifically, the dominant domain is computer-science researcher subfield (259/600 probes, $43\%$). The remaining T5--T7 Wikidata probes span 16 fact types after the audit (Section~\ref{sec:wikidata-audit}): grounded founding years (sports clubs, museums, bridges, universities, journals, abbeys, tramways), and entity-attribute questions (river mouths, film directors, capes, sculptors, founders, mountains, lakes, painters, screenwriters, composers, named-after eponyms, headquarters cities, architects). The original T6/T7 Wikidata composition was a 100\% founding-year monoculture; the audit replaced ${\sim}50\%$ of T6/T7 Wikidata probes with diverse fact types to remove that source bias while preserving the unambiguous-answer property.

\subsection{Verification Procedure}

Gold-answer verification differs by source:

\begin{itemize}[leftmargin=*,topsep=2pt,itemsep=2pt]
    \item \textbf{Wikidata probes}: gold answer is the queried property value as of the SPARQL extraction date. Conflicts (e.g., a museum with multiple founding-date claims) are resolved by selecting the value with the most reference citations; probes with unresolved conflicts are dropped. After an audit (Section~\ref{sec:wikidata-audit}), every T5--T7 Wikidata probe was independently web-cross-checked against Wikipedia and primary sources; probes with a Wikidata-vs-Wikipedia disagreement on the answer were either corrected (4 cases, e.g.\ Makran Medical College: Wikidata 2015, Wikipedia 2017---kept Wikipedia's answer) or dropped (e.g., probes attributing a sculpture to its bronze foundry rather than its sculptor). All probes whose answers were corrected were re-calibrated through the landmark ladder, since changing the gold answer changes which landmarks are scored as correct and therefore the assigned tier.
    \item \textbf{Researcher probes}: gold subfield is derived from each researcher's dominant DBLP venue tag in their highest-cited 5 papers, mapped to one of 14 CS subfields. Researchers whose top-5 papers split across $\geq 3$ unrelated subfields are dropped (only researchers with a clear primary specialization are retained).
    \item \textbf{Phase A probes}: gold answers are produced by the generator and independently re-verified by a different LLM (Claude Opus 4.6) plus a manual spot-check of $\sim 10\%$.
    \item \textbf{Manual probes}: directly authored with citation evidence.
\end{itemize}

\subsection{Sample Probes}
\label{app:sample-probes}

\begin{table}[htbp]
\centering
\small
\setlength{\tabcolsep}{4pt}
\renewcommand{\arraystretch}{1.15}
\begin{tabular}{@{}c l p{0.55\linewidth} l@{}}
\toprule
\textbf{Tier} & \textbf{Source} & \textbf{Question} & \textbf{Answer} \\
\midrule
\multirow{3}{*}{T1} & Wikidata   & What is the capital of Norway? & Oslo \\
                    & LLM        & In what year was the Meiji Constitution promulgated? & 1889 \\
                    & LLM        & Who designed the Sydney Opera House? & Jørn Utzon \\
\midrule
\multirow{3}{*}{T2} & LLM        & Who composed the Enigma Variations? & Edward Elgar \\
                    & LLM        & What is the most common isotope of hydrogen? & Protium \\
                    & Manual     & Who was the first Prime Minister of India? & Jawaharlal Nehru \\
\midrule
\multirow{3}{*}{T3} & LLM        & In what year was the Battle of Hastings fought? & 1066 \\
                    & Wikidata   & In what year was Hunan Normal University founded? & 1938 \\
                    & Manual     & What year was the Camp David Accords signed? & 1978 \\
\midrule
\multirow{3}{*}{T4} & Researcher & In computer science, what is the research subfield of Peter Druschel, and name one paper, system, institution, or co-author associated with their work? If you don't know who this person is, say so. & computer networking \\
                    & Wikidata   & In what year was Lawrence Heritage State Park founded? & 1980 \\
                    & Wikidata   & In what year was the journal \textit{Cancer} first published? & 1948 \\
\midrule
\multirow{3}{*}{T5} & Wikidata   & In what year was the town of Eliot in York County, Maine founded? & 1810 \\
                    & Researcher & In computer science, what is the research subfield of Jianliang Xu, and name one paper, system, institution, or co-author associated with their work? If you don't know who this person is, say so. & data mining \\
                    & Wikidata   & In what year was State Polytechnic of Indramayu founded? & 2014 \\
\midrule
\multirow{3}{*}{T6} & Researcher & In computer science, what is the research subfield of Jeffrey Helt, and name one paper, system, institution, or co-author associated with their work? If you don't know who this person is, say so. & distributed systems \\
                    & Wikidata   & In what year was Museu da Misericórdia founded? & 2006 \\
                    & Wikidata   & In what year was Universitas Imelda Medan founded? & 2019 \\
\midrule
\multirow{3}{*}{T7} & Wikidata   & In what country is the mountain Tadekho Hill located? & Canada \\
                    & Researcher & In computer science, what is the research subfield of Christopher Bengel, and name one paper, system, institution, or co-author associated with their work? If you don't know who this person is, say so. & computer architecture \\
                    & Wikidata   & Into what body of water does the Bolshaya Khopa in Arkhangelsk Oblast, Russia flow? & Soyana \\
\bottomrule
\end{tabular}
\caption{Three sample probes per tier, one per source where possible. Each row is a single probe as it appears in the released probe set.}
\label{tab:sample-probes}
\end{table}

Table~\ref{tab:sample-probes} lists three representative probes per tier, drawing from all three probe sources (LLM-generated, Wikidata-grounded, researcher subfield) so that each contributes.

\section{Hallucination Penalty Sensitivity}
\label{app:lambda}

The hallucination penalty $\lambda$ (a wrong answer scores $\lambda$; correct $=+1$, refusal $=0$) is a scoring-convention choice. This appendix reports its sensitivity: it sweeps $\lambda$ over $\{0,\,-0.25,\,-0.5,\,-1.0,\,-2.0\}$, refits the log-linear calibration on the same 93-model open-weight set for each value, and re-inverts for four flagship proprietary models (Table~\ref{tab:lambda-sweep}). The released dataset, website, and toolkit default to $\lambda = 0$ (raw accuracy---directly the fraction of probes answered correctly); the table and figure let results under any penalty be read off directly.

\begin{table}[H]
\centering
\small
\caption{Sensitivity of the IKP calibration and of four flagship proprietary estimates to the hallucination penalty $\lambda$. Each row refits $A = \alpha \log_{10}(N) + \beta$ on the same 93 open-weight models, recomputing every model's accuracy from stored verdicts under the listed $\lambda$ (correct $=+1$, refusal $=0$, wrong $=\lambda$), and re-inverts the fit for each flagship model. ``Slope'' is in percentage points per decade; ``LOO$\times$'' is the median multiplicative parameter-prediction error in leave-one-out cross-validation; ``PI$\times$'' is the 90\% prediction-interval factor. $\lambda = 0$ (starred) is the default used by the released dataset, website, and toolkit.}
\label{tab:lambda-sweep}
{\setlength{\tabcolsep}{4pt}\input{tables/lambda_sensitivity}}
\end{table}

\begin{figure}[H]
\centering
\includegraphics[width=\textwidth]{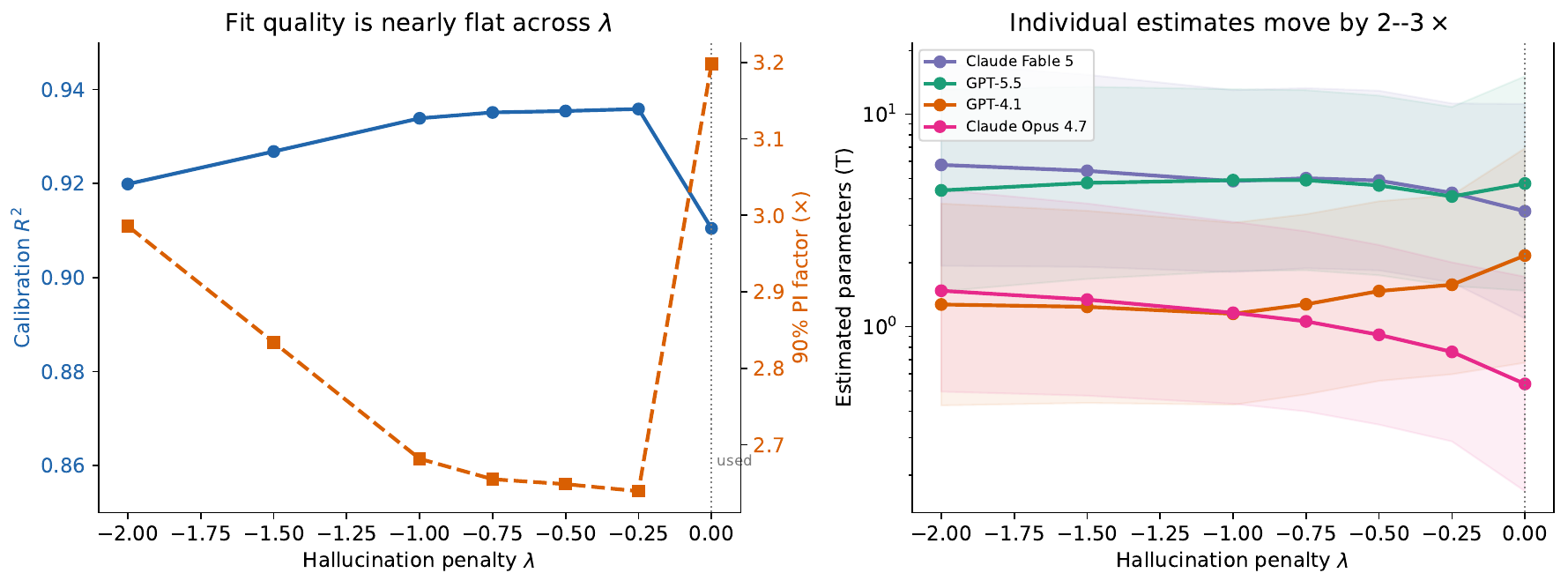}
\caption{What changes across the hallucination penalty $\lambda$. \textbf{Left:} calibration fit quality is nearly flat---$R^2$ stays in $[0.91,0.94]$ and the 90\% prediction-interval factor in $[2.6,3.2]\times$ (floored scoring for $\lambda<0$; $\lambda=0$ is floor-free). \textbf{Right:} individual frontier estimates (90\% bands shaded) move by $2$--$3\times$ and \emph{re-order} by vendor: as $\lambda\to 0$ the confident guesser GPT-4.1 rises while the refusal-heavy Claude Opus~4.7 falls.}
\label{fig:lambda}
\end{figure}

\textbf{Findings.} Fit quality is near-flat across the band---$R^2$ ranges only ${\sim}0.91$--$0.94$ and the 90\% PI factor ${\sim}2.6$--$3.2\times$ (Figure~\ref{fig:lambda}, left)---while individual estimates shift $2$--$3\times$ in a \emph{vendor-dependent} way (Figure~\ref{fig:lambda}, right): a heavier penalty raises models that refuse rather than bluff (Claude Opus~4.7, $0.55 \to 1.5$T from $\lambda = 0$ to $-2$) and lowers confident guessers (GPT-4.1, $2.2 \to 1.2$T), reattributing above-trend scores from hallucination to genuine capacity.

\textbf{The $\lambda$-choice principle.} We fix $\lambda$ by \emph{parsimony}, not by tuning: adopt the parameter-free operating point---$\lambda = 0$, which also eliminates the flooring rule (per-tier scores are then $\geq 0$ by construction)---\emph{unless} a nonzero penalty materially improves the calibration. It does not (the flat fit above; $R^2$ within $0.03$, ordering preserved). A nonzero $\lambda$ therefore buys at most $0.03$ in $R^2$ at the price of \emph{two} tunable knobs---the penalty magnitude and the floor---that, as Table~\ref{tab:floor-ablation} shows, can together be dialed to inflate frontier estimates by $3$--$5\times$. Because IKP is a measurement instrument, we prefer the estimator a reader cannot tune toward a desired answer, and $\lambda = 0$ is that Schelling point; readers who prefer a hallucination-aware score may read any $\lambda$ column directly, since no qualitative conclusion of the paper depends on the choice.

\subsection{Flooring and its interaction with $\lambda$}
\label{app:flooring}

A second scoring degree of freedom is whether per-tier scores are floored at zero. Under a negative penalty a strongly-bluffing model can score negative on the hard tiers; flooring those scores at zero raises small/aggressive models' accuracy, which flattens the calibration slope and---because the curve is inverted to estimate size---inflates frontier estimates. Table~\ref{tab:floor-ablation} crosses $\lambda$ with flooring on/off.

\begin{table}[H]
\centering
\small
\caption{Calibration and two flagship estimates under $\lambda \times$ flooring. Flooring only matters under a negative penalty; unfloored, the fit degrades sharply as $|\lambda|$ grows ($R^2$ from $0.91$ to $0.72$, PI factor from $3.2\times$ to $9.9\times$) because negative hard-tier scores dominate the regression, and estimates fall by up to ${\sim}3\times$. At $\lambda = 0$ (starred) the floor is a no-op, so the ambiguity vanishes entirely---the principal reason we adopt no-penalty scoring.}
\label{tab:floor-ablation}
{\setlength{\tabcolsep}{5pt}\input{tables/lambda_floor_ablation}}
\end{table}

\textbf{Relation to prior discrepancy.} An earlier version of IKP reported floored scores under $\lambda = -1.0$ while the accompanying text described the scores as unfloored~\citep{sturb2026sanity}; the two differ materially (e.g.\ GPT-5.5 $4.9$T floored vs.\ $1.5$T unfloored at $\lambda=-1.0$). Adopting $\lambda = 0$ removes the discrepancy at its root: with no penalty every per-tier score is $\text{correct}/\text{total} \geq 0$, so floored and unfloored scoring are identical and there is nothing left to specify.

\section{Dense vs.\ MoE Calibration}
\label{app:moe}

The calibration pools dense and Mixture-of-Experts models; Table~\ref{tab:moe-fits} fits them separately. The per-decade slope is nearly identical (dense $0.151$, MoE-total $0.145$), but MoE-total sits slightly higher---an MoE knows ${\sim}2$--$4$ pp more per \emph{total} parameter than a same-size dense model---and fits more loosely ($R^2=0.67$ vs $0.88$). Active parameters predict MoE knowledge far worse ($R^2=0.41$), confirming that factual capacity scales with \emph{total}, not active, weights: an MoE with $T$ total and $A$ active parameters behaves, for recall, like a dense model of ${\sim}T$ parameters. Notably the combined-set slope ($0.159$) is \emph{steeper} than either subset---a mixture effect, since MoE models cluster at large $N$/high accuracy and dense at small $N$/low accuracy.

\begin{table}[H]
\centering\small
\caption{Log-linear calibration fit by architecture ($\lambda=0$, cleaned set).}
\label{tab:moe-fits}
\input{tables/moe_dense_fits}
\end{table}

\textbf{Which curve should estimate (MoE) frontier models?} Frontier proprietary models are widely believed to be MoE, so one might fit only on MoE anchors. Empirically this does \emph{not} help. In leave-one-out over the 41 known-size open MoE models, the MoE-only curve recovers their sizes no better than the combined curve (median fold error $1.58\times$ vs.\ $1.52\times$ overall; $1.55\times$ vs.\ $1.55\times$ for the 34 models ${\geq}100$B). Dropping the 52 dense anchors to erase a ${\sim}2$--$4$ pp offset costs more in slope stability than the offset is worth, and per-model deviations from post-training, distillation, and refusal behavior---GLM over-knowing for its size by ${\sim}3\times$, DeepSeek V4 Pro's degraded non-thinking variant ${\sim}10\times$ low---dwarf the architecture gap. We therefore keep the combined curve as the headline: it is the \emph{validated} choice, not merely a convenient one.

\textbf{MoE-deflation caveat.} Because the combined slope is steeper, it assigns \emph{lower} sizes to the highest-accuracy models than a pure MoE curve would (Table~\ref{tab:moe-sens}): switching to the MoE-total curve raises the top estimates by ${\sim}18\%$ (e.g.\ Claude Fable 5 from ${\sim}3.5$T to ${\sim}4.0$T; GPT-5.5 from ${\sim}4.7$T to ${\sim}5.5$T). These shifts are well within the $3.2\times$ prediction interval and cannot be validated out-of-sample, so we report combined-curve values as headline while noting that presumed-MoE frontier models may be modestly under-placed.

\begin{table}[H]
\centering\small
\caption{Frontier estimate sensitivity to the calibration subset: headline combined curve vs.\ the MoE-total curve.}
\label{tab:moe-sens}
\input{tables/moe_dense_sens}
\end{table}

\section{Wikidata Long-Tail Audit}
\label{app:wikidata-audit}

This appendix expands Section~\ref{sec:wikidata-audit}: it documents the failure-mode taxonomy uncovered when auditing the T5--T7 Wikidata-sourced probes against Wikipedia and primary sources, the per-round telemetry of the 10-round repair cycle, and the empirical observation that an answer-field correction necessarily invalidates the probe's prior tier assignment.

\subsection{Failure-Mode Taxonomy}

Every problematic probe falls into exactly one of five buckets. The taxonomy is intended to be a re-usable diagnostic for any future Wikidata-sourced evaluation set.

\begin{description}[leftmargin=2em,topsep=2pt,itemsep=2pt,style=nextline]
    \item[A. Wikidata-itself-wrong.] Outright incorrect facts in Wikidata that are falsifiable via Wikipedia or a primary source.
    \begin{itemize}[leftmargin=*,topsep=1pt,itemsep=1pt]
        \item \textbf{A1. Stale data} (\texttt{P159} headquarters typically): Roku still listed as headquartered in Los Gatos despite a 2019 move to San Jose; Vista Outdoor in Clearfield despite a move to Anoka. Frequency: $\sim 75\%$ of headquarters probes were stale.
        \item \textbf{A2. Wrong field semantics}: \texttt{P170} (creator) for sculptures often returns the bronze foundry, not the sculptor---Giacometti's \emph{L'Homme qui marche I} attributed to Susse Fr\`{e}res; Bourdelle's \emph{Monument to Alvear} attributed to Eug\`{e}ne Rudier.
        \item \textbf{A3. Self-referential / circular}: ``Na Klang named after Na Klang''---a Thai descriptive place-name (``middle field'') incorrectly marked as having an eponymous referent.
        \item \textbf{A4. Plain wrong year/value}: Makran Medical College listed as 2015 when Wikipedia infobox says 2017; Lady Snowblood screenplay attributed to the manga illustrator Kazuo Kamimura rather than the actual screenwriter Norio Osada.
    \end{itemize}
    \item[B. Pipeline extraction bugs (on our side).] \texttt{P571} inception was correct in Wikidata, but our SPARQL pipeline extracted a different date field (e.g., the year the building was constructed rather than the year the museum was chartered: George Eastman Museum stored as 1905, but Wikidata \texttt{P571}=1947).
    \item[C. Probe-construction ambiguity (Wikidata correct, question unanswerable).] The most prevalent class.
    \begin{itemize}[leftmargin=*,topsep=1pt,itemsep=1pt]
        \item \textbf{C1. Generic religious-art title}: ``Madonna and Child'', ``Holy Trinity'', ``Saint Apollonia''---each refers to dozens of works by different painters across centuries.
        \item \textbf{C2. Generic geographic-feature name}: ``Stone Bridge'' (Latvia? Russia? UK?), ``South Island'' (New Zealand? or the one in Lake Turkana, Kenya?), ``Hog Island'' (Falklands, or one of several US ones?), ``Devil's Lake''.
        \item \textbf{C3. Bare common-name place}: ``Putnam'' (the Connecticut town, or Putnam County?), ``Norwich'' (Vermont? or the much more famous Norwich, England?), ``Wells''.
        \item \textbf{C4. Generic institution name}: ``Maritime Museum'' (Jakarta? San Diego? Greenwich?), ``St. Lawrence University'' (Kampala, Uganda? or the famous one in NY?).
    \end{itemize}
    \item[D. Politically/legally contested entities.] Wikidata picks one answer where reasonable models will give a different one.
    \begin{itemize}[leftmargin=*,topsep=1pt,itemsep=1pt]
        \item \textbf{D1. Disputed sovereignty}: Loaita Cay listed under Vietnam (the cay is one of the Spratlys; claimed by China, Vietnam, and the Philippines). Cape Plaka listed under Ukraine (in Crimea, internationally recognized as Ukraine but Russian-occupied since 2014).
        \item \textbf{D2. Historical-vs-current statehood}: Mount Ashley listed under ``United Kingdom''---technically the South Georgia archipelago is its own British Overseas Territory, not part of the UK proper.
    \end{itemize}
    \item[E. Multi-actor attribution.] A genuinely correct attribution that is one of several equally valid answers; reasonable models will give a different one.
    \begin{itemize}[leftmargin=*,topsep=1pt,itemsep=1pt]
        \item \textbf{E1. Co-founder selection}: Cin\'{e}math\`{e}que de Tanger has 3 founders (Lahlou, Auriol, Barrada); the National Council of Women of Canada has Adelaide Hoodless (treasurer at founding) and Lady Aberdeen (first president).
        \item \textbf{E2. Co-author / story-vs-screenplay}: Twenty-Four Eyes screenplay attributed to the novelist Sakae Tsuboi rather than the actual screenwriter (and director) Keisuke Kinoshita; The Amazing Panda Adventure screenplay attributed to John Wilcox (story credit) rather than Jeff Rothberg / Laurice Elehwany (actual screenplay).
    \end{itemize}
\end{description}

The audit response by class: A and B are resolved by correcting the answer field (and re-calibrating); C is resolved by adding a discriminating qualifier to the question template (``the town of Putnam in Connecticut''); D is resolved by either dropping the probe (Loaita Cay) or accepting Wikidata's choice with the caveat that the LLM judge accepts the documented answer; E is resolved by accepting any of the multiply-valid answers via the judge's instruction to credit co-contributors.

\subsection{Per-Round Sourcing Telemetry}

Replacement probes were sourced via 10 SPARQL rounds, each producing 100--300 candidates that were calibrated through the landmark ladder. Per-fact-type web-verification pass rates emerged empirically and informed which fact types to over-source in later rounds.

\begin{table}[H]
\centering
\small
\setlength{\tabcolsep}{4pt}
\begin{tabularx}{\textwidth}{@{}>{\raggedright\arraybackslash}X r r l >{\raggedright\arraybackslash}X@{}}
\toprule
\textbf{Round} & \textbf{Sourced} & \textbf{Calib.\ valid} & \textbf{T6/T7 yield} & \textbf{Web-verified pass} \\
\midrule
R1 (founding-year)               & 186 & 155 & 70 (T6=25, T7=45) & --- (most rejected post-mandate) \\
R2 (mixed)                       & 152 & 138 & 39 (T6=20, T7=19) & 18/27 diverse (67\%) \\
R3 (composer/birthplace)         & 0   & --- & ---               & --- (SPARQL timeouts) \\
R4 (diverse-only)                & 140 & 124 & 24 (T6=10, T7=14) & 15/24 (62\%) \\
R5 (12 fact types)               & 250 & 228 & 68 (T6=39, T7=29) & 21/29 T7 (72\%) \\
R6 (11 fact types + castle)      & 275 & 256 & 64 (T6=36, T7=28) & 18/28 T7 (64\%) \\
R7 (7 reliable types only)       & 140 & 130 & 21 (T6=9, T7=12)  & 11/12 T7 (\textbf{92\%}) \\
R8 (aggressive ambiguity filter) & 100 & 92  & 24 (T6=13, T7=11) & 7/11 T7 (64\%) \\
R9 (most reliable types)         & 75  & 71  & 26 (T6=8, T7=18)  & 18/18 T7 (\textbf{100\%}) \\
R10 (grounded SPARQL templates)  & 45  & 15  & 5 (T6=3, T7=2)    & --- (low yield; alt path used) \\
\bottomrule
\end{tabularx}
\caption{Per-round sourcing telemetry. Total API spend across all rounds: ${\sim}10{,}000$ landmark+judge calls and ${\sim}120$ manual web verifications.}
\label{tab:audit-telemetry}
\end{table}

\subsection{Per-Fact-Type Reliability Ranking}

Aggregating across rounds 4--7, the empirical reliability of each Wikidata fact type at the long tail (T6/T7 obscurity):

\begin{enumerate}[leftmargin=*,topsep=2pt,itemsep=1pt]
    \item \textbf{\texttt{P403} river mouth} --- 100\% pass rate. Geography is well-curated.
    \item \textbf{\texttt{P57} director (film)} --- 67--100\% on attribution; ambiguity from generic film titles (``Night Train'', ``Hit Man'') is the dominant failure mode and is resolved by adding year + country + genre to the template.
    \item \textbf{\texttt{P17} country (cape/island/lake/mountain)} --- 75--100\%. Caveat: politically contested entities (Crimea, Spratlys, South Georgia) consistently fail web cross-check because Wikidata picks one claimant.
    \item \textbf{\texttt{P571} inception (founding year)} --- ${\sim}95\%$. The most-curated date field. Most failures trace to pipeline-side date-field selection bugs (Bucket B), not Wikidata error.
    \item \textbf{\texttt{P112} founder (organization)} --- 75\%. Sometimes returns the founding \emph{organization} rather than a person (e.g., St John Clinic founded by ``Order of the Holy Ghost''), making the question awkward when phrased ``Who founded \ldots?''.
    \item \textbf{\texttt{P170} creator (sculpture)} --- ${\sim}70\%$. The major Bucket-A2 failure: Wikidata's creator field for sculptures frequently records the bronze foundry rather than the sculptor.
    \item \textbf{\texttt{P58} screenwriter} --- 50\%. Routinely conflates ``story by'' credit with ``screenplay by'' credit (e.g., the manga author of \emph{Lady Snowblood} listed as the film's screenwriter).
    \item \textbf{\texttt{P159} headquarters} --- 25\%. \emph{Stale}. Wikidata frequently reflects a company's HQ from years prior; modern moves are not propagated quickly.
    \item \textbf{\texttt{P170} creator (painting)} --- ${\sim}70\%$ on attribution, but ${\sim}100\%$ title-collision rate at T6/T7 obscurity (generic religious-art titles). Painter probes are essentially unusable for T7 obscurity.
    \item \textbf{\texttt{P138} named after} --- mixed; self-referential and descriptive-placename errors observed.
\end{enumerate}

\section{Full Model Results}
\label{app:full-results}

Tables~\ref{tab:full-accuracy} and~\ref{tab:full-hallucination} present the complete evaluation results for all 201 models.
Models are grouped by vendor and sorted by overall accuracy (descending) within each group.
Vendors are ordered by the highest accuracy achieved by any model in that group.
The \emph{Acc.}\ column reports the tier-mean accuracy at $\lambda = 0$ (no penalty; Section~\ref{sec:setup}); \emph{Raw} is overall correct/total.
The \emph{Raw} column reports raw accuracy without hallucination penalty.
Per-tier accuracy (T1--T7) is reported in the range $[0, 1]$.

\begin{figure}[H]
    \centering
    \includegraphics[width=0.85\textwidth]{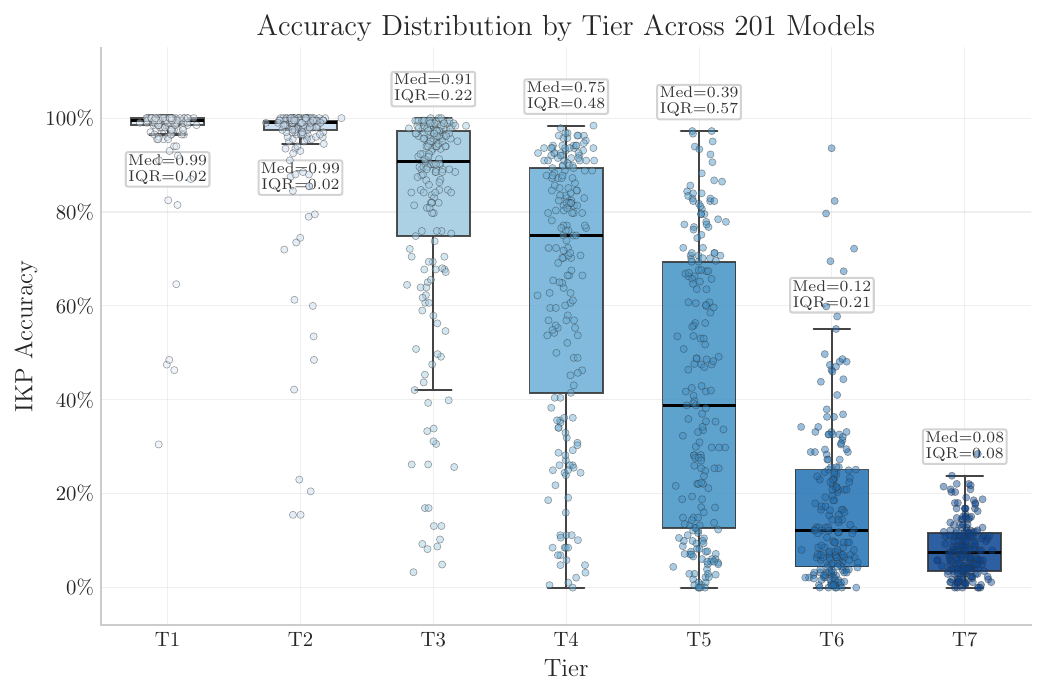}
    \caption{Accuracy distribution by tier across 201 models. T1--T2 are compressed at ceiling (median $>98\%$). T4 has the widest interquartile range, making it the best population discriminator. T6--T7 discriminate only the strongest models.}
    \label{fig:tier-boxplots}
\end{figure}

\begin{figure}[H]
    \centering
    \includegraphics[width=\textwidth]{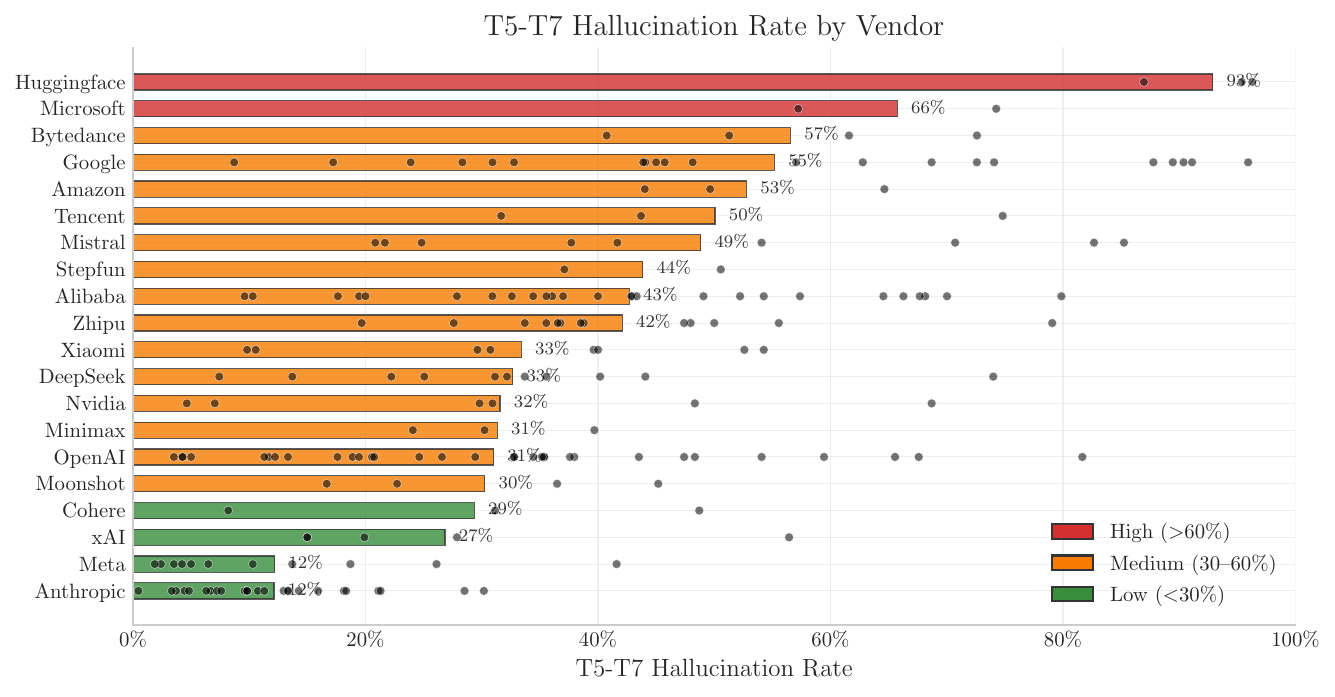}
    \caption{T5--T7 hallucination rate by vendor. Individual model points are overlaid on vendor means. Anthropic models hallucinate at only 10\% on probes beyond their knowledge (preferring to refuse), while Google models hallucinate at 58--66\%. This vendor-specific calibration directly impacts IKP scoring.}
    \label{fig:vendor-hallucination}
\end{figure}

Figure~\ref{fig:tier-boxplots} shows the distribution of per-tier accuracy across all models, illustrating how each tier discriminates at different capability levels. Figure~\ref{fig:vendor-hallucination} shows the hallucination rate by vendor on T5--T7 probes, revealing stark differences in safety calibration across vendors.

\begin{landscape}
\input{tables/full_accuracy}

\end{landscape}

\begin{landscape}
\input{tables/full_hallucination}
\end{landscape}

\section{Densing Law Falsification: Full Regression Results}
\label{app:densing}

This appendix provides the full regression tables supporting the Densing-Law falsification in Section~\ref{sec:densing-falsification}.

\paragraph{Dataset.} We restrict to open-weight models with published parameter counts and completed IKP evaluations: $n = 100$ models from 23 vendors, with release dates spanning 2023-09-27 (Mistral-7B) to 2026-06-12. Release dates were manually collected and verified against authoritative sources (vendor blog posts, HuggingFace model cards, and archived press releases). For `-think' variants that share weights with a non-thinking base, the release date equals that of the base model. The full dated configuration, the analysis script, and the tidy data CSV are released with the paper.

\paragraph{Specifications.} All fits are OLS with IKP accuracy ($\lambda=0$) as the dependent variable. \texttt{months} is the release date minus 2024-01-01, in months. \texttt{thinking} is a 0/1 indicator; \texttt{MoE} is a 0/1 indicator for mixture-of-experts architecture. The baseline M0 contains only $\log_{10}(N_B)$.

\begin{table}[ht]
    \centering
    \small
    \begin{tabular}{l c c c c c}
    \toprule
    Specification & $\log_{10}(N_B)$ & months & thinking & MoE & $R^2$ \\
    \midrule
    M0: log10(params) only        & $+0.1494^{***}$ & --- & --- & --- & $0.8148$ \\
    M1: + months                  & $+0.1445^{***}$ & $+0.0013$ & --- & --- & $0.8181$ \\
    M2: + thinking                & $+0.1421^{***}$ & $+0.0008$ & $+0.0277$ & --- & $0.8237$ \\
    M3: + MoE                     & $+0.1415^{***}$ & $+0.0007$ & $+0.0276$ & $+0.0022$ & $0.8237$ \\
    M$_t$: time only              & ---             & $+0.0095^{***}$ & --- & --- & $0.2180$ \\
    \bottomrule
    \end{tabular}
    \caption{Full regression results for the Densing-Law analysis ($n = 100$). Entries are point estimates; $^{***}p < 0.001$, $^{**}p < 0.01$. The time coefficient is statistically indistinguishable from zero in every specification that controls for parameter count (M1--M3). The time-only model (M$_t$) produces a nominally significant but much smaller slope than the Densing prediction, and its explanatory power ($R^2 = 0.22$) collapses once log-params is included---indicating the residual time correlation is a selection effect (newer models skew larger), not a real per-parameter trend.}
    \label{tab:densing-full}
\end{table}

\paragraph{Tests against the Densing prediction.} Under the observed $\log_{10}(N)$ slope of $\beta_1 = 0.149$, the Densing Law predicts a monthly accuracy gain of $\beta_2^{\mathrm{Densing}} = \beta_1 \cdot \log_{10}(2)/3.5 \approx +0.01285$/month. Table~\ref{tab:densing-tests} reports two-sided $t$-tests of the fitted time coefficient against both $0$ and this Densing target.

\begin{table}[ht]
    \centering
    \small
    \begin{tabular}{l r r r r r}
    \toprule
    Spec & $\hat\beta_2$ (months) & SE & $p$ (vs 0) & $t$ (vs Densing) & $p$ (vs Densing) \\
    \midrule
    M1 & $+0.00131$ & $0.00099$ & $0.19$ & $-11.64$ & ${<}10^{-15}$ \\
    M2 & $+0.00075$ & $0.00103$ & $0.47$ & $-11.72$ & ${<}10^{-15}$ \\
    M3 & $+0.00069$ & $0.00107$ & $0.52$ & $-10.93$ & ${<}10^{-15}$ \\
    \bottomrule
    \end{tabular}
    \caption{Hypothesis tests on the time coefficient. In every specification, the null of zero time trend is not rejected (all $p > 0.3$), while the Densing-Law prediction is rejected with effectively certainty ($|t| > 10$).}
    \label{tab:densing-tests}
\end{table}

\paragraph{Bootstrap inference.} A 4000-replicate nonparametric bootstrap on M1 gives a 95\% CI for the time coefficient of $[-0.00037, +0.00326]$, spanning zero and well below the Densing target of $+0.01285$. The gap between the upper bound and the Densing prediction is roughly $11\times$ the bootstrap standard error.

\paragraph{Robustness.} Three additional checks confirm the main finding: (i) excluding the two DeepSeek-R1 distilled models ($n = 98$) leaves $\hat\beta_2 = +0.00135$/month ($p = 0.18$ vs.\ zero); (ii) excluding the six sub-1B anchor models ($n = 94$) leaves $\hat\beta_2 = +0.00197$/month ($p = 0.06$). In both cases the Densing prediction is rejected at $p < 10^{-14}$. Check (iii) uses $\log_{10}(\text{active\_B})$ in place of total parameters and yields $\hat\beta_2 = +0.0084$/month---nominally significant but still only two-thirds of the Densing target, and also rejecting the Densing prediction ($p \approx 10^{-4}$). The apparent time effect on the active-params regression is a known artifact: MoE models store knowledge across all experts, not just the ones activated per token (Section~\ref{sec:results}), so active-params overstates $N_\mathrm{fact}$ for dense models and understates it for MoEs in a way that correlates with release date (newer models are more often MoE). This is why our preferred specification uses total parameters.

\paragraph{Date uncertainty.} Release dates for post-January-2026 models (e.g., Claude Opus 4.7, Qwen 3.5 series, Kimi K2.6, GLM-5.1, Ling-2.6-flash) were collected contemporaneously; other dates are vendor-announced release dates. A small amount of date noise (${\pm}1$ month per observation) would inflate $\mathrm{SE}(\hat\beta_2)$ but leaves the point estimate unchanged in expectation. To hide a Densing-magnitude effect would require systematic backdating of post-2025 releases by ${>}3$ months each, which is not what our sources support.

\section{Cross-Generation and Model Family Analysis}
\label{app:generations}

\begin{figure}[H]
    \centering
    \includegraphics[width=\textwidth]{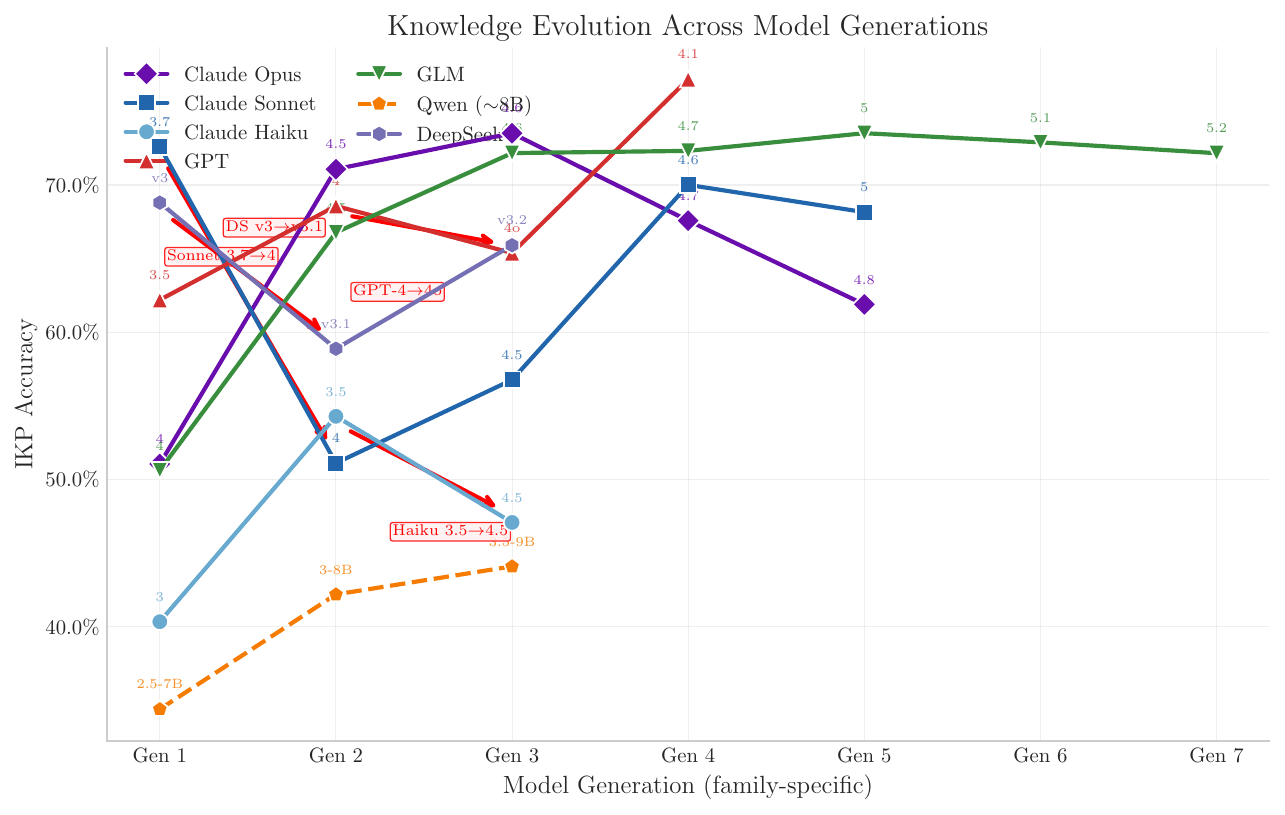}
    \caption{Knowledge evolution across model generations. Red boxes highlight regressions: GPT-4$\to$4o (smaller efficiency-optimized successor), Claude Sonnet 3.7$\to$4 (RLHF conservatism), Claude Haiku 3.5$\to$4.5 (smaller post-training). GLM shows the steadiest improvement across 6 generations.}
    \label{fig:gen-trajectories}
\end{figure}

\begin{figure}[H]
    \centering
    \includegraphics[width=0.9\textwidth]{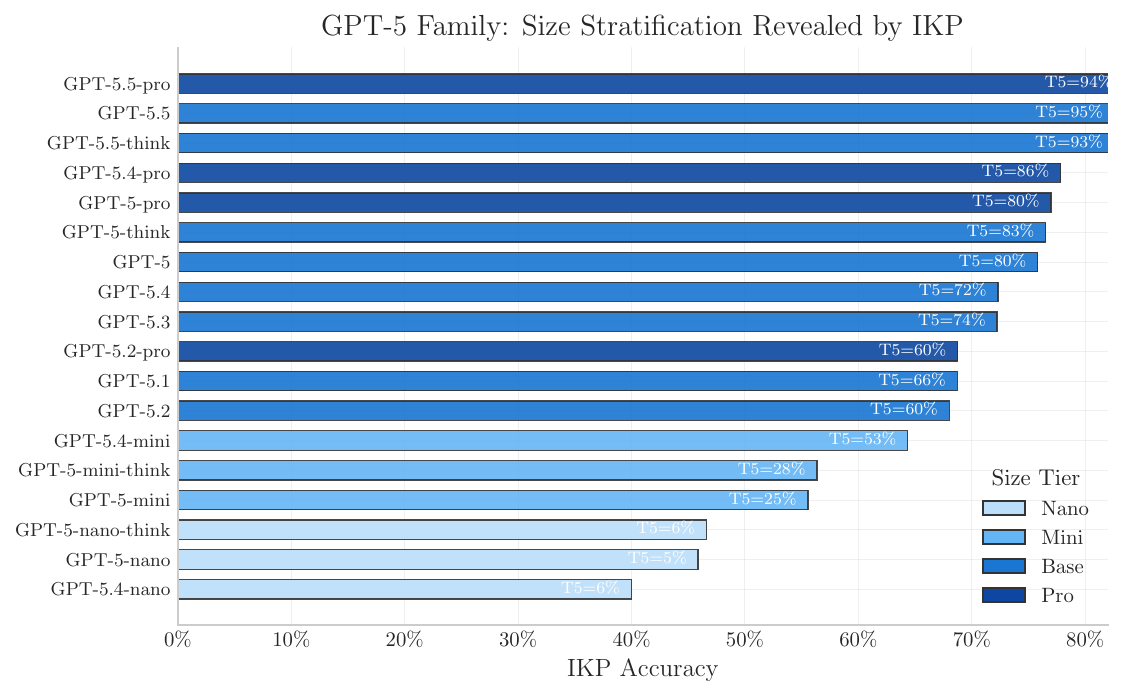}
    \caption{GPT-5 family size stratification revealed by IKP. T5 accuracy (annotated) shows clear tiers: nano (5--6\%), mini (25--53\%), base (60--95\%), pro (60--94\%). GPT-5, GPT-5-pro, and GPT-5-think score nearly identically, suggesting the same base model with different inference configurations.}
    \label{fig:gpt5-family}
\end{figure}

Figure~\ref{fig:gen-trajectories} traces knowledge evolution across model generations for 7 major families. Most families show steady improvement, but several notable regressions occur when vendors optimize for efficiency or safety at the expense of factual recall. Figure~\ref{fig:gpt5-family} reveals the internal size stratification of the GPT-5 family.

\paragraph{GPT-5 family internal structure.} The 15 GPT-5 variants in our evaluation reveal OpenAI's model lineup structure through both coarse (overall-accuracy) and fine (knowledge-fingerprint) analysis. Clear size stratification appears at T5: nano (5\%), mini (25\%), base (80\%)---a ${\sim}16\times$ knowledge gap---with a ``pro'' shelf above the base (GPT-5.4-pro at 77.8\% overall, the strongest OpenAI model below the 5.5 generation). Each \texttt{.x} iteration generally improves overall accuracy, but GPT-5.4-nano regresses from GPT-5-nano on T3 (59.0\% vs 75.4\%), suggesting a different distillation approach between generations. The fingerprint evidence (Section~\ref{sec:fingerprint}, Appendix~\ref{app:fingerprint}) is consistent with a shared-base GPT-5 / 5-pro / 5-think cluster followed by a sequence of retrains: each \texttt{.x} release (5.1, 5.2, 5.3, 5.4) sits in the retrained regime relative to its predecessor, with only the ``pro'' variants (5.2-pro, 5.4-pro) tied to their base by lineage-level HSS.

\section{Knowledge Fingerprinting: Detailed Tables}
\label{app:fingerprint}

This appendix expands Section~\ref{sec:fingerprint} with the complete consecutive-pair tables for each tracked model family and the full list of cross-vendor outliers. For every pair of models we report, on the 400 T5--T6 probes, the Jaccard similarity $J$ of correct-answer sets, the lift ($= \text{observed}/\text{expected}$, under the null of independence) of the correct-set intersection, the hallucination similarity $\mathrm{HSS}$, the number of joint-wrong probes (``both\_w''), and a regime classification (shared-base, lineage, retrained, independent). The thresholds are those stated in Section~\ref{sec:fingerprint}: shared-base requires $\mathrm{HSS} \geq 0.30$ and $J \geq 0.60$; lineage requires $\mathrm{HSS} \geq 0.10$ and $J \geq 0.50$; retrained is $\mathrm{HSS} < 0.10$ with $\geq 10$ joint-wrong probes (limited statistical resolution when both\_w is much smaller).

\paragraph{Why HSS?} Raw Jaccard on correct-answer sets is dominated by probes that nearly all frontier models get right, producing high $J$ even for truly independent models (for example, GPT-5 vs.\ Claude Opus 4.5 scores $J = 0.56$ despite being independent trainings by different vendors). Lift partially corrects for this but is noisy when either model's correct set is small. Hallucination similarity---the rate at which two models produce the same normalized wrong answer on a probe that both got wrong---is a much cleaner signal: independent models almost never converge on the same wrong rare fact, while weight-sharing siblings do so on 30--55\% of joint-wrong probes. A bright threshold around $\mathrm{HSS} = 0.30$ separates the two regimes empirically.

\paragraph{When the three metrics disagree.} The three metrics are complementary, not redundant. Heavy post-training can preserve rare-fact knowledge (high $J$ and high lift) while rewriting the generation style enough that a model gives \emph{differently worded} wrong answers on hard probes---driving $\mathrm{HSS}$ down. Nemotron-70B vs.\ Llama-3.1-70B is the textbook example: $J = 0.81$ and lift = 6.8 (clear shared base) but $\mathrm{HSS} = 0.08$ (SFT has scrambled the surface form of wrong answers). Conversely, a high-$\mathrm{HSS}$ pair with low $J$ and low lift is a weaker signal than it sounds, because it may reflect a single over-represented hallucination (e.g., a common name collision) rather than broad knowledge overlap. We classify a pair as \emph{shared base} only when all three metrics point the same way; one-sided signals are reported as ``possible lineage'' and should be investigated case-by-case.

\paragraph{Release-practice summary.} Applied to consecutive-generation pairs across all tracked families (see Table~\ref{tab:fp-all-families}), three release patterns recur:
\begin{itemize}[leftmargin=*]
    \item \emph{Fine-tune-on-same-base releases} (lineage): Claude Opus 4 $\to$ 4.1, Sonnet 4 $\to$ 4.5, DeepSeek V3 $\to$ V3.1 $\to$ V3.2, GLM 4.6 $\to$ 4.7, Kimi K2 $\to$ K2.5 $\to$ K2.6. These preserve a large fraction of the base's rare-fact mistakes.
    \item \emph{Inference/alignment variants of a shared base} (shared-base): GPT-5 / GPT-5-pro / GPT-5-think; Claude Opus 4.5 / 4.5-think; DeepSeek V3.2 / V3.2-think. These show the highest $\mathrm{HSS}$ in the dataset.
    \item \emph{Full retrains} (retrained): every GPT-5.$x$ $\to$ 5.$(x{+}1)$ transition for $x = 0,1,2,3$; Claude Opus 4.6 $\to$ 4.7; every cross-generation Gemini pair; GLM 4.7 $\to$ 5 and 5 $\to$ 5.1. For these, the same-wrong-answer count is statistically indistinguishable from what we see between models built by different vendors, despite the shared version-family label.
\end{itemize}

\paragraph{Cross-vendor outliers: how to read the table.} Applying the same test to the ${\sim}14{,}000$ cross-vendor pairs flags a small fraction at $\mathrm{HSS} \geq 0.20$ with $\geq 10$ joint-wrong probes (the full list is Table~\ref{tab:fp-cross-full}). Interpretation requires care: a single high-$\mathrm{HSS}$ pair between two independent frontier models is not strong evidence of cross-family distillation given the number of pairs tested. More suggestive is the \emph{pattern} across pairs involving the same model on one side:

\begin{itemize}[leftmargin=*]
    \item \textbf{Baidu ERNIE 4.5-300B-A47B} is above the cross-vendor threshold against four independent models simultaneously (Llama-3-70B $\mathrm{HSS}=0.50$, GPT-4o $0.46$, Mistral-Small-24B $0.40$, Qwen-Max $0.36$). This is the pattern expected when a model is trained on a large mixture of distilled outputs from Western frontier models rather than on a single teacher.
    \item \textbf{Llama 3.1 70B} appears as an apparent ``teacher'' against many other models ($\mathrm{HSS} \geq 0.30$ vs.\ grok-3, gemini-2.0-flash, qwen3-max, GPT-4.1-nano, and several other models). This is consistent with Llama 3.1 being the most widely-used open base for synthetic-data generation in 2024--2025, so its characteristic hallucinations leak into many downstream datasets. It is \emph{not} evidence that any single downstream vendor distilled from it directly.
    \item \textbf{deepseek-r1-distill-llama-70b vs.\ Llama-3.1-70B} ($\mathrm{HSS} = 0.31$) is a known distillation-into-base pairing and serves as a positive control for the method.
    \item Individual high-$\mathrm{HSS}$ frontier-vs-frontier pairs (GPT-5 vs.\ Grok-4 at $0.38$; GPT-5-pro vs.\ Kimi-K2.6 at $0.32$) are individually above threshold but have too few joint-wrong probes, at our probe count, to reject an innocent-explanation null (e.g., training on similar recent web snapshots or on overlapping synthetic-data mixtures).
\end{itemize}

\paragraph{Limitations.} The method identifies \emph{statistical} overlap in knowledge and hallucination patterns; it does not distinguish between direct distillation and shared training data (e.g., two vendors independently scraping the same recent Wikipedia dump or licensing the same curated corpus). It also has limited resolution when the number of joint-wrong rare-fact probes is small: frontier models that each score 70--80\% on T5--T6 jointly get only 20--50 probes wrong, and $\mathrm{HSS}$ estimated from small samples is noisy. Probe-pool expansion into T6--T7 would tighten these estimates and is a natural next step.

\paragraph{Known-provenance positive controls.} As a sanity check, Table~\ref{tab:fp-controls} reports metrics for six pairs whose parent-child relationship is publicly disclosed. Every pair with $\geq 10$ joint-wrong probes shows elevated lift (2.1--7.9) and $J$ (0.15--0.81) consistent with shared training signal; $\mathrm{HSS}$ on heavily post-trained derivatives (Nemotron, R1-distill) is suppressed for the reason discussed above, which is why we recommend using all three metrics jointly rather than thresholding $\mathrm{HSS}$ alone.

\input{tables/fp_controls.tex}

\paragraph{Comprehensive all-family table.} Table~\ref{tab:fp-all-families} reports consecutive-generation metrics for every tracked family (open-weight and closed-weight) where at least one consecutive pair has $\geq 8$ joint-wrong probes. Three patterns recur across the ${\sim}30$ families:
\begin{itemize}[leftmargin=*]
    \item \emph{Frequent retrains}: Gemini Flash/Pro, Gemma 2/3/4, Qwen 2.5/3/3.5, most GPT-5 \texttt{.x} transitions, and the Opus 4.6/4.7 step all sit in the retrained regime. These vendors appear to prefer full retrains over incremental post-training when bumping version numbers.
    \item \emph{Incremental lineages}: DeepSeek V3 $\to$ V3.1 $\to$ V3.2, GLM 4.6 $\to$ 4.7, Kimi K2 $\to$ K2.5 $\to$ K2.6, and the Nova / Command families show consistent lineage-regime HSS, consistent with continued pretraining or large-scale post-training on a stable base.
    \item \emph{Shared-base variants}: Thinking / reasoning variants (Opus 4.5 / 4.5-think, GPT-5 / GPT-5-pro, DeepSeek V3.2 / V3.2-think) and ``pro'' variants of the same base generation concentrate at $\mathrm{HSS} \geq 0.30$.
\end{itemize}

\input{tables/fp_all_families.tex}

\paragraph{All cross-vendor outliers.} Table~\ref{tab:fp-cross-full} gives the full ranked list of cross-vendor pairs above our reporting threshold.

\input{tables/fp_cross_vendor.tex}

\section{Case Studies}
\label{app:case-studies}

This section presents concrete examples of probe questions and model responses to illustrate key findings from the main text. All responses are verbatim model outputs (truncated for space).

\subsection{USTC Hackergame --- Watching a Fact Arrive Over Three Years}
\label{app:hackergame-case}

The motivating observation in Section~\ref{sec:intro} traces the arrival of a single piece of long-tail knowledge---specific challenge names from \emph{USTC Hackergame}, an annual Capture-the-Flag competition organized by the USTC Linux User Group since 2014---into frontier model parameters across release cycles. We probed 12 representative models (across 2022--2026) with a uniform prompt asking for specific challenge names per year, then verified each response against the official writeup repositories at \url{https://github.com/USTC-Hackergame} and \url{https://github.com/ustclug/hackergame}. Results are summarized in Table~\ref{tab:hackergame-case}.

\paragraph{Three observations from this case study.}
\begin{enumerate}[leftmargin=*,topsep=2pt,itemsep=2pt]
    \item \textbf{The transition is sharp and dateable.} GPT-4o (May 2024) hallucinates a deterministic fake list, repeated across independent probes. Claude 3.7 Sonnet (February 2025) is the first model to list real per-year challenges, with $\sim$19 verified 2023 titles. Whatever change to the training pipeline absorbed Hackergame writeups happened in this $\sim$9-month window, not gradually across the field; later models (Claude Opus 4.6/4.7, GPT-5.5, Kimi K2.6) inherited and extended this knowledge.
    \item \textbf{Knowing the meta-fact does not imply knowing the content.} DeepSeek V4 Pro (1.6T) is the only model that correctly states 2014 as the start year (a single Wikipedia-grade meta-fact) but fabricates per-year challenges. The structural fact and the long-tail content are stored independently.
    \item \textbf{Refusal vs hallucination is a vendor-level choice that interacts with measurement.} GPT-4o and DeepSeek V4 Pro confidently fabricate; Claude Opus 4.7 says ``I don't have reliable detailed memory and don't want to fabricate specifics''; GPT-5/5.5 burn output budget on reasoning and produce empty content.
\end{enumerate}

This case is one motivating example among many; the systematic IKP probe set (Section~\ref{sec:probe-generation}) operationalizes the same kind of observation across 1{,}400 facts and 201 models, with quantitative scoring rather than human verification.

\begin{table}[H]
\centering
\small
\setlength{\tabcolsep}{4pt}
\begin{tabular}{l l p{10cm}}
\toprule
\textbf{Model} & \textbf{Released} & \textbf{Verification of LLM Responses} \\
\midrule
GPT-3.5 Turbo & 2022-11 & Hallucinates: claims contest started in ``2006'' (actual: 2014); claims ``offline finals'' (contest is online only). \\
GPT-4 & 2023-03 & Admits no specific knowledge. Generic CTF description. \\
GPT-4o & 2024-05 & Knows USTC + CTF format. Asked for per-year challenges, returns the \emph{identical} fabricated list (``Hello World'', ``Maze'', ``Calculator'', ``Reverse Engineering'', ``Quantum Computing'') for nine different years (2015--2023). None of these are real Hackergame challenges; the pattern is deterministic fabrication, repeated across two independent probes. \\
\textbf{Claude 3.7 Sonnet} & \textbf{2025-02} & \textbf{First model to list real per-year challenges.} 2023: ``Kitty Quiz'', ``Travel Photo'', ``Cyber Tic-Tac-Toe'', ``Committee Simulator'', ``High-Frequency Planet'', ``Streaming Planet'', ``Small LM Planet'', ``Word-Frugal'', ``Grandma's Bedtime flag Story'', ``Alien Detour'' (all verified real, English glosses of Chinese originals). Also accurate 2022 (HeiLang, Xcaptcha, plus original-Chinese titles) and 2021 (similar). \\
Gemini 2.5 Pro & 2025-03 & Lists $\sim$6 real challenges per year for 2021--2023 (e.g., HeiLang, ``JSON $\subset$ YAML?'', plus various original-Chinese titles), with one likely fabrication (a non-existent planet-series title). \\
GPT-5 & 2025-08 & Returns reasoning trace only; content cut off before any answer is written. \\
\textbf{Claude Opus 4.6} & \textbf{2025-11} & \textbf{Lists 16+ real Hackergame 2023 challenges, all verified}: ``Kitty Quiz'', ``Deeper Darker'', ``Travel Photo 3.0'', ``Cyber Tic-Tac-Toe'', ``Grandma's Bedtime flag Story'', ``Committee Simulator'', ``Worm'', ``JSON $\subset$ YAML?'', ``Git? Git!'', ``HTTP Stamp Album'', ``Docker for Everyone'', ``Word-Frugal 2.0'', ``High-Frequency Planet'', ``Streaming Planet'', ``Komm süsser satisfiability'', ``Alien Detour''. Earlier years also accurate. \\
Gemini 3.1 Pro & 2026-01 & Year-by-year breakdown 2018--2023 with specific real challenge names (``Travel Photo'', ``Selling Melons'', ``Transparent File'', ``Kitty Quiz'' series). Concise but accurate. \\
Kimi K2.6-think & 2026-03 & 30+ specific names across 2020--2023, mostly verified real (HeiLang, Xcaptcha, ``Cup-Window Goose-Shadow''/Wine, ``Komm süsser Flagge''), with occasional year-mixing. \\
DeepSeek V4 Pro & 2026-04 & Hallucinates: knows the meta-fact (correctly says contest started 2014, the only model that does) but fabricates per-year challenges as generic CTF tropes (``check-in problem'', ``classical cryptography'', ``web dog daily'', ``PWN intro'')---none are actual Hackergame names. \\
GPT-5.5 & 2026-04 & Lists real 2023 names (``Sign-In'', ``Kitty Quiz'', ``Deeper Darker'', ``Travel Photo 3.0'', ``JSON $\subset$ YAML?'', ``Git?'', ``HTTP Stamp Album'', ``Docker for Everyone'', ``Word-Frugal 2.0''). \\
Claude Opus 4.7 & 2026-04 & Year-by-year list 2018--2024. Specific names per year mostly real, with occasional cross-year mixing. Says contest started 2018 (incorrect: that was the year of the LUG renaming; contest itself dates to 2014). \\
\bottomrule
\end{tabular}
\caption{USTC Hackergame probe across 12 representative frontier models. Verification was performed against the official writeup repositories (\texttt{github.com/USTC-Hackergame/hackergame20\{18..23\}-writeups}) and the platform repo (\texttt{github.com/ustclug/hackergame}).}
\label{tab:hackergame-case}
\end{table}

\subsection{Tier Discrimination: Universal Knowledge (T1)}
\label{app:case-t1}

T1 probes test universal knowledge that every model, regardless of size, should know. Even the smallest 1B-parameter models answer these correctly.

\begin{quote}
\textbf{Probe:} What is the capital of France? \\
\textbf{Tier:} T1 \quad \textbf{Gold answer:} Paris \\
\textbf{Domain:} general \\
\end{quote}

\begin{description}[style=nextline, leftmargin=1.5cm]
\item[\textbf{Gemma-3 1B}] \textcolor{green!60!black}{\textsc{correct}} --- Paris.
\item[\textbf{Llama-3.2 3B}] \textcolor{green!60!black}{\textsc{correct}} --- Paris.
\item[\textbf{GPT-4o}] \textcolor{green!60!black}{\textsc{correct}} --- The capital of France is Paris.
\item[\textbf{Claude Opus 4.6}] \textcolor{green!60!black}{\textsc{correct}} --- Paris.
\end{description}

\subsection{Tier Discrimination: Mid-Tier Knowledge (T3)}
\label{app:case-t3}

T3 probes separate small models from medium and large ones. Small models ($<$10B) typically refuse or hallucinate, while models above ${\sim}30$B answer correctly. This probe tests a moderately obscure scientific fact.

\begin{quote}
\textbf{Probe:} Who first produced liquid helium? \\
\textbf{Tier:} T3 \quad \textbf{Gold answer:} Heike Kamerlingh Onnes \\
\textbf{Domain:} science \\
\end{quote}

\begin{description}[style=nextline, leftmargin=1.5cm]
\item[\textbf{Gemma-3 4B}] \textcolor{red!70!black}{\textsc{wrong}} --- William Ramsay and Lord Rayleigh.
\item[\textbf{Llama-3.2 3B}] \textcolor{orange!80!black}{\textsc{refusal}} --- I don't know.
\item[\textbf{GPT-4o}] \textcolor{green!60!black}{\textsc{correct}} --- Heike Kamerlingh Onnes first produced liquid helium in 1908.
\item[\textbf{Claude Opus 4.6}] \textcolor{green!60!black}{\textsc{correct}} --- Heike Kamerlingh Onnes first produced liquid helium on July 10, 1908, at his laboratory in Leiden, the Netherlands. This achievement earned him widespread recognition and contributed to his receiving the Nobel Prize in P...
\end{description}

\subsection{Tier Discrimination: Frontier-Only Knowledge (T6)}
\label{app:case-t6}

T6 probes are only answered correctly by the largest frontier models. These typically involve obscure researchers or rarely documented facts. Even strong models like GPT-4o refuse on these probes, while the top-scoring Gemini models retrieve the information.

\begin{quote}
\textbf{Probe:} In computer science, what is the research subfield of Janick Edinger, and name one paper, system, institution, or co-author associated with their work? If you don't know who this person is, say so. \\
\textbf{Tier:} T6 \quad \textbf{Gold answer:} distributed systems \\
\textbf{Domain:} computer\_science \\
\end{quote}

\begin{description}[style=nextline, leftmargin=1.5cm]
\item[\textbf{Llama-3.3 70B}] \textcolor{orange!80!black}{\textsc{refusal}} --- I don't know.
\item[\textbf{GPT-4o}] \textcolor{orange!80!black}{\textsc{refusal}} --- I don't know.
\item[\textbf{Gemini 3 Flash}] \textcolor{green!60!black}{\textsc{correct}} --- Janick Edinger's research subfield is distributed systems, with a specific focus on edge computing, pervasive computing, and mobile computing.
\item[\textbf{Gemini 3.1 Pro}] \textcolor{green!60!black}{\textsc{correct}} --- Janick Edinger's research subfields in computer science are primarily distributed systems, edge computing, pervasive/ubiquitous computing, and the Internet of Things (IoT).
\end{description}

\subsection{Hallucination vs.\ Refusal}
\label{app:case-hallucination}

A critical distinction in IKP evaluation is between hallucination (confidently wrong) and refusal (saying one does not know). Our scoring penalizes hallucination but not refusal, reflecting the real-world risk of confident misinformation. Here, Claude Opus 4.6 answers correctly, Llama-3.3 70B appropriately refuses, but GPT-4o-mini hallucinates a plausible but wrong subfield.

\begin{quote}
\textbf{Probe:} In computer science, what is the research subfield of Zhongzhi Luan, and name one paper, system, institution, or co-author associated with their work? If you don't know who this person is, say so. \\
\textbf{Tier:} T5 \quad \textbf{Gold answer:} computer architecture \\
\textbf{Domain:} computer\_science \\
\end{quote}

\begin{description}[style=nextline, leftmargin=1.5cm]
\item[\textbf{Claude Opus 4.6}] \textcolor{green!60!black}{\textsc{correct}} --- Zhongzhi Luan is a researcher in the field of high-performance computing (HPC). His research interests include parallel computing, large-scale cluster systems, and related areas in computer architecture and systems. He h...
\item[\textbf{Llama-3.3 70B}] \textcolor{orange!80!black}{\textsc{refusal}} --- I don't know.
\item[\textbf{GPT-4o-mini}] \textcolor{red!70!black}{\textsc{wrong}} --- Zhongzhi Luan is known for his research in the field of artificial intelligence, particularly in areas related to machine learning, data mining, and complex networks.
\end{description}

\subsection{Name Collision in Researcher Probes}
\label{app:case-name-collision}

Researcher probes are particularly susceptible to name collision: the model recognizes the name but associates it with a different person in the same or a different field. Here, ``Geoffrey Ye Li'' is a wireless communications researcher, but GPT-4o-mini confuses him with Geoffrey Hinton-adjacent ML researchers, producing a confident but wrong answer about ``deep learning and reinforcement learning.''

\begin{quote}
\textbf{Probe:} In computer science, what is the research subfield of Geoffrey Ye Li, and name one paper, system, institution, or co-author associated with their work? If you don't know who this person is, say so. \\
\textbf{Tier:} T3 \quad \textbf{Gold answer:} computer networking \\
\textbf{Domain:} computer\_science \\
\end{quote}

\begin{description}[style=nextline, leftmargin=1.5cm]
\item[\textbf{GPT-4o-mini}] \textcolor{red!70!black}{\textsc{wrong}} --- Geoffrey Ye Li is known for his research in the field of machine learning, particularly in areas related to deep learning, reinforcement learning, and their applications in various domains.
\item[\textbf{GPT-4o}] \textcolor{green!60!black}{\textsc{correct}} --- Geoffrey Ye Li is known for his research in the field of wireless communications and signal processing.
\item[\textbf{Claude Opus 4.6}] \textcolor{green!60!black}{\textsc{correct}} --- Geoffrey Ye Li is a prominent researcher in the field of wireless communications and signal processing. His research subfields include:  - OFDM (Orthogonal Frequency-Division Multiplexing) - MIMO (Multiple-Input Multiple...
\item[\textbf{Llama-3.3 70B}] \textcolor{green!60!black}{\textsc{correct}} --- Geoffrey Ye Li's research subfield is wireless communications and signal processing.
\end{description}

\subsection{Thinking Mode Enabling Correct Recall}
\label{app:case-thinking}

Extended thinking (chain-of-thought reasoning) can help models retrieve knowledge they would otherwise fail to access. Here, the base Claude Sonnet 4 refuses the question about a classical Japanese poetry anthology, but the thinking variant correctly identifies Fujiwara no Teika as the compiler. The thinking trace allows the model to reason through partial knowledge to arrive at the correct answer.

\begin{quote}
\textbf{Probe:} What Japanese poet wrote the Hyakunin Isshu anthology? \\
\textbf{Tier:} T3 \quad \textbf{Gold answer:} Fujiwara no Teika \\
\textbf{Domain:} Non-Western Cultural Works \\
\end{quote}

\begin{description}[style=nextline, leftmargin=1.5cm]
\item[\textbf{Claude Sonnet 4}] \textcolor{orange!80!black}{\textsc{refusal}} --- I don't know who compiled the Hyakunin Isshu anthology. While I know it's a famous collection of 100 Japanese poems by 100 different poets, I'm not certain about who assembled the anthology itself.
\item[\textbf{Claude Sonnet 4 (think)}] \textcolor{green!60!black}{\textsc{correct}} --- The Hyakunin Isshu anthology was compiled by Fujiwara no Teika (also known as Fujiwara no Sadaie) around 1235. He selected one poem each from 100 different poets, creating this famous collection of classical Japanese wak...
\end{description}

\subsection{Thinking Mode Correcting Hallucination}
\label{app:case-thinking-correct}

Thinking mode not only helps with recall but can also correct hallucinations. Here, the base DeepSeek-V3.2 answers 1529 (the first Ottoman siege) instead of the gold answer 1683 (the second, more famous siege). While both sieges were technically ``repelled,'' the probe targets the 1683 Battle of Vienna---a more common reference in training data. The thinking variant correctly disambiguates between the two events.

\begin{quote}
\textbf{Probe:} In what year was the Siege of Vienna by the Ottomans that was repelled? \\
\textbf{Tier:} T1 \quad \textbf{Gold answer:} 1683 \\
\textbf{Domain:} Historical Dates \\
\end{quote}

\begin{description}[style=nextline, leftmargin=1.5cm]
\item[\textbf{DeepSeek-V3.2}] \textcolor{red!70!black}{\textsc{wrong}} --- The Siege of Vienna by the Ottomans that was successfully repelled occurred in 1529.
\item[\textbf{DeepSeek-V3.2 (think)}] \textcolor{green!60!black}{\textsc{correct}} --- The Siege of Vienna by the Ottomans that was repelled occurred in 1683.
\end{description}

\section{Ethical Considerations}
\label{app:ethics}

This work estimates proprietary model sizes from public API access, which raises considerations for competitive intelligence and intellectual property. We note that: (1)~parameter count is only one dimension of model capability, and our estimates are effective capacity (not literal architecture); (2)~the method provides ${\sim}3.2\times$ precision, insufficient for reverse-engineering proprietary designs; and (3)~similar estimates are already publicly available from inference-economics analyses~\citep{epochai2024}. Knowledge fingerprinting for cross-model lineage analysis could be used to enforce licensing terms (a legitimate use) or to make unfounded IP claims (a misuse). We encourage responsible use of these tools for transparency and governance rather than competitive surveillance.

All information presented in this paper is drawn from publicly available sources. The author has no internal access to, confidential briefings about, or non-public collaboration with any frontier LLM vendor; in particular, the parameter and effective-capacity estimates for proprietary models are produced \emph{entirely} from external behavioral measurements and the open-weight calibration set.

%% file: tables/lambda_sensitivity.tex
\begin{tabular}{rccccrrrr}
\toprule
$\lambda$ & Slope & $R^2$ & LOO$\times$ & PI$\times$ & GPT-5.5 Pro & Claude Fable 5 & GPT-4.1 & Claude Opus 4.7 \\
 & (pp/dec) & & (med.) & (90\%) & est. & est. & est. & est. \\
\midrule
$0.00$\;$^\star$ & 15.9 & 0.910 & 1.48 & 3.20 & 5.3T & 3.5T & 2.2T & 538B \\
$-0.25$ & 16.7 & 0.936 & 1.48 & 2.64 & 4.6T & 4.3T & 1.6T & 761B \\
$-0.50$ & 16.6 & 0.935 & 1.57 & 2.65 & 5.2T & 4.9T & 1.5T & 916B \\
$-1.00$ & 16.5 & 0.934 & 1.65 & 2.68 & 5.7T & 4.8T & 1.2T & 1.2T \\
$-2.00$ & 16.0 & 0.920 & 1.68 & 2.99 & 5.4T & 5.8T & 1.3T & 1.5T \\
\bottomrule
\end{tabular}

%% file: tables/lambda_floor_ablation.tex
\begin{tabular}{rlrrrrrr}
\toprule
$\lambda$ & floor & $R^2$ & Slope & PI$\times$ & LOO$\times$ & GPT-5.5 & Gemini 3.1 Pro \\
 & & & (pp/dec) & & & est. & est. \\
\midrule
$0.00$$^\star$ & on & 0.910 & 15.9 & 3.20 & 1.48 & 4.7T & 7.3T \\
\multicolumn{8}{l}{\footnotesize\quad(at $\lambda=0$ the floor is a no-op: scores are $\text{correct}/\text{total}\geq 0$)} \\
$0.00$ & off & 0.910 & 15.9 & 3.20 & 1.48 & 4.7T & 7.3T \\
$-0.25$ & on & 0.936 & 16.7 & 2.64 & 1.48 & 4.1T & 9.4T \\
$-0.25$ & off & 0.931 & 19.3 & 2.74 & 1.56 & 3.0T & 6.2T \\
$-0.50$ & on & 0.935 & 16.6 & 2.65 & 1.57 & 4.6T & 12.0T \\
$-0.50$ & off & 0.905 & 22.7 & 3.32 & 1.55 & 2.2T & 5.5T \\
$-1.00$ & on & 0.934 & 16.5 & 2.68 & 1.65 & 4.9T & 18.0T \\
$-1.00$ & off & 0.830 & 29.5 & 5.35 & 1.77 & 1.5T & 4.8T \\
$-1.50$ & on & 0.927 & 16.3 & 2.83 & 1.63 & 4.8T & 25.4T \\
$-1.50$ & off & 0.768 & 36.4 & 7.66 & 2.12 & 1.2T & 4.3T \\
$-2.00$ & on & 0.920 & 16.0 & 2.99 & 1.68 & 4.4T & 34.5T \\
$-2.00$ & off & 0.723 & 43.2 & 9.91 & 2.40 & 972B & 4.1T \\
\bottomrule
\end{tabular}

%% file: tables/moe_dense_fits.tex
\begin{tabular}{lrrrr}
\toprule
Calibration subset & $n$ & Slope & Intercept & $R^2$ \\
\midrule
Dense & 52 & 0.151 & 0.245 & 0.875 \\
MoE (total params) & 41 & 0.145 & 0.282 & 0.667 \\
MoE (active params) & 41 & 0.144 & 0.453 & 0.412 \\
Combined (headline) & 93 & 0.159 & 0.242 & 0.910 \\
\bottomrule
\end{tabular}

%% file: tables/moe_dense_sens.tex
\begin{tabular}{lrrr}
\toprule
Model & Accuracy & Combined & MoE-curve \\
\midrule
Claude Fable 5 & 80.5\% & ${\sim}3.5$T & ${\sim}4.0$T \\
GPT-5.5 Pro & 83.4\% & ${\sim}5.3$T & ${\sim}6.3$T \\
GPT-5.5 & 82.6\% & ${\sim}4.7$T & ${\sim}5.5$T \\
Gemini 2.5 Pro & 79.5\% & ${\sim}3.0$T & ${\sim}3.4$T \\
GPT-4.1 & 77.2\% & ${\sim}2.2$T & ${\sim}2.4$T \\
Claude Opus 4.7 & 67.6\% & ${\sim}538$B & ${\sim}515$B \\
\bottomrule
\end{tabular}

%% file: tables/full_accuracy.tex
\small
\setlength{\tabcolsep}{4pt}
\begin{longtable}{@{}ll r r l c rr rrrrrrr@{}}
\caption{Full model results: per-tier accuracy at $\lambda=0$ (no penalty). Models grouped by vendor, sorted by accuracy. \textbf{Acc.}\ is the tier-mean accuracy used throughout; \textbf{Raw} is overall correct/total. The \textbf{IKP Pred.} column shows the calibration's predicted parameter count and ratio to actual for open-weight models; ratios $>2\times$ or $<0.5\times$ are bolded as systematic outliers. Proprietary models show `--'.}
\label{tab:full-accuracy} \\
\toprule
\textbf{Vendor} & \textbf{Model} & \textbf{Params} & \textbf{IKP Pred.} & \textbf{Arch} & \textbf{Think} & \textbf{Acc.} & \textbf{Raw} & \textbf{T1} & \textbf{T2} & \textbf{T3} & \textbf{T4} & \textbf{T5} & \textbf{T6} & \textbf{T7} \\
\midrule
\endfirsthead
\toprule
\textbf{Vendor} & \textbf{Model} & \textbf{Params} & \textbf{IKP Pred.} & \textbf{Arch} & \textbf{Think} & \textbf{Acc.} & \textbf{Raw} & \textbf{T1} & \textbf{T2} & \textbf{T3} & \textbf{T4} & \textbf{T5} & \textbf{T6} & \textbf{T7} \\
\midrule
\endhead
\bottomrule
\endfoot
Google & gemini-3-flash-think & -- & -- & -- & Y & 0.862 & 0.870 & 1.00 & 0.99 & 0.99 & 0.96 & 0.97 & 0.82 & 0.28 \\
 & gemini-3.1-pro & -- & -- & -- &  & 0.856 & 0.867 & 1.00 & 0.99 & 1.00 & 0.98 & 0.97 & 0.94 & 0.11 \\
 & gemini-3-flash & -- & -- & -- &  & 0.851 & 0.860 & 1.00 & 1.00 & 0.99 & 0.96 & 0.97 & 0.80 & 0.24 \\
 & gemini-2.5-pro & -- & -- & -- &  & 0.795 & 0.805 & 0.99 & 1.00 & 0.99 & 0.95 & 0.82 & 0.58 & 0.22 \\
 & gemini-2.5-pro-think & -- & -- & -- & Y & 0.789 & 0.799 & 0.99 & 1.00 & 0.99 & 0.94 & 0.83 & 0.55 & 0.22 \\
 & gemini-3.1-flash-lite & -- & -- & -- &  & 0.768 & 0.779 & 0.99 & 0.99 & 0.98 & 0.93 & 0.87 & 0.50 & 0.13 \\
 & gemini-2.5-flash-think & -- & -- & -- & Y & 0.702 & 0.714 & 1.00 & 0.99 & 0.95 & 0.84 & 0.66 & 0.33 & 0.15 \\
 & gemini-2.0-flash & -- & -- & -- &  & 0.663 & 0.675 & 0.99 & 0.98 & 0.95 & 0.81 & 0.47 & 0.26 & 0.17 \\
 & gemini-2.5-flash & -- & -- & -- &  & 0.621 & 0.635 & 0.99 & 0.99 & 0.90 & 0.80 & 0.39 & 0.19 & 0.08 \\
 & gemini-2.5-flash-lite-think & -- & -- & -- & Y & 0.599 & 0.613 & 1.00 & 0.99 & 0.90 & 0.75 & 0.37 & 0.12 & 0.06 \\
 & gemini-2.5-flash-lite & -- & -- & -- &  & 0.557 & 0.571 & 0.99 & 0.98 & 0.87 & 0.62 & 0.30 & 0.10 & 0.03 \\
 & gemma-4-31b & 31B & 27B (0.9$\times$) & dense &  & 0.470 & 0.486 & 0.99 & 0.99 & 0.68 & 0.34 & 0.13 & 0.09 & 0.08 \\
 & gemma-3-27b & 27B & 25B (0.9$\times$) & dense &  & 0.465 & 0.481 & 0.98 & 0.98 & 0.64 & 0.36 & 0.15 & 0.06 & 0.08 \\
 & gemma-4-26b-a4b & 26B & 21B (0.8$\times$) & moe &  & 0.452 & 0.468 & 0.99 & 0.98 & 0.62 & 0.35 & 0.10 & 0.07 & 0.05 \\
 & gemma-2-27b & 27B & 15B (0.5$\times$) & dense &  & 0.428 & 0.486 & 0.99 & 0.95 & 0.65 & 0.32 & 0.07 & 0.01 & 0.00 \\
 & gemma-3-12b & 12B & 10B (0.8$\times$) & dense &  & 0.400 & 0.416 & 0.97 & 0.94 & 0.48 & 0.24 & 0.07 & 0.05 & 0.06 \\
 & gemma-3n-e4b & 4B & 5B (1.2$\times$) & dense &  & 0.348 & 0.393 & 0.99 & 0.88 & 0.34 & 0.09 & 0.05 & 0.03 & 0.06 \\
 & gemma-3-4b & 4B & 4B (0.9$\times$) & dense &  & 0.330 & 0.345 & 0.98 & 0.79 & 0.26 & 0.07 & 0.06 & 0.03 & 0.13 \\
 & gemma-2-2b & 3B & 3B (1.2$\times$) & dense &  & 0.320 & 0.333 & 0.94 & 0.72 & 0.17 & 0.11 & 0.10 & 0.07 & 0.12 \\
 & gemma-3-1b & 1B & 572M (0.6$\times$) & dense &  & 0.204 & 0.216 & 0.81 & 0.48 & 0.09 & 0.01 & 0.01 & 0.01 & 0.01 \\
 & gemma-3-270m & 268M & 237M (0.9$\times$) & dense &  & 0.143 & 0.145 & 0.30 & 0.15 & 0.08 & 0.09 & 0.17 & 0.08 & 0.12 \\
OpenAI & gpt-5.5-pro & -- & -- & -- & Y & 0.834 & 0.844 & 1.00 & 1.00 & 0.98 & 0.98 & 0.94 & 0.72 & 0.22 \\
 & gpt-5.5 & -- & -- & -- &  & 0.826 & 0.835 & 1.00 & 1.00 & 0.99 & 0.96 & 0.95 & 0.67 & 0.20 \\
 & gpt-5.5-think & -- & -- & -- & Y & 0.826 & 0.835 & 1.00 & 1.00 & 0.98 & 0.96 & 0.93 & 0.70 & 0.21 \\
 & gpt-5.4-pro & -- & -- & -- &  & 0.778 & 0.789 & 1.00 & 0.99 & 0.99 & 0.94 & 0.86 & 0.49 & 0.18 \\
 & gpt-4.1 & -- & -- & -- &  & 0.772 & 0.782 & 1.00 & 0.98 & 0.99 & 0.92 & 0.82 & 0.47 & 0.21 \\
 & gpt-5-pro & -- & -- & -- &  & 0.770 & 0.780 & 1.00 & 0.99 & 0.99 & 0.94 & 0.80 & 0.48 & 0.19 \\
 & o3 & -- & -- & -- &  & 0.769 & 0.780 & 1.00 & 0.99 & 0.99 & 0.91 & 0.82 & 0.48 & 0.18 \\
 & gpt-5-think & -- & -- & -- & Y & 0.765 & 0.776 & 0.99 & 0.99 & 0.98 & 0.93 & 0.83 & 0.46 & 0.17 \\
 & gpt-5 & -- & -- & -- &  & 0.758 & 0.769 & 0.99 & 0.99 & 0.99 & 0.92 & 0.80 & 0.44 & 0.17 \\
 & o1 & -- & -- & -- & Y & 0.736 & 0.748 & 0.99 & 1.00 & 0.98 & 0.94 & 0.77 & 0.32 & 0.15 \\
 & gpt-5.4 & -- & -- & -- &  & 0.723 & 0.735 & 1.00 & 0.98 & 0.97 & 0.91 & 0.72 & 0.34 & 0.13 \\
 & gpt-5.3 & -- & -- & -- &  & 0.722 & 0.734 & 1.00 & 0.99 & 0.99 & 0.90 & 0.74 & 0.29 & 0.13 \\
 & gpt-4-turbo & -- & -- & -- &  & 0.722 & 0.733 & 0.99 & 1.00 & 0.98 & 0.91 & 0.78 & 0.27 & 0.12 \\
 & gpt-5.2-pro & -- & -- & -- &  & 0.688 & 0.700 & 0.99 & 0.99 & 0.97 & 0.90 & 0.60 & 0.26 & 0.10 \\
 & gpt-5.1 & -- & -- & -- &  & 0.687 & 0.699 & 0.99 & 0.98 & 0.97 & 0.86 & 0.66 & 0.27 & 0.08 \\
 & gpt-4 & 1.8T & -- & moe &  & 0.686 & 0.698 & 0.99 & 0.99 & 0.97 & 0.82 & 0.67 & 0.26 & 0.09 \\
 & gpt-5.2 & -- & -- & -- &  & 0.680 & 0.693 & 0.99 & 0.99 & 0.98 & 0.87 & 0.60 & 0.23 & 0.09 \\
 & gpt-4.1-mini & -- & -- & -- &  & 0.671 & 0.683 & 0.99 & 0.99 & 0.94 & 0.81 & 0.54 & 0.25 & 0.17 \\
 & gpt-4o & -- & -- & -- &  & 0.654 & 0.667 & 0.99 & 0.98 & 0.91 & 0.80 & 0.56 & 0.25 & 0.09 \\
 & o4-mini-think & -- & -- & -- & Y & 0.650 & 0.663 & 1.00 & 0.99 & 0.96 & 0.82 & 0.43 & 0.21 & 0.13 \\
 & gpt-5.4-mini & -- & -- & -- &  & 0.644 & 0.657 & 1.00 & 0.99 & 0.91 & 0.82 & 0.53 & 0.17 & 0.09 \\
 & gpt-3.5-turbo & -- & -- & -- &  & 0.622 & 0.635 & 0.99 & 0.99 & 0.94 & 0.75 & 0.48 & 0.13 & 0.08 \\
 & gpt-oss-120b-think & 120B & 106B (0.9$\times$) & dense & Y & 0.564 & 0.578 & 0.97 & 0.98 & 0.85 & 0.63 & 0.31 & 0.12 & 0.09 \\
 & gpt-5-mini-think & -- & -- & -- & Y & 0.563 & 0.578 & 0.99 & 0.99 & 0.92 & 0.64 & 0.28 & 0.07 & 0.05 \\
 & gpt-5-mini & -- & -- & -- &  & 0.555 & 0.571 & 0.99 & 0.99 & 0.89 & 0.65 & 0.25 & 0.07 & 0.04 \\
 & gpt-4o-mini & -- & -- & -- &  & 0.554 & 0.569 & 0.98 & 0.98 & 0.89 & 0.65 & 0.25 & 0.07 & 0.05 \\
 & gpt-4.1-nano & -- & -- & -- &  & 0.536 & 0.551 & 1.00 & 0.96 & 0.82 & 0.54 & 0.28 & 0.08 & 0.06 \\
 & o3-mini & -- & -- & -- &  & 0.527 & 0.543 & 0.99 & 0.99 & 0.89 & 0.60 & 0.12 & 0.05 & 0.04 \\
 & gpt-5-nano-think & -- & -- & -- & Y & 0.466 & 0.483 & 1.00 & 0.99 & 0.76 & 0.43 & 0.06 & 0.01 & 0.01 \\
 & gpt-5-nano & -- & -- & -- &  & 0.459 & 0.475 & 0.99 & 0.98 & 0.75 & 0.41 & 0.05 & 0.01 & 0.01 \\
 & gpt-oss-20b-think & 20B & 16B (0.8$\times$) & dense & Y & 0.435 & 0.451 & 0.99 & 0.95 & 0.64 & 0.26 & 0.07 & 0.06 & 0.06 \\
 & gpt-5.4-nano & -- & -- & -- &  & 0.400 & 0.416 & 0.95 & 0.93 & 0.59 & 0.25 & 0.06 & 0.01 & 0.01 \\
Anthropic & claude-fable-5 & -- & -- & -- &  & 0.805 & 0.815 & 0.99 & 0.99 & 0.99 & 0.94 & 0.91 & 0.60 & 0.20 \\
 & claude-opus-4.6-think & -- & -- & -- & Y & 0.751 & 0.763 & 1.00 & 1.00 & 0.99 & 0.97 & 0.81 & 0.38 & 0.12 \\
 & claude-opus-4.6 & -- & -- & -- &  & 0.735 & 0.747 & 1.00 & 0.99 & 0.99 & 0.96 & 0.77 & 0.33 & 0.10 \\
 & claude-3.7-sonnet & -- & -- & -- &  & 0.726 & 0.738 & 0.99 & 1.00 & 0.97 & 0.87 & 0.77 & 0.37 & 0.11 \\
 & claude-opus-4.5 & -- & -- & -- &  & 0.711 & 0.722 & 0.99 & 0.99 & 0.97 & 0.89 & 0.72 & 0.29 & 0.12 \\
 & claude-opus-4.1-think & -- & -- & -- & Y & 0.707 & 0.719 & 0.99 & 1.00 & 0.98 & 0.93 & 0.71 & 0.25 & 0.09 \\
 & claude-opus-4.5-think & -- & -- & -- & Y & 0.706 & 0.718 & 1.00 & 1.00 & 0.99 & 0.93 & 0.70 & 0.23 & 0.09 \\
 & claude-sonnet-4.6-think & -- & -- & -- & Y & 0.700 & 0.712 & 0.99 & 0.99 & 0.96 & 0.93 & 0.71 & 0.22 & 0.09 \\
 & claude-sonnet-4.6 & -- & -- & -- &  & 0.700 & 0.712 & 1.00 & 1.00 & 0.96 & 0.93 & 0.70 & 0.23 & 0.09 \\
 & claude-opus-4.7-think & -- & -- & -- & Y & 0.698 & 0.710 & 1.00 & 0.99 & 0.99 & 0.94 & 0.65 & 0.24 & 0.08 \\
 & claude-sonnet-5 & -- & -- & -- &  & 0.682 & 0.694 & 1.00 & 0.99 & 0.99 & 0.89 & 0.66 & 0.17 & 0.07 \\
 & claude-opus-4.7 & -- & -- & -- &  & 0.676 & 0.689 & 1.00 & 0.99 & 0.97 & 0.91 & 0.59 & 0.20 & 0.07 \\
 & claude-sonnet-4.5-think & -- & -- & -- & Y & 0.662 & 0.675 & 0.99 & 1.00 & 0.98 & 0.91 & 0.56 & 0.13 & 0.05 \\
 & claude-opus-4.8 & -- & -- & -- &  & 0.619 & 0.633 & 0.99 & 1.00 & 0.94 & 0.81 & 0.41 & 0.13 & 0.05 \\
 & claude-3.7-sonnet-think & -- & -- & -- & Y & 0.616 & 0.629 & 0.99 & 0.99 & 0.91 & 0.77 & 0.48 & 0.12 & 0.06 \\
 & claude-opus-4-think & -- & -- & -- & Y & 0.611 & 0.625 & 0.99 & 0.99 & 0.97 & 0.75 & 0.40 & 0.12 & 0.04 \\
 & claude-sonnet-4-think & -- & -- & -- & Y & 0.574 & 0.589 & 0.99 & 1.00 & 0.89 & 0.70 & 0.34 & 0.06 & 0.03 \\
 & claude-sonnet-4.5 & -- & -- & -- &  & 0.568 & 0.583 & 0.99 & 1.00 & 0.90 & 0.71 & 0.28 & 0.06 & 0.03 \\
 & claude-opus-4.1 & -- & -- & -- &  & 0.561 & 0.576 & 0.99 & 1.00 & 0.89 & 0.66 & 0.25 & 0.07 & 0.06 \\
 & claude-3.5-haiku & -- & -- & -- &  & 0.543 & 0.605 & 0.99 & 0.99 & 0.80 & 0.63 & 0.30 & 0.07 & 0.02 \\
 & claude-sonnet-4 & -- & -- & -- &  & 0.511 & 0.527 & 0.99 & 0.99 & 0.85 & 0.56 & 0.15 & 0.02 & 0.02 \\
 & claude-opus-4 & -- & -- & -- &  & 0.511 & 0.526 & 0.99 & 0.99 & 0.83 & 0.55 & 0.14 & 0.03 & 0.04 \\
 & claude-haiku-4.5-think & -- & -- & -- & Y & 0.508 & 0.524 & 0.98 & 0.99 & 0.84 & 0.54 & 0.13 & 0.04 & 0.03 \\
 & claude-haiku-4.5 & -- & -- & -- &  & 0.471 & 0.487 & 0.99 & 0.98 & 0.74 & 0.46 & 0.08 & 0.02 & 0.01 \\
 & claude-3-haiku & -- & -- & -- &  & 0.404 & 0.459 & 0.99 & 0.96 & 0.58 & 0.24 & 0.03 & 0.01 & 0.01 \\
DeepSeek & deepseek-v4-pro-think & 1.6T & 2.2T (1.4$\times$) & moe & Y & 0.773 & 0.808 & 1.00 & 0.99 & 0.97 & 0.96 & 0.84 & 0.46 & 0.18 \\
 & deepseek-v4-flash & 284B & \textbf{1.5T (5.4$\times$)} & moe &  & 0.748 & 0.759 & 0.99 & 1.00 & 0.99 & 0.90 & 0.78 & 0.44 & 0.13 \\
 & deepseek-v4-flash-think & 284B & \textbf{1.2T (4.1$\times$)} & moe & Y & 0.730 & 0.741 & 1.00 & 0.99 & 0.99 & 0.94 & 0.71 & 0.36 & 0.11 \\
 & deepseek-r1-think & 671B & 793B (1.2$\times$) & moe & Y & 0.703 & 0.757 & 1.00 & 0.97 & 0.96 & 0.87 & 0.68 & 0.25 & 0.19 \\
 & deepseek-v3.2-think & 671B & 663B (1.0$\times$) & moe & Y & 0.690 & 0.703 & 1.00 & 0.99 & 0.97 & 0.86 & 0.66 & 0.24 & 0.10 \\
 & deepseek-v3 & 671B & 642B (1.0$\times$) & moe &  & 0.688 & 0.720 & 0.99 & 0.99 & 0.96 & 0.89 & 0.64 & 0.23 & 0.10 \\
 & deepseek-v3.2 & 671B & 422B (0.6$\times$) & moe &  & 0.659 & 0.672 & 0.99 & 1.00 & 0.96 & 0.83 & 0.61 & 0.17 & 0.06 \\
 & deepseek-v4-pro & 1.6T & \textbf{181B (0.1$\times$)} & moe &  & 0.601 & 0.648 & 0.99 & 0.98 & 0.90 & 0.77 & 0.34 & 0.16 & 0.06 \\
 & deepseek-v3.1 & -- & -- & -- &  & 0.589 & 0.603 & 0.99 & 0.99 & 0.89 & 0.72 & 0.35 & 0.13 & 0.04 \\
 & deepseek-r1-distill-llama-70b-think & 70B & 119B (1.7$\times$) & dense & Y & 0.572 & 0.586 & 0.98 & 0.96 & 0.85 & 0.66 & 0.36 & 0.14 & 0.04 \\
 & deepseek-r1-distill-qwen-32b-think & 32B & 16B (0.5$\times$) & dense & Y & 0.435 & 0.488 & 0.98 & 0.95 & 0.70 & 0.24 & 0.08 & 0.03 & 0.05 \\
xAI & grok-3 & -- & -- & -- &  & 0.768 & 0.826 & 1.00 & 1.00 & 0.98 & 0.94 & 0.86 & 0.47 & 0.12 \\
 & grok-4 & -- & -- & -- &  & 0.766 & 0.821 & 1.00 & 0.99 & 0.97 & 0.94 & 0.88 & 0.41 & 0.17 \\
 & grok-4.20 & -- & -- & -- &  & 0.693 & 0.743 & 1.00 & 0.97 & 0.91 & 0.76 & 0.68 & 0.31 & 0.22 \\
 & grok-4.20-think & -- & -- & -- & Y & 0.665 & 0.727 & 1.00 & 0.99 & 0.97 & 0.83 & 0.61 & 0.19 & 0.08 \\
 & grok-3-mini-think & -- & -- & -- & Y & 0.608 & 0.672 & 1.00 & 0.97 & 0.95 & 0.77 & 0.42 & 0.12 & 0.03 \\
Xiaomi & mimo-v2.5-pro & -- & -- & -- &  & 0.744 & 0.755 & 1.00 & 0.99 & 0.99 & 0.95 & 0.84 & 0.33 & 0.10 \\
 & mimo-v2.5-pro-think & -- & -- & -- & Y & 0.742 & 0.754 & 1.00 & 0.99 & 0.99 & 0.94 & 0.84 & 0.33 & 0.10 \\
 & mimo-v2.5 & -- & -- & -- &  & 0.695 & 0.706 & 1.00 & 0.98 & 0.98 & 0.88 & 0.67 & 0.22 & 0.12 \\
 & mimo-v2.5-think & -- & -- & -- & Y & 0.684 & 0.696 & 1.00 & 0.99 & 0.99 & 0.85 & 0.67 & 0.21 & 0.08 \\
 & mimo-v2-flash & 309B & 348B (1.1$\times$) & moe &  & 0.646 & 0.658 & 0.99 & 0.96 & 0.92 & 0.81 & 0.56 & 0.16 & 0.11 \\
 & mimo-v2-flash-think & 309B & 339B (1.1$\times$) & moe & Y & 0.644 & 0.657 & 1.00 & 0.99 & 0.93 & 0.80 & 0.53 & 0.14 & 0.11 \\
 & mimo-v2-pro & -- & -- & -- &  & 0.635 & 0.648 & 0.99 & 1.00 & 0.96 & 0.82 & 0.49 & 0.14 & 0.04 \\
 & mimo-v2-pro-think & -- & -- & -- & Y & 0.624 & 0.638 & 0.99 & 0.99 & 0.94 & 0.82 & 0.46 & 0.13 & 0.03 \\
Moonshot & kimi-k2.5-think & 1.0T & 1.4T (1.4$\times$) & moe & Y & 0.742 & 0.753 & 0.99 & 0.99 & 0.99 & 0.97 & 0.92 & 0.21 & 0.12 \\
 & kimi-k2.7-code & 1.0T & 1.1T (1.0$\times$) & moe &  & 0.724 & 0.735 & 0.99 & 1.00 & 0.97 & 0.91 & 0.81 & 0.23 & 0.15 \\
 & kimi-k2.6-think & 1.0T & 909B (0.9$\times$) & moe & Y & 0.712 & 0.724 & 0.99 & 1.00 & 1.00 & 0.94 & 0.83 & 0.16 & 0.06 \\
 & kimi-k2 & 1.0T & 682B (0.7$\times$) & moe &  & 0.692 & 0.751 & 0.99 & 0.99 & 0.97 & 0.85 & 0.69 & 0.22 & 0.12 \\
Tencent & hy3-preview & -- & -- & -- &  & 0.741 & 0.752 & 1.00 & 0.99 & 0.98 & 0.90 & 0.80 & 0.36 & 0.15 \\
 & hunyuan-a13b-think & 80B & \textbf{6B (0.1$\times$)} & moe & Y & 0.368 & 0.384 & 0.96 & 0.85 & 0.44 & 0.16 & 0.06 & 0.04 & 0.06 \\
 & hunyuan-a13b & 80B & \textbf{4B (0.0$\times$)} & moe &  & 0.331 & 0.347 & 0.95 & 0.85 & 0.33 & 0.10 & 0.02 & 0.03 & 0.02 \\
Z.ai & glm-5 & 744B & 1.3T (1.8$\times$) & moe &  & 0.737 & 0.748 & 0.99 & 0.99 & 0.97 & 0.89 & 0.82 & 0.30 & 0.19 \\
 & glm-5-think & 744B & 1.3T (1.7$\times$) & moe & Y & 0.735 & 0.746 & 0.99 & 0.99 & 0.98 & 0.91 & 0.77 & 0.34 & 0.16 \\
 & glm-5.1-think & 754B & 1.2T (1.5$\times$) & moe & Y & 0.729 & 0.740 & 0.96 & 0.98 & 0.99 & 0.94 & 0.78 & 0.33 & 0.12 \\
 & glm-4.7-think & 358B & \textbf{1.1T (3.0$\times$)} & moe & Y & 0.723 & 0.735 & 0.99 & 0.99 & 0.98 & 0.92 & 0.71 & 0.33 & 0.15 \\
 & glm-4.6-think & 357B & \textbf{1.0T (2.9$\times$)} & moe & Y & 0.722 & 0.733 & 0.99 & 0.98 & 0.96 & 0.93 & 0.71 & 0.33 & 0.15 \\
 & glm-5.2 & -- & -- & -- &  & 0.722 & 0.733 & 1.00 & 1.00 & 1.00 & 0.88 & 0.75 & 0.29 & 0.13 \\
 & glm-5.1 & 754B & 1.0T (1.4$\times$) & moe &  & 0.721 & 0.732 & 1.00 & 0.99 & 0.98 & 0.91 & 0.76 & 0.27 & 0.12 \\
 & glm-5-turbo & -- & -- & -- &  & 0.711 & 0.723 & 1.00 & 0.98 & 0.98 & 0.92 & 0.69 & 0.28 & 0.11 \\
 & glm-5-turbo-think & -- & -- & -- & Y & 0.710 & 0.769 & 0.99 & 0.99 & 0.98 & 0.88 & 0.70 & 0.31 & 0.11 \\
 & glm-4.5-think & 355B & 479B (1.3$\times$) & moe & Y & 0.668 & 0.726 & 0.99 & 0.99 & 0.96 & 0.80 & 0.60 & 0.21 & 0.12 \\
 & glm-4.5-air-think & 106B & 139B (1.3$\times$) & moe & Y & 0.583 & 0.597 & 0.99 & 0.99 & 0.90 & 0.70 & 0.34 & 0.11 & 0.04 \\
 & glm-4.7-flash-think & 30B & \textbf{75B (2.5$\times$)} & moe & Y & 0.540 & 0.555 & 0.98 & 0.99 & 0.90 & 0.54 & 0.22 & 0.06 & 0.08 \\
 & glm-4-32b & 32B & 46B (1.4$\times$) & dense &  & 0.506 & 0.565 & 0.97 & 0.98 & 0.80 & 0.52 & 0.19 & 0.03 & 0.05 \\
Alibaba & qwen3.7-max & -- & -- & -- &  & 0.710 & 0.723 & 1.00 & 1.00 & 0.97 & 0.90 & 0.71 & 0.29 & 0.10 \\
 & qwen3-max & -- & -- & -- &  & 0.684 & 0.745 & 1.00 & 0.99 & 0.96 & 0.87 & 0.67 & 0.18 & 0.11 \\
 & qwen3.7-plus & -- & -- & -- &  & 0.681 & 0.693 & 1.00 & 1.00 & 0.95 & 0.88 & 0.61 & 0.21 & 0.12 \\
 & qwen3.5-plus-think & -- & -- & -- & Y & 0.663 & 0.720 & 0.99 & 0.99 & 0.96 & 0.84 & 0.54 & 0.17 & 0.15 \\
 & qwen3.5-397b-a17b-think & 397B & 356B (0.9$\times$) & moe & Y & 0.647 & 0.705 & 0.99 & 0.99 & 0.92 & 0.82 & 0.49 & 0.17 & 0.15 \\
 & qwen3.6-plus-think & -- & -- & -- & Y & 0.641 & 0.704 & 0.99 & 1.00 & 0.97 & 0.85 & 0.49 & 0.10 & 0.09 \\
 & qwen-plus & -- & -- & -- &  & 0.629 & 0.694 & 0.98 & 0.99 & 0.94 & 0.85 & 0.48 & 0.12 & 0.04 \\
 & qwen3.5-122b-a10b-think & 122B & 216B (1.8$\times$) & moe & Y & 0.613 & 0.675 & 0.99 & 0.99 & 0.95 & 0.80 & 0.39 & 0.09 & 0.08 \\
 & qwen3-next-80b-a3b & 80B & 106B (1.3$\times$) & moe &  & 0.564 & 0.579 & 0.98 & 0.99 & 0.92 & 0.71 & 0.22 & 0.07 & 0.04 \\
 & qwen3.5-flash-think & -- & -- & -- & Y & 0.548 & 0.606 & 1.00 & 0.99 & 0.89 & 0.57 & 0.22 & 0.07 & 0.10 \\
 & qwen3.5-35b-a3b-think & 35B & \textbf{84B (2.4$\times$)} & moe & Y & 0.548 & 0.603 & 0.99 & 0.97 & 0.90 & 0.55 & 0.21 & 0.05 & 0.15 \\
 & qwen3-235b-a22b-think & 235B & \textbf{83B (0.4$\times$)} & moe & Y & 0.547 & 0.612 & 0.99 & 0.99 & 0.89 & 0.72 & 0.19 & 0.04 & 0.02 \\
 & qwen-max & -- & -- & -- &  & 0.528 & 0.589 & 0.99 & 0.99 & 0.84 & 0.61 & 0.19 & 0.03 & 0.04 \\
 & qwen3.5-27b-think & 27B & \textbf{55B (2.0$\times$)} & dense & Y & 0.518 & 0.575 & 0.99 & 0.97 & 0.84 & 0.49 & 0.17 & 0.08 & 0.08 \\
 & qwen-2.5-72b & 73B & 40B (0.6$\times$) & dense &  & 0.497 & 0.513 & 1.00 & 0.99 & 0.81 & 0.49 & 0.14 & 0.02 & 0.03 \\
 & qwq-32b-think & 32B & 29B (0.9$\times$) & dense &  & 0.475 & 0.490 & 0.99 & 0.97 & 0.76 & 0.40 & 0.13 & 0.04 & 0.03 \\
 & qwen3-32b-think & 32B & 26B (0.8$\times$) & dense & Y & 0.466 & 0.481 & 0.99 & 0.97 & 0.87 & 0.29 & 0.08 & 0.03 & 0.02 \\
 & qwen3-30b-a3b-think & 30B & 20B (0.7$\times$) & moe & Y & 0.448 & 0.465 & 0.99 & 0.98 & 0.76 & 0.33 & 0.06 & 0.01 & 0.02 \\
 & qwen-turbo & -- & -- & -- &  & 0.442 & 0.497 & 0.96 & 0.96 & 0.66 & 0.36 & 0.11 & 0.02 & 0.03 \\
 & qwen3.5-9b-think & 9B & 18B (2.0$\times$) & dense & Y & 0.441 & 0.494 & 0.98 & 0.95 & 0.68 & 0.28 & 0.08 & 0.05 & 0.06 \\
 & qwen3-14b-think & 14B & 16B (1.2$\times$) & dense & Y & 0.436 & 0.452 & 0.97 & 0.97 & 0.72 & 0.29 & 0.04 & 0.02 & 0.03 \\
 & qwen3-8b-think & 8B & 14B (1.7$\times$) & dense & Y & 0.422 & 0.471 & 0.97 & 0.94 & 0.55 & 0.25 & 0.09 & 0.04 & 0.10 \\
 & qwen-2.5-7b & 8B & 4B (0.6$\times$) & dense &  & 0.344 & 0.396 & 0.98 & 0.97 & 0.31 & 0.11 & 0.03 & 0.01 & 0.00 \\
 & qwen-2.5-0.5b & 494M & 349M (0.7$\times$) & dense &  & 0.170 & 0.180 & 0.94 & 0.15 & 0.03 & 0.03 & 0.00 & 0.02 & 0.01 \\
 & qwen3.5-0.8b & 800M & \textbf{231M (0.3$\times$)} & dense &  & 0.141 & 0.213 & 0.46 & 0.42 & 0.10 & 0.00 & 0.00 & 0.00 & 0.00 \\
 & qwen3-0.6b & 596M & \textbf{146M (0.2$\times$)} & dense &  & 0.110 & 0.116 & 0.47 & 0.20 & 0.05 & 0.01 & 0.01 & 0.02 & 0.01 \\
NVIDIA & nemotron-3-ultra & 550B & 771B (1.4$\times$) & moe &  & 0.701 & 0.713 & 1.00 & 1.00 & 0.97 & 0.91 & 0.70 & 0.25 & 0.07 \\
 & nemotron-3-super-120b & 120B & 203B (1.7$\times$) & moe &  & 0.609 & 0.622 & 0.99 & 0.98 & 0.94 & 0.72 & 0.42 & 0.09 & 0.12 \\
 & nemotron-70b & 70B & 86B (1.2$\times$) & dense &  & 0.549 & 0.564 & 0.99 & 0.99 & 0.86 & 0.68 & 0.24 & 0.05 & 0.03 \\
 & nemotron-super-49b-think & 49B & \textbf{24B (0.5$\times$)} & dense & Y & 0.461 & 0.477 & 0.98 & 0.99 & 0.69 & 0.36 & 0.16 & 0.02 & 0.02 \\
 & nemotron-3-nano-30b & 30B & 21B (0.7$\times$) & moe &  & 0.452 & 0.468 & 0.99 & 0.98 & 0.67 & 0.31 & 0.12 & 0.04 & 0.05 \\
 & nemotron-nano-9b-v2 & 9B & \textbf{4B (0.5$\times$)} & dense &  & 0.340 & 0.357 & 0.98 & 0.91 & 0.39 & 0.09 & 0.00 & 0.00 & 0.01 \\
ByteDance & seed-2.0-lite-think & -- & -- & -- & Y & 0.697 & 0.757 & 1.00 & 0.99 & 0.95 & 0.89 & 0.65 & 0.29 & 0.10 \\
 & seed-2.0-mini-think & -- & -- & -- & Y & 0.629 & 0.686 & 0.99 & 0.97 & 0.88 & 0.74 & 0.52 & 0.19 & 0.11 \\
 & seed-1.6-think & -- & -- & -- & Y & 0.600 & 0.664 & 0.99 & 0.99 & 0.94 & 0.76 & 0.38 & 0.10 & 0.03 \\
 & seed-1.6-flash-think & -- & -- & -- & Y & 0.457 & 0.512 & 1.00 & 0.96 & 0.70 & 0.30 & 0.12 & 0.04 & 0.06 \\
StepFun & step-3.7-flash & -- & -- & -- &  & 0.667 & 0.679 & 1.00 & 0.99 & 0.95 & 0.77 & 0.64 & 0.21 & 0.11 \\
 & step-3.5-flash-think & 197B & \textbf{429B (2.2$\times$)} & moe & Y & 0.660 & 0.676 & 1.00 & 0.98 & 0.96 & 0.81 & 0.56 & 0.20 & 0.12 \\
Amazon & nova-premier & -- & -- & -- &  & 0.656 & 0.715 & 1.00 & 0.98 & 0.95 & 0.84 & 0.48 & 0.24 & 0.11 \\
 & nova-pro & -- & -- & -- &  & 0.568 & 0.629 & 0.99 & 0.99 & 0.89 & 0.66 & 0.27 & 0.12 & 0.05 \\
 & nova-micro & -- & -- & -- &  & 0.396 & 0.450 & 0.99 & 0.94 & 0.50 & 0.26 & 0.08 & 0.00 & 0.01 \\
Meta & llama-4-maverick & 402B & 333B (0.8$\times$) & moe &  & 0.643 & 0.656 & 0.99 & 0.99 & 0.90 & 0.80 & 0.51 & 0.21 & 0.09 \\
 & llama-3-70b & 71B & 116B (1.6$\times$) & dense &  & 0.570 & 0.634 & 0.99 & 0.98 & 0.86 & 0.72 & 0.32 & 0.08 & 0.03 \\
 & llama-3.3-70b & 71B & 113B (1.6$\times$) & dense &  & 0.568 & 0.583 & 0.98 & 0.99 & 0.84 & 0.71 & 0.30 & 0.10 & 0.06 \\
 & llama-3.1-70b & 71B & 58B (0.8$\times$) & dense &  & 0.522 & 0.538 & 0.99 & 0.99 & 0.81 & 0.57 & 0.22 & 0.05 & 0.02 \\
 & hermes-4-405b & 405B & \textbf{58B (0.1$\times$)} & dense &  & 0.522 & 0.538 & 0.99 & 0.98 & 0.82 & 0.61 & 0.15 & 0.07 & 0.03 \\
 & llama-4-scout & 109B & \textbf{35B (0.3$\times$)} & moe &  & 0.488 & 0.504 & 0.98 & 0.93 & 0.75 & 0.45 & 0.22 & 0.06 & 0.02 \\
 & hermes-3-405b & 405B & \textbf{22B (0.1$\times$)} & dense &  & 0.456 & 0.472 & 0.99 & 0.98 & 0.69 & 0.40 & 0.07 & 0.03 & 0.01 \\
 & llama-3-8b & 8B & \textbf{17B (2.1$\times$)} & dense &  & 0.438 & 0.455 & 0.98 & 0.96 & 0.61 & 0.36 & 0.12 & 0.02 & 0.02 \\
 & llama-3.1-8b & 8B & 9B (1.1$\times$) & dense &  & 0.392 & 0.409 & 0.97 & 0.96 & 0.51 & 0.26 & 0.03 & 0.01 & 0.01 \\
 & llama-3.2-3b & 3B & 3B (0.9$\times$) & dense &  & 0.318 & 0.366 & 0.95 & 0.88 & 0.31 & 0.07 & 0.02 & 0.00 & 0.00 \\
 & llama-3.2-1b & 1B & 908M (0.7$\times$) & dense &  & 0.236 & 0.273 & 0.87 & 0.60 & 0.13 & 0.05 & 0.00 & 0.00 & 0.00 \\
MiniMax & minimax-m3 & -- & -- & -- &  & 0.639 & 0.652 & 1.00 & 0.99 & 0.96 & 0.82 & 0.49 & 0.14 & 0.06 \\
 & minimax-m2.7-think & 230B & \textbf{113B (0.5$\times$)} & moe & Y & 0.568 & 0.635 & 0.99 & 0.99 & 0.88 & 0.65 & 0.33 & 0.06 & 0.08 \\
 & minimax-m1-think & 456B & \textbf{16B (0.0$\times$)} & moe & Y & 0.434 & 0.479 & 0.65 & 0.61 & 0.68 & 0.55 & 0.35 & 0.10 & 0.09 \\
Mistral & mistral-medium-3.1 & -- & -- & -- &  & 0.631 & 0.691 & 0.99 & 0.98 & 0.93 & 0.85 & 0.43 & 0.12 & 0.11 \\
 & mixtral-8x22b & 141B & 216B (1.5$\times$) & moe &  & 0.613 & 0.674 & 0.99 & 1.00 & 0.90 & 0.78 & 0.42 & 0.12 & 0.08 \\
 & mistral-large & 123B & 103B (0.8$\times$) & dense &  & 0.562 & 0.624 & 0.98 & 0.98 & 0.90 & 0.69 & 0.28 & 0.06 & 0.04 \\
 & mixtral-8x7b & 47B & 68B (1.5$\times$) & moe &  & 0.533 & 0.548 & 0.96 & 0.98 & 0.82 & 0.60 & 0.25 & 0.06 & 0.06 \\
 & mistral-small-24b & 24B & 36B (1.5$\times$) & dense &  & 0.489 & 0.505 & 0.98 & 0.99 & 0.81 & 0.50 & 0.10 & 0.03 & 0.01 \\
 & ministral-8b & 8B & \textbf{23B (2.9$\times$)} & dense &  & 0.459 & 0.508 & 0.97 & 0.97 & 0.64 & 0.34 & 0.09 & 0.06 & 0.14 \\
 & mistral-nemo-12b & 12B & 15B (1.2$\times$) & dense &  & 0.428 & 0.444 & 0.96 & 0.94 & 0.56 & 0.38 & 0.09 & 0.02 & 0.03 \\
 & mistral-7b & 7B & 7B (1.0$\times$) & dense &  & 0.376 & 0.422 & 0.93 & 0.88 & 0.45 & 0.19 & 0.07 & 0.04 & 0.07 \\
 & ministral-3b & 3B & 6B (2.0$\times$) & dense &  & 0.365 & 0.412 & 0.96 & 0.89 & 0.40 & 0.11 & 0.11 & 0.03 & 0.05 \\
Cohere & command-a & 111B & 215B (1.9$\times$) & dense &  & 0.612 & 0.675 & 0.99 & 0.99 & 0.93 & 0.77 & 0.39 & 0.15 & 0.05 \\
 & command-r-plus & 104B & \textbf{26B (0.2$\times$)} & dense &  & 0.466 & 0.524 & 0.99 & 0.95 & 0.68 & 0.46 & 0.14 & 0.03 & 0.02 \\
 & command-r7b & 7B & 6B (0.8$\times$) & dense &  & 0.364 & 0.410 & 0.96 & 0.84 & 0.42 & 0.19 & 0.03 & 0.03 & 0.08 \\
AI21 & jamba-large & 398B & \textbf{129B (0.3$\times$)} & moe &  & 0.578 & 0.639 & 0.99 & 0.99 & 0.92 & 0.70 & 0.28 & 0.10 & 0.06 \\
Prime-intellect & intellect-3-think & 106B & 87B (0.8$\times$) & moe & Y & 0.550 & 0.564 & 0.98 & 0.96 & 0.82 & 0.58 & 0.26 & 0.09 & 0.16 \\
Baidu & ernie-4.5-300b-a47b & 300B & \textbf{73B (0.2$\times$)} & moe &  & 0.538 & 0.601 & 0.99 & 0.98 & 0.84 & 0.60 & 0.28 & 0.06 & 0.01 \\
Inclusionai & ling-2.6-flash & 104B & 65B (0.6$\times$) & moe &  & 0.530 & 0.751 & 0.99 & 0.99 & 0.87 & 0.57 & 0.30 & 0.00 & 0.00 \\
Microsoft & phi-4 & 15B & 16B (1.1$\times$) & dense &  & 0.432 & 0.448 & 0.97 & 0.96 & 0.62 & 0.26 & 0.10 & 0.05 & 0.06 \\
 & phi-3-mini & 4B & 2B (0.5$\times$) & dense &  & 0.289 & 0.304 & 0.92 & 0.73 & 0.26 & 0.05 & 0.01 & 0.01 & 0.03 \\
AllenAI & olmo-3.1-32b & 32B & \textbf{12B (0.4$\times$)} & dense &  & 0.415 & 0.432 & 0.98 & 0.97 & 0.61 & 0.27 & 0.05 & 0.01 & 0.01 \\
Nex-agi & deepseek-v3.1-nex-n1 & 671B & \textbf{8B (0.0$\times$)} & moe &  & 0.384 & 0.401 & 0.98 & 0.97 & 0.49 & 0.22 & 0.01 & 0.01 & 0.01 \\
HuggingFace & smollm2-1.7b & 2B & 2B (1.4$\times$) & dense &  & 0.301 & 0.317 & 0.95 & 0.80 & 0.17 & 0.06 & 0.06 & 0.04 & 0.03 \\
 & smollm2-360m & 362M & \textbf{2B (5.1$\times$)} & dense &  & 0.284 & 0.294 & 0.82 & 0.54 & 0.13 & 0.11 & 0.13 & 0.08 & 0.18 \\
 & smollm2-135m & 135M & 182M (1.4$\times$) & dense &  & 0.125 & 0.131 & 0.48 & 0.23 & 0.09 & 0.02 & 0.02 & 0.01 & 0.02 \\
Ibm & granite-3.3-2b & 2B & 2B (1.2$\times$) & dense &  & 0.300 & 0.315 & 0.91 & 0.74 & 0.26 & 0.09 & 0.04 & 0.02 & 0.04 \\
\end{longtable}

%% file: tables/full_hallucination.tex
\small
\setlength{\tabcolsep}{4pt}
\begin{longtable}{@{}ll rr rrrrrrr@{}}
\caption{Per-tier hallucination rate = wrong / (wrong + correct + refusal). Higher = more confident incorrect answers. Models sorted as in Table~\ref{tab:full-accuracy}.}
\label{tab:full-hallucination} \\
\toprule
\textbf{Vendor} & \textbf{Model} & \textbf{Acc.} & \textbf{Raw} & \textbf{T1} & \textbf{T2} & \textbf{T3} & \textbf{T4} & \textbf{T5} & \textbf{T6} & \textbf{T7} \\
\midrule
\endfirsthead
\toprule
\textbf{Vendor} & \textbf{Model} & \textbf{Acc.} & \textbf{Raw} & \textbf{T1} & \textbf{T2} & \textbf{T3} & \textbf{T4} & \textbf{T5} & \textbf{T6} & \textbf{T7} \\
\midrule
\endhead
\bottomrule
\endfoot
Google & gemini-3-flash-think & 0.862 & 0.870 & 0.00 & 0.01 & 0.01 & 0.04 & 0.03 & 0.17 & 0.68 \\
 & gemini-3.1-pro & 0.856 & 0.867 & 0.00 & 0.01 & 0.00 & 0.02 & 0.01 & 0.02 & 0.24 \\
 & gemini-3-flash & 0.851 & 0.860 & 0.00 & 0.00 & 0.01 & 0.04 & 0.03 & 0.19 & 0.73 \\
 & gemini-2.5-pro & 0.795 & 0.805 & 0.01 & 0.00 & 0.01 & 0.05 & 0.17 & 0.41 & 0.76 \\
 & gemini-2.5-pro-think & 0.789 & 0.799 & 0.01 & 0.00 & 0.01 & 0.06 & 0.17 & 0.44 & 0.75 \\
 & gemini-3.1-flash-lite & 0.768 & 0.779 & 0.00 & 0.01 & 0.02 & 0.07 & 0.12 & 0.47 & 0.74 \\
 & gemini-2.5-flash-think & 0.702 & 0.714 & 0.00 & 0.01 & 0.05 & 0.16 & 0.31 & 0.56 & 0.58 \\
 & gemini-2.0-flash & 0.663 & 0.675 & 0.01 & 0.01 & 0.05 & 0.18 & 0.44 & 0.70 & 0.74 \\
 & gemini-2.5-flash & 0.621 & 0.635 & 0.01 & 0.01 & 0.07 & 0.13 & 0.15 & 0.18 & 0.19 \\
 & gemini-2.5-flash-lite-think & 0.599 & 0.613 & 0.00 & 0.01 & 0.10 & 0.24 & 0.62 & 0.79 & 0.81 \\
 & gemini-2.5-flash-lite & 0.557 & 0.571 & 0.01 & 0.01 & 0.10 & 0.27 & 0.36 & 0.58 & 0.43 \\
 & gemma-4-31b & 0.470 & 0.486 & 0.00 & 0.01 & 0.30 & 0.59 & 0.74 & 0.61 & 0.72 \\
 & gemma-3-27b & 0.465 & 0.481 & 0.01 & 0.02 & 0.36 & 0.64 & 0.85 & 0.93 & 0.91 \\
 & gemma-4-26b-a4b & 0.452 & 0.468 & 0.01 & 0.01 & 0.34 & 0.53 & 0.65 & 0.55 & 0.52 \\
 & gemma-2-27b & 0.428 & 0.486 & 0.01 & 0.04 & 0.27 & 0.48 & 0.29 & 0.23 & 0.16 \\
 & gemma-3-12b & 0.400 & 0.416 & 0.01 & 0.07 & 0.50 & 0.74 & 0.92 & 0.90 & 0.89 \\
 & gemma-3n-e4b & 0.348 & 0.393 & 0.01 & 0.12 & 0.65 & 0.89 & 0.91 & 0.94 & 0.86 \\
 & gemma-3-4b & 0.330 & 0.345 & 0.02 & 0.21 & 0.73 & 0.91 & 0.90 & 0.91 & 0.82 \\
 & gemma-2-2b & 0.320 & 0.333 & 0.06 & 0.28 & 0.81 & 0.85 & 0.81 & 0.63 & 0.74 \\
 & gemma-3-1b & 0.204 & 0.216 & 0.17 & 0.51 & 0.89 & 0.97 & 0.97 & 0.97 & 0.94 \\
 & gemma-3-270m & 0.143 & 0.145 & 0.29 & 0.35 & 0.51 & 0.42 & 0.27 & 0.26 & 0.46 \\
OpenAI & gpt-5.5-pro & 0.834 & 0.844 & 0.00 & 0.00 & 0.02 & 0.02 & 0.06 & 0.27 & 0.76 \\
 & gpt-5.5 & 0.826 & 0.835 & 0.00 & 0.00 & 0.01 & 0.04 & 0.04 & 0.31 & 0.72 \\
 & gpt-5.5-think & 0.826 & 0.835 & 0.00 & 0.00 & 0.02 & 0.04 & 0.05 & 0.28 & 0.72 \\
 & gpt-5.4-pro & 0.778 & 0.789 & 0.00 & 0.01 & 0.01 & 0.06 & 0.12 & 0.48 & 0.73 \\
 & gpt-4.1 & 0.772 & 0.782 & 0.00 & 0.01 & 0.01 & 0.08 & 0.18 & 0.52 & 0.77 \\
 & gpt-5-pro & 0.770 & 0.780 & 0.00 & 0.01 & 0.01 & 0.04 & 0.08 & 0.31 & 0.35 \\
 & o3 & 0.769 & 0.780 & 0.00 & 0.01 & 0.01 & 0.08 & 0.08 & 0.37 & 0.44 \\
 & gpt-5-think & 0.765 & 0.776 & 0.00 & 0.01 & 0.01 & 0.06 & 0.04 & 0.26 & 0.32 \\
 & gpt-5 & 0.758 & 0.769 & 0.00 & 0.01 & 0.01 & 0.06 & 0.05 & 0.26 & 0.31 \\
 & o1 & 0.736 & 0.748 & 0.01 & 0.00 & 0.02 & 0.05 & 0.05 & 0.14 & 0.18 \\
 & gpt-5.4 & 0.723 & 0.735 & 0.00 & 0.01 & 0.03 & 0.07 & 0.14 & 0.42 & 0.51 \\
 & gpt-5.3 & 0.722 & 0.734 & 0.00 & 0.01 & 0.01 & 0.07 & 0.12 & 0.27 & 0.41 \\
 & gpt-4-turbo & 0.722 & 0.733 & 0.00 & 0.00 & 0.02 & 0.07 & 0.13 & 0.53 & 0.47 \\
 & gpt-5.2-pro & 0.688 & 0.700 & 0.00 & 0.01 & 0.02 & 0.07 & 0.13 & 0.22 & 0.23 \\
 & gpt-5.1 & 0.687 & 0.699 & 0.00 & 0.01 & 0.02 & 0.08 & 0.04 & 0.14 & 0.16 \\
 & gpt-4 & 0.686 & 0.698 & 0.01 & 0.01 & 0.03 & 0.09 & 0.09 & 0.25 & 0.22 \\
 & gpt-5.2 & 0.680 & 0.693 & 0.00 & 0.01 & 0.01 & 0.07 & 0.13 & 0.20 & 0.20 \\
 & gpt-4.1-mini & 0.671 & 0.683 & 0.01 & 0.01 & 0.06 & 0.19 & 0.46 & 0.74 & 0.83 \\
 & gpt-4o & 0.654 & 0.667 & 0.01 & 0.01 & 0.03 & 0.06 & 0.02 & 0.15 & 0.18 \\
 & o4-mini-think & 0.650 & 0.663 & 0.00 & 0.01 & 0.03 & 0.11 & 0.31 & 0.40 & 0.42 \\
 & gpt-5.4-mini & 0.644 & 0.657 & 0.00 & 0.01 & 0.07 & 0.12 & 0.20 & 0.47 & 0.31 \\
 & gpt-3.5-turbo & 0.622 & 0.635 & 0.01 & 0.01 & 0.05 & 0.21 & 0.25 & 0.64 & 0.53 \\
 & gpt-oss-120b-think & 0.564 & 0.578 & 0.02 & 0.01 & 0.14 & 0.36 & 0.47 & 0.60 & 0.55 \\
 & gpt-5-mini-think & 0.563 & 0.578 & 0.00 & 0.01 & 0.01 & 0.04 & 0.04 & 0.04 & 0.06 \\
 & gpt-5-mini & 0.555 & 0.571 & 0.00 & 0.00 & 0.01 & 0.05 & 0.02 & 0.03 & 0.06 \\
 & gpt-4o-mini & 0.554 & 0.569 & 0.01 & 0.02 & 0.10 & 0.32 & 0.48 & 0.71 & 0.60 \\
 & gpt-4.1-nano & 0.536 & 0.551 & 0.00 & 0.04 & 0.16 & 0.43 & 0.55 & 0.73 & 0.69 \\
 & o3-mini & 0.527 & 0.543 & 0.01 & 0.01 & 0.08 & 0.28 & 0.41 & 0.24 & 0.34 \\
 & gpt-5-nano-think & 0.466 & 0.483 & 0.00 & 0.01 & 0.05 & 0.08 & 0.03 & 0.03 & 0.07 \\
 & gpt-5-nano & 0.459 & 0.475 & 0.01 & 0.01 & 0.03 & 0.11 & 0.06 & 0.01 & 0.07 \\
 & gpt-oss-20b-think & 0.435 & 0.451 & 0.01 & 0.04 & 0.35 & 0.69 & 0.89 & 0.82 & 0.73 \\
 & gpt-5.4-nano & 0.400 & 0.416 & 0.01 & 0.03 & 0.12 & 0.24 & 0.11 & 0.14 & 0.15 \\
Anthropic & claude-fable-5 & 0.805 & 0.815 & 0.01 & 0.01 & 0.01 & 0.05 & 0.05 & 0.15 & 0.20 \\
 & claude-opus-4.6-think & 0.751 & 0.763 & 0.00 & 0.00 & 0.01 & 0.03 & 0.09 & 0.26 & 0.28 \\
 & claude-opus-4.6 & 0.735 & 0.747 & 0.00 & 0.00 & 0.01 & 0.04 & 0.07 & 0.29 & 0.28 \\
 & claude-3.7-sonnet & 0.726 & 0.738 & 0.01 & 0.00 & 0.03 & 0.09 & 0.08 & 0.19 & 0.22 \\
 & claude-opus-4.5 & 0.711 & 0.722 & 0.00 & 0.01 & 0.01 & 0.05 & 0.09 & 0.16 & 0.30 \\
 & claude-opus-4.1-think & 0.707 & 0.719 & 0.01 & 0.00 & 0.00 & 0.04 & 0.06 & 0.11 & 0.17 \\
 & claude-opus-4.5-think & 0.706 & 0.718 & 0.00 & 0.00 & 0.01 & 0.04 & 0.06 & 0.11 & 0.27 \\
 & claude-sonnet-4.6-think & 0.700 & 0.712 & 0.00 & 0.01 & 0.04 & 0.07 & 0.16 & 0.36 & 0.33 \\
 & claude-sonnet-4.6 & 0.700 & 0.712 & 0.00 & 0.00 & 0.03 & 0.05 & 0.15 & 0.41 & 0.34 \\
 & claude-opus-4.7-think & 0.698 & 0.710 & 0.00 & 0.01 & 0.01 & 0.03 & 0.07 & 0.17 & 0.16 \\
 & claude-sonnet-5 & 0.682 & 0.694 & 0.00 & 0.01 & 0.01 & 0.09 & 0.12 & 0.26 & 0.17 \\
 & claude-opus-4.7 & 0.676 & 0.689 & 0.00 & 0.01 & 0.02 & 0.03 & 0.05 & 0.10 & 0.15 \\
 & claude-sonnet-4.5-think & 0.662 & 0.675 & 0.00 & 0.00 & 0.02 & 0.04 & 0.09 & 0.15 & 0.16 \\
 & claude-opus-4.8 & 0.619 & 0.633 & 0.01 & 0.00 & 0.00 & 0.03 & 0.04 & 0.07 & 0.10 \\
 & claude-3.7-sonnet-think & 0.616 & 0.629 & 0.01 & 0.01 & 0.01 & 0.05 & 0.02 & 0.04 & 0.08 \\
 & claude-opus-4-think & 0.611 & 0.625 & 0.00 & 0.00 & 0.01 & 0.05 & 0.04 & 0.05 & 0.11 \\
 & claude-sonnet-4-think & 0.574 & 0.589 & 0.00 & 0.00 & 0.04 & 0.06 & 0.02 & 0.05 & 0.08 \\
 & claude-sonnet-4.5 & 0.568 & 0.583 & 0.00 & 0.00 & 0.02 & 0.05 & 0.03 & 0.07 & 0.13 \\
 & claude-opus-4.1 & 0.561 & 0.576 & 0.00 & 0.00 & 0.02 & 0.05 & 0.01 & 0.05 & 0.13 \\
 & claude-3.5-haiku & 0.543 & 0.605 & 0.01 & 0.01 & 0.08 & 0.12 & 0.14 & 0.06 & 0.08 \\
 & claude-sonnet-4 & 0.511 & 0.527 & 0.00 & 0.01 & 0.00 & 0.05 & 0.01 & 0.03 & 0.06 \\
 & claude-opus-4 & 0.511 & 0.526 & 0.01 & 0.01 & 0.01 & 0.04 & 0.01 & 0.04 & 0.07 \\
 & claude-haiku-4.5-think & 0.508 & 0.524 & 0.01 & 0.01 & 0.08 & 0.15 & 0.13 & 0.07 & 0.09 \\
 & claude-haiku-4.5 & 0.471 & 0.487 & 0.01 & 0.01 & 0.09 & 0.13 & 0.11 & 0.12 & 0.09 \\
 & claude-3-haiku & 0.404 & 0.459 & 0.01 & 0.04 & 0.07 & 0.08 & 0.01 & 0.00 & 0.01 \\
DeepSeek & deepseek-v4-pro-think & 0.773 & 0.808 & 0.00 & 0.01 & 0.02 & 0.04 & 0.12 & 0.40 & 0.44 \\
 & deepseek-v4-flash & 0.748 & 0.759 & 0.01 & 0.00 & 0.01 & 0.10 & 0.15 & 0.41 & 0.52 \\
 & deepseek-v4-flash-think & 0.730 & 0.741 & 0.00 & 0.01 & 0.01 & 0.06 & 0.20 & 0.44 & 0.56 \\
 & deepseek-r1-think & 0.703 & 0.757 & 0.00 & 0.03 & 0.03 & 0.08 & 0.15 & 0.31 & 0.33 \\
 & deepseek-v3.2-think & 0.690 & 0.703 & 0.00 & 0.01 & 0.03 & 0.07 & 0.17 & 0.38 & 0.47 \\
 & deepseek-v3 & 0.688 & 0.720 & 0.00 & 0.01 & 0.03 & 0.09 & 0.16 & 0.40 & 0.43 \\
 & deepseek-v3.2 & 0.659 & 0.672 & 0.01 & 0.00 & 0.02 & 0.09 & 0.10 & 0.25 & 0.33 \\
 & deepseek-v4-pro & 0.601 & 0.648 & 0.01 & 0.01 & 0.00 & 0.02 & 0.02 & 0.08 & 0.15 \\
 & deepseek-v3.1 & 0.589 & 0.603 & 0.00 & 0.01 & 0.03 & 0.08 & 0.04 & 0.11 & 0.27 \\
 & deepseek-r1-distill-llama-70b-think & 0.572 & 0.586 & 0.01 & 0.04 & 0.14 & 0.30 & 0.36 & 0.52 & 0.43 \\
 & deepseek-r1-distill-qwen-32b-think & 0.435 & 0.488 & 0.02 & 0.04 & 0.27 & 0.72 & 0.76 & 0.78 & 0.64 \\
xAI & grok-3 & 0.768 & 0.826 & 0.00 & 0.00 & 0.02 & 0.06 & 0.11 & 0.43 & 0.34 \\
 & grok-4 & 0.766 & 0.821 & 0.00 & 0.01 & 0.02 & 0.06 & 0.05 & 0.26 & 0.15 \\
 & grok-4.20 & 0.693 & 0.743 & 0.00 & 0.03 & 0.09 & 0.24 & 0.32 & 0.69 & 0.78 \\
 & grok-4.20-think & 0.665 & 0.727 & 0.00 & 0.01 & 0.02 & 0.10 & 0.13 & 0.20 & 0.11 \\
 & grok-3-mini-think & 0.608 & 0.672 & 0.00 & 0.03 & 0.04 & 0.13 & 0.20 & 0.23 & 0.15 \\
Xiaomi & mimo-v2.5-pro & 0.744 & 0.755 & 0.00 & 0.01 & 0.01 & 0.04 & 0.12 & 0.36 & 0.41 \\
 & mimo-v2.5-pro-think & 0.742 & 0.754 & 0.00 & 0.01 & 0.01 & 0.05 & 0.09 & 0.38 & 0.45 \\
 & mimo-v2.5 & 0.695 & 0.706 & 0.00 & 0.01 & 0.02 & 0.11 & 0.25 & 0.47 & 0.49 \\
 & mimo-v2.5-think & 0.684 & 0.696 & 0.00 & 0.01 & 0.01 & 0.13 & 0.23 & 0.45 & 0.51 \\
 & mimo-v2-flash & 0.646 & 0.658 & 0.01 & 0.04 & 0.08 & 0.18 & 0.36 & 0.65 & 0.62 \\
 & mimo-v2-flash-think & 0.644 & 0.657 & 0.00 & 0.01 & 0.05 & 0.19 & 0.37 & 0.63 & 0.58 \\
 & mimo-v2-pro & 0.635 & 0.648 & 0.01 & 0.00 & 0.01 & 0.07 & 0.08 & 0.05 & 0.17 \\
 & mimo-v2-pro-think & 0.624 & 0.638 & 0.01 & 0.01 & 0.01 & 0.05 & 0.06 & 0.08 & 0.18 \\
Moonshot & kimi-k2.5-think & 0.742 & 0.753 & 0.00 & 0.01 & 0.01 & 0.03 & 0.07 & 0.58 & 0.45 \\
 & kimi-k2.7-code & 0.724 & 0.735 & 0.01 & 0.00 & 0.03 & 0.09 & 0.17 & 0.63 & 0.56 \\
 & kimi-k2.6-think & 0.712 & 0.724 & 0.01 & 0.00 & 0.00 & 0.03 & 0.03 & 0.28 & 0.19 \\
 & kimi-k2 & 0.692 & 0.751 & 0.01 & 0.01 & 0.03 & 0.13 & 0.14 & 0.33 & 0.22 \\
Tencent & hy3-preview & 0.741 & 0.752 & 0.00 & 0.01 & 0.02 & 0.10 & 0.19 & 0.53 & 0.59 \\
 & hunyuan-a13b-think & 0.368 & 0.384 & 0.04 & 0.14 & 0.55 & 0.83 & 0.86 & 0.74 & 0.64 \\
 & hunyuan-a13b & 0.331 & 0.347 & 0.04 & 0.13 & 0.49 & 0.59 & 0.34 & 0.34 & 0.27 \\
Z.ai & glm-5 & 0.737 & 0.748 & 0.01 & 0.01 & 0.02 & 0.11 & 0.17 & 0.60 & 0.66 \\
 & glm-5-think & 0.735 & 0.746 & 0.01 & 0.01 & 0.02 & 0.09 & 0.22 & 0.55 & 0.69 \\
 & glm-5.1-think & 0.729 & 0.740 & 0.03 & 0.02 & 0.01 & 0.06 & 0.18 & 0.41 & 0.42 \\
 & glm-4.7-think & 0.723 & 0.735 & 0.01 & 0.01 & 0.02 & 0.08 & 0.27 & 0.58 & 0.66 \\
 & glm-4.6-think & 0.722 & 0.733 & 0.01 & 0.01 & 0.04 & 0.07 & 0.28 & 0.62 & 0.77 \\
 & glm-5.2 & 0.722 & 0.733 & 0.00 & 0.00 & 0.00 & 0.10 & 0.15 & 0.49 & 0.42 \\
 & glm-5.1 & 0.721 & 0.732 & 0.00 & 0.01 & 0.02 & 0.08 & 0.18 & 0.54 & 0.44 \\
 & glm-5-turbo & 0.711 & 0.723 & 0.00 & 0.01 & 0.02 & 0.07 & 0.21 & 0.51 & 0.45 \\
 & glm-5-turbo-think & 0.710 & 0.769 & 0.00 & 0.01 & 0.02 & 0.11 & 0.21 & 0.49 & 0.43 \\
 & glm-4.5-think & 0.668 & 0.726 & 0.00 & 0.01 & 0.01 & 0.12 & 0.14 & 0.24 & 0.23 \\
 & glm-4.5-air-think & 0.583 & 0.597 & 0.01 & 0.01 & 0.07 & 0.20 & 0.24 & 0.32 & 0.26 \\
 & glm-4.7-flash-think & 0.540 & 0.555 & 0.01 & 0.01 & 0.10 & 0.45 & 0.74 & 0.83 & 0.80 \\
 & glm-4-32b & 0.506 & 0.565 & 0.01 & 0.02 & 0.17 & 0.37 & 0.42 & 0.42 & 0.19 \\
Alibaba & qwen3.7-max & 0.710 & 0.723 & 0.00 & 0.00 & 0.02 & 0.07 & 0.12 & 0.20 & 0.20 \\
 & qwen3-max & 0.684 & 0.745 & 0.00 & 0.01 & 0.04 & 0.12 & 0.17 & 0.52 & 0.22 \\
 & qwen3.7-plus & 0.681 & 0.693 & 0.00 & 0.00 & 0.05 & 0.10 & 0.29 & 0.42 & 0.35 \\
 & qwen3.5-plus-think & 0.663 & 0.720 & 0.01 & 0.01 & 0.04 & 0.14 & 0.31 & 0.59 & 0.38 \\
 & qwen3.5-397b-a17b-think & 0.647 & 0.705 & 0.01 & 0.01 & 0.07 & 0.16 & 0.35 & 0.58 & 0.34 \\
 & qwen3.6-plus-think & 0.641 & 0.704 & 0.00 & 0.00 & 0.01 & 0.09 & 0.12 & 0.09 & 0.09 \\
 & qwen-plus & 0.629 & 0.694 & 0.01 & 0.01 & 0.05 & 0.13 & 0.29 & 0.47 & 0.24 \\
 & qwen3.5-122b-a10b-think & 0.613 & 0.675 & 0.01 & 0.01 & 0.05 & 0.20 & 0.50 & 0.71 & 0.50 \\
 & qwen3-next-80b-a3b & 0.564 & 0.579 & 0.01 & 0.01 & 0.05 & 0.24 & 0.36 & 0.52 & 0.40 \\
 & qwen3.5-flash-think & 0.548 & 0.606 & 0.00 & 0.01 & 0.11 & 0.41 & 0.65 & 0.76 & 0.59 \\
 & qwen3.5-35b-a3b-think & 0.548 & 0.603 & 0.00 & 0.03 & 0.10 & 0.44 & 0.64 & 0.76 & 0.54 \\
 & qwen3-235b-a22b-think & 0.547 & 0.612 & 0.01 & 0.01 & 0.10 & 0.21 & 0.39 & 0.46 & 0.17 \\
 & qwen-max & 0.528 & 0.589 & 0.01 & 0.01 & 0.10 & 0.20 & 0.14 & 0.08 & 0.06 \\
 & qwen3.5-27b-think & 0.518 & 0.575 & 0.01 & 0.03 & 0.16 & 0.49 & 0.79 & 0.81 & 0.79 \\
 & qwen-2.5-72b & 0.497 & 0.513 & 0.00 & 0.01 & 0.15 & 0.36 & 0.25 & 0.13 & 0.20 \\
 & qwq-32b-think & 0.475 & 0.490 & 0.01 & 0.03 & 0.21 & 0.51 & 0.59 & 0.55 & 0.49 \\
 & qwen3-32b-think & 0.466 & 0.481 & 0.01 & 0.03 & 0.11 & 0.62 & 0.55 & 0.53 & 0.38 \\
 & qwen3-30b-a3b-think & 0.448 & 0.465 & 0.01 & 0.01 & 0.19 & 0.48 & 0.42 & 0.30 & 0.26 \\
 & qwen-turbo & 0.442 & 0.497 & 0.04 & 0.03 & 0.30 & 0.46 & 0.36 & 0.26 & 0.17 \\
 & qwen3.5-9b-think & 0.441 & 0.494 & 0.00 & 0.03 & 0.29 & 0.63 & 0.75 & 0.67 & 0.67 \\
 & qwen3-14b-think & 0.436 & 0.452 & 0.03 & 0.03 & 0.23 & 0.57 & 0.44 & 0.38 & 0.38 \\
 & qwen3-8b-think & 0.422 & 0.471 & 0.01 & 0.06 & 0.41 & 0.62 & 0.58 & 0.49 & 0.47 \\
 & qwen-2.5-7b & 0.344 & 0.396 & 0.01 & 0.03 & 0.61 & 0.66 & 0.44 & 0.42 & 0.18 \\
 & qwen-2.5-0.5b & 0.170 & 0.180 & 0.06 & 0.64 & 0.79 & 0.77 & 0.69 & 0.73 & 0.63 \\
 & qwen3.5-0.8b & 0.141 & 0.213 & 0.10 & 0.24 & 0.33 & 0.26 & 0.42 & 0.57 & 0.81 \\
 & qwen3-0.6b & 0.110 & 0.116 & 0.29 & 0.34 & 0.38 & 0.24 & 0.18 & 0.14 & 0.28 \\
NVIDIA & nemotron-3-ultra & 0.701 & 0.713 & 0.00 & 0.00 & 0.03 & 0.08 & 0.13 & 0.41 & 0.35 \\
 & nemotron-3-super-120b & 0.609 & 0.622 & 0.00 & 0.01 & 0.05 & 0.24 & 0.46 & 0.50 & 0.49 \\
 & nemotron-70b & 0.549 & 0.564 & 0.00 & 0.00 & 0.04 & 0.10 & 0.03 & 0.06 & 0.12 \\
 & nemotron-super-49b-think & 0.461 & 0.477 & 0.01 & 0.01 & 0.28 & 0.47 & 0.33 & 0.32 & 0.28 \\
 & nemotron-3-nano-30b & 0.452 & 0.468 & 0.01 & 0.02 & 0.32 & 0.65 & 0.72 & 0.71 & 0.63 \\
 & nemotron-nano-9b-v2 & 0.340 & 0.357 & 0.01 & 0.07 & 0.11 & 0.15 & 0.03 & 0.03 & 0.08 \\
ByteDance & seed-2.0-lite-think & 0.697 & 0.757 & 0.00 & 0.01 & 0.04 & 0.10 & 0.29 & 0.53 & 0.43 \\
 & seed-2.0-mini-think & 0.629 & 0.686 & 0.01 & 0.03 & 0.10 & 0.26 & 0.44 & 0.75 & 0.71 \\
 & seed-1.6-think & 0.600 & 0.664 & 0.00 & 0.01 & 0.05 & 0.23 & 0.44 & 0.62 & 0.48 \\
 & seed-1.6-flash-think & 0.457 & 0.512 & 0.00 & 0.04 & 0.28 & 0.66 & 0.75 & 0.70 & 0.72 \\
StepFun & step-3.7-flash & 0.667 & 0.679 & 0.00 & 0.01 & 0.05 & 0.20 & 0.27 & 0.39 & 0.46 \\
 & step-3.5-flash-think & 0.660 & 0.676 & 0.00 & 0.01 & 0.04 & 0.18 & 0.34 & 0.59 & 0.60 \\
Amazon & nova-premier & 0.656 & 0.715 & 0.00 & 0.01 & 0.05 & 0.15 & 0.49 & 0.71 & 0.80 \\
 & nova-pro & 0.568 & 0.629 & 0.01 & 0.01 & 0.09 & 0.27 & 0.40 & 0.67 & 0.39 \\
 & nova-micro & 0.396 & 0.450 & 0.01 & 0.06 & 0.42 & 0.63 & 0.44 & 0.51 & 0.33 \\
Meta & llama-4-maverick & 0.643 & 0.656 & 0.01 & 0.01 & 0.08 & 0.14 & 0.14 & 0.20 & 0.22 \\
 & llama-3-70b & 0.570 & 0.634 & 0.01 & 0.01 & 0.08 & 0.15 & 0.09 & 0.12 & 0.10 \\
 & llama-3.3-70b & 0.568 & 0.583 & 0.02 & 0.01 & 0.11 & 0.14 & 0.09 & 0.13 & 0.19 \\
 & llama-3.1-70b & 0.522 & 0.538 & 0.01 & 0.01 & 0.04 & 0.09 & 0.04 & 0.05 & 0.10 \\
 & hermes-4-405b & 0.522 & 0.538 & 0.01 & 0.02 & 0.03 & 0.04 & 0.02 & 0.04 & 0.09 \\
 & llama-4-scout & 0.488 & 0.504 & 0.02 & 0.07 & 0.25 & 0.44 & 0.43 & 0.44 & 0.37 \\
 & hermes-3-405b & 0.456 & 0.472 & 0.01 & 0.02 & 0.03 & 0.04 & 0.00 & 0.02 & 0.05 \\
 & llama-3-8b & 0.438 & 0.455 & 0.01 & 0.04 & 0.29 & 0.35 & 0.40 & 0.11 & 0.28 \\
 & llama-3.1-8b & 0.392 & 0.409 & 0.01 & 0.03 & 0.12 & 0.12 & 0.03 & 0.02 & 0.06 \\
 & llama-3.2-3b & 0.318 & 0.366 & 0.02 & 0.07 & 0.23 & 0.22 & 0.03 & 0.01 & 0.01 \\
 & llama-3.2-1b & 0.236 & 0.273 & 0.08 & 0.20 & 0.42 & 0.24 & 0.06 & 0.03 & 0.04 \\
MiniMax & minimax-m3 & 0.639 & 0.652 & 0.00 & 0.01 & 0.03 & 0.10 & 0.17 & 0.30 & 0.25 \\
 & minimax-m2.7-think & 0.568 & 0.635 & 0.01 & 0.01 & 0.09 & 0.23 & 0.31 & 0.31 & 0.27 \\
 & minimax-m1-think & 0.434 & 0.479 & 0.34 & 0.38 & 0.30 & 0.39 & 0.37 & 0.47 & 0.33 \\
Mistral & mistral-medium-3.1 & 0.631 & 0.691 & 0.01 & 0.01 & 0.05 & 0.11 & 0.32 & 0.31 & 0.56 \\
 & mixtral-8x22b & 0.613 & 0.674 & 0.01 & 0.00 & 0.05 & 0.14 & 0.18 & 0.40 & 0.13 \\
 & mistral-large & 0.562 & 0.624 & 0.01 & 0.01 & 0.03 & 0.18 & 0.25 & 0.17 & 0.19 \\
 & mixtral-8x7b & 0.533 & 0.548 & 0.02 & 0.02 & 0.14 & 0.31 & 0.40 & 0.43 & 0.42 \\
 & mistral-small-24b & 0.489 & 0.505 & 0.01 & 0.01 & 0.11 & 0.26 & 0.23 & 0.18 & 0.24 \\
 & ministral-8b & 0.459 & 0.508 & 0.03 & 0.03 & 0.34 & 0.64 & 0.88 & 0.85 & 0.82 \\
 & mistral-nemo-12b & 0.428 & 0.444 & 0.03 & 0.04 & 0.34 & 0.48 & 0.61 & 0.50 & 0.51 \\
 & mistral-7b & 0.376 & 0.422 & 0.06 & 0.10 & 0.51 & 0.73 & 0.76 & 0.63 & 0.73 \\
 & ministral-3b & 0.365 & 0.412 & 0.04 & 0.12 & 0.58 & 0.85 & 0.80 & 0.84 & 0.85 \\
Cohere & command-a & 0.612 & 0.675 & 0.00 & 0.01 & 0.05 & 0.15 & 0.31 & 0.30 & 0.34 \\
 & command-r-plus & 0.466 & 0.524 & 0.01 & 0.02 & 0.09 & 0.19 & 0.08 & 0.10 & 0.06 \\
 & command-r7b & 0.364 & 0.410 & 0.04 & 0.15 & 0.49 & 0.65 & 0.48 & 0.62 & 0.29 \\
AI21 & jamba-large & 0.578 & 0.639 & 0.01 & 0.01 & 0.02 & 0.08 & 0.08 & 0.07 & 0.02 \\
Prime-intellect & intellect-3-think & 0.550 & 0.564 & 0.01 & 0.04 & 0.18 & 0.41 & 0.72 & 0.87 & 0.79 \\
Baidu & ernie-4.5-300b-a47b & 0.538 & 0.601 & 0.01 & 0.01 & 0.05 & 0.16 & 0.11 & 0.21 & 0.09 \\
Inclusionai & ling-2.6-flash & 0.530 & 0.751 & 0.01 & 0.01 & 0.12 & 0.26 & 0.27 & 0.20 & 0.00 \\
Microsoft & phi-4 & 0.432 & 0.448 & 0.02 & 0.04 & 0.34 & 0.66 & 0.78 & 0.73 & 0.72 \\
 & phi-3-mini & 0.289 & 0.304 & 0.07 & 0.26 & 0.64 & 0.74 & 0.57 & 0.57 & 0.57 \\
AllenAI & olmo-3.1-32b & 0.415 & 0.432 & 0.01 & 0.03 & 0.29 & 0.49 & 0.31 & 0.21 & 0.28 \\
Nex-agi & deepseek-v3.1-nex-n1 & 0.384 & 0.401 & 0.01 & 0.01 & 0.02 & 0.04 & 0.00 & 0.01 & 0.07 \\
HuggingFace & smollm2-1.7b & 0.301 & 0.317 & 0.04 & 0.20 & 0.83 & 0.94 & 0.94 & 0.95 & 0.97 \\
 & smollm2-360m & 0.284 & 0.294 & 0.17 & 0.46 & 0.87 & 0.88 & 0.87 & 0.91 & 0.82 \\
 & smollm2-135m & 0.125 & 0.131 & 0.51 & 0.77 & 0.91 & 0.97 & 0.96 & 0.98 & 0.95 \\
Ibm & granite-3.3-2b & 0.300 & 0.315 & 0.09 & 0.25 & 0.73 & 0.88 & 0.89 & 0.89 & 0.87 \\
\end{longtable}

%% file: tables/fp_controls.tex
\begin{table}[ht]\centering\small
\begin{tabular}{p{0.30\textwidth} p{0.22\textwidth} r r r r}
\toprule Student (derived) & Parent (base/teacher) & $J$ & lift & HSS & both\_w \\ \midrule
deepseek-r1-distill-llama-70b-think & llama-3.3-70b & 0.323 & 2.26 & 0.114 & 35 \\
deepseek-r1-distill-qwen-32b-think & qwq-32b-think & 0.325 & 7.88 & 0.037 & 164 \\
nemotron-70b & llama-3.1-70b & 0.810 & 6.84 & 0.083 & 12 \\
nemotron-super-49b-think & llama-3.3-70b & 0.153 & 2.13 & 0.034 & 29 \\
hermes-3-405b & llama-3.1-70b & 0.278 & 6.32 & 1.000 & 2 \\
hermes-4-405b & hermes-3-405b & 0.356 & 8.02 & 0.750 & 4 \\
\bottomrule \end{tabular}
\caption{Known-provenance positive controls. Each pair is a publicly-disclosed distillation or fine-tune. All show either lineage or cross-vendor-suspect HSS ($\geq 0.10$) when both\_w permits statistical resolution, confirming that the HSS metric detects shared training signal.}\label{tab:fp-controls}\end{table}

%% file: tables/fp_all_families.tex
{\small
\begin{longtable}{@{}p{0.20\textwidth} p{0.20\textwidth} r r r r l@{}}
\caption{Comprehensive within-family fingerprint metrics for all tracked model families (T5--T6 probes). $J$: Jaccard on correct-answer sets; lift: observed/expected-under-independence; $\mathrm{HSS}$: hallucination similarity; both\_w: joint-wrong probes; class: shared-base / lineage / retrained / independent. Families with all pairs having both\_w $<$ 8 are omitted.}\label{tab:fp-all-families}\\
\toprule From & To & $J$ & lift & HSS & both\_w & class \\ \midrule \endfirsthead
\toprule From & To & $J$ & lift & HSS & both\_w & class \\ \midrule \endhead
\multicolumn{7}{l}{\emph{OpenAI GPT base}} \\
\quad gpt-3.5-turbo & gpt-4 & 0.448 & 1.72 & 0.078 & 51 & retrained \\
\quad gpt-4 & gpt-4-turbo & 0.647 & 1.59 & 0.089 & 45 & retrained \\
\quad gpt-4-turbo & gpt-4o & 0.482 & 1.47 & 0.200 & 30 & independent \\
\quad gpt-4o & gpt-4.1 & 0.538 & 1.46 & 0.318 & 22 & lineage \\
\quad gpt-4.1 & gpt-5 & 0.758 & 1.38 & 0.000 & 38 & retrained \\
\quad gpt-5 & gpt-5.1 & 0.656 & 1.53 & 0.000 & 22 & retrained \\
\quad gpt-5.1 & gpt-5.2 & 0.555 & 1.69 & 0.091 & 11 & retrained \\
\quad gpt-5.2 & gpt-5.3 & 0.596 & 1.67 & 0.000 & 23 & retrained \\
\quad gpt-5.3 & gpt-5.4 & 0.715 & 1.63 & 0.083 & 48 & retrained \\
\multicolumn{7}{l}{\emph{OpenAI GPT-5 base/pro/think cluster}} \\
\quad gpt-5 & gpt-5-pro & 0.836 & 1.48 & 0.561 & 41 & shared-base \\
\quad gpt-5-pro & gpt-5-think & 0.846 & 1.46 & 0.523 & 44 & shared-base \\
\multicolumn{7}{l}{\emph{OpenAI GPT-5 .x transitions}} \\
\quad gpt-5 & gpt-5.1 & 0.656 & 1.53 & 0.000 & 22 & retrained \\
\quad gpt-5.1 & gpt-5.2 & 0.555 & 1.69 & 0.091 & 11 & retrained \\
\quad gpt-5.2 & gpt-5.2-pro & 0.776 & 2.15 & 0.286 & 42 & lineage \\
\quad gpt-5.2-pro & gpt-5.3 & 0.597 & 1.64 & 0.038 & 26 & retrained \\
\quad gpt-5.3 & gpt-5.4 & 0.715 & 1.63 & 0.083 & 48 & retrained \\
\quad gpt-5.4 & gpt-5.4-pro & 0.715 & 1.44 & 0.101 & 69 & lineage \\
\multicolumn{7}{l}{\emph{OpenAI GPT-mini}} \\
\quad gpt-4o-mini & gpt-4.1-mini & 0.304 & 2.06 & 0.070 & 143 & retrained \\
\quad gpt-4.1-mini & gpt-5-mini & 0.361 & 2.45 & 0.286 & 7 & independent \\
\quad gpt-5-mini & gpt-5.4-mini & 0.311 & 2.27 & 0.000 & 5 & independent \\
\multicolumn{7}{l}{\emph{OpenAI GPT-nano}} \\
\quad gpt-4.1-nano & gpt-5-nano & 0.107 & 4.38 & 0.000 & 8 & independent \\
\quad gpt-5-nano & gpt-5.4-nano & 0.100 & 6.67 & 0.000 & 4 & independent (small $n$) \\
\multicolumn{7}{l}{\emph{OpenAI o-series}} \\
\quad o1 & o3-mini & 0.131 & 1.62 & 0.000 & 17 & retrained \\
\quad o3-mini & o3 & 0.120 & 1.47 & 0.000 & 29 & retrained \\
\quad o3 & o4-mini-think & 0.404 & 1.38 & 0.000 & 46 & retrained \\
\multicolumn{7}{l}{\emph{Anthropic Claude Opus}} \\
\quad claude-opus-4 & claude-opus-4.1 & 0.525 & 6.56 & 0.889 & 9 & lineage \\
\quad claude-opus-4.1 & claude-opus-4.5 & 0.282 & 1.87 & 0.111 & 9 & independent \\
\quad claude-opus-4.5 & claude-opus-4.6 & 0.757 & 1.62 & 0.091 & 33 & retrained \\
\quad claude-opus-4.6 & claude-opus-4.7 & 0.604 & 1.64 & 0.000 & 15 & retrained \\
\multicolumn{7}{l}{\emph{Anthropic Claude Sonnet}} \\
\quad claude-3.7-sonnet & claude-sonnet-4 & 0.141 & 1.72 & 0.250 & 4 & independent (small $n$) \\
\quad claude-sonnet-4 & claude-sonnet-4.5 & 0.449 & 5.61 & 0.429 & 7 & independent \\
\quad claude-sonnet-4.5 & claude-sonnet-4.6 & 0.324 & 2.00 & 0.133 & 15 & independent \\
\multicolumn{7}{l}{\emph{Anthropic Claude Haiku}} \\
\quad claude-3-haiku & claude-3.5-haiku & 0.041 & 2.82 & 0.000 & 1 & independent (small $n$) \\
\quad claude-3.5-haiku & claude-haiku-4.5 & 0.136 & 2.95 & 0.167 & 18 & independent \\
\multicolumn{7}{l}{\emph{Google Gemini Flash}} \\
\quad gemini-2.0-flash & gemini-2.5-flash & 0.497 & 2.14 & 0.021 & 48 & retrained \\
\quad gemini-2.5-flash & gemini-3-flash & 0.297 & 1.09 & 0.067 & 15 & retrained \\
\multicolumn{7}{l}{\emph{Google Gemini Flash-Lite}} \\
\quad gemini-2.5-flash-lite & gemini-3.1-flash-lite & 0.237 & 1.29 & 0.029 & 70 & retrained \\
\multicolumn{7}{l}{\emph{Google Gemma}} \\
\quad gemma-2-2b & gemma-3-1b & 0.048 & 4.10 & 0.014 & 284 & retrained \\
\quad gemma-3-1b & gemma-3-4b & 0.000 & 0.00 & 0.006 & 349 & retrained \\
\quad gemma-3-4b & gemma-3-12b & 0.200 & 6.48 & 0.042 & 334 & retrained \\
\quad gemma-3-12b & gemma-2-27b & 0.000 & 0.00 & 0.044 & 90 & retrained \\
\quad gemma-2-27b & gemma-3-27b & 0.118 & 3.81 & 0.047 & 85 & retrained \\
\quad gemma-3-27b & gemma-4-31b & 0.076 & 1.33 & 0.000 & 238 & retrained \\
\multicolumn{7}{l}{\emph{Meta Llama 3 (70B)}} \\
\quad llama-3-70b & llama-3.1-70b & 0.424 & 4.06 & 0.625 & 8 & independent \\
\quad llama-3.1-70b & llama-3.3-70b & 0.500 & 4.42 & 0.400 & 15 & lineage \\
\multicolumn{7}{l}{\emph{Meta Llama 3 (8B)}} \\
\quad llama-3-8b & llama-3.1-8b & 0.214 & 12.70 & 0.125 & 8 & independent \\
\multicolumn{7}{l}{\emph{Meta Llama 4}} \\
\quad llama-4-scout & llama-4-maverick & 0.255 & 2.01 & 0.065 & 46 & retrained \\
\multicolumn{7}{l}{\emph{Alibaba Qwen dense (7B)}} \\
\quad qwen-2.5-7b & qwen3-8b-think & 0.143 & 10.26 & 0.040 & 100 & retrained \\
\multicolumn{7}{l}{\emph{Alibaba Qwen dense (70B)}} \\
\quad qwen-2.5-72b & qwq-32b-think & 0.218 & 4.28 & 0.032 & 62 & retrained \\
\multicolumn{7}{l}{\emph{Alibaba Qwen Max}} \\
\quad qwen-max & qwen3-max & 0.224 & 2.34 & 0.231 & 13 & independent \\
\multicolumn{7}{l}{\emph{Alibaba Qwen Plus}} \\
\quad qwen-plus & qwen3.5-plus-think & 0.464 & 2.10 & 0.000 & 94 & retrained \\
\quad qwen3.5-plus-think & qwen3.6-plus-think & 0.545 & 2.36 & 0.105 & 19 & lineage \\
\multicolumn{7}{l}{\emph{Alibaba Qwen large MoE}} \\
\quad qwen3-235b-a22b-think & qwen3.5-397b-a17b-think & 0.243 & 2.40 & 0.011 & 87 & retrained \\
\multicolumn{7}{l}{\emph{DeepSeek V3}} \\
\quad deepseek-v3 & deepseek-v3.1 & 0.380 & 1.82 & 0.227 & 22 & independent \\
\quad deepseek-v3.1 & deepseek-v3.2 & 0.503 & 2.25 & 0.333 & 21 & lineage \\
\multicolumn{7}{l}{\emph{Zhipu GLM}} \\
\quad glm-4.5-think & glm-4.6-think & 0.572 & 1.66 & 0.041 & 49 & retrained \\
\quad glm-4.6-think & glm-4.7-think & 0.707 & 1.58 & 0.209 & 129 & lineage \\
\quad glm-4.7-think & glm-5-think & 0.561 & 1.33 & 0.011 & 92 & retrained \\
\quad glm-5-think & glm-5.1-think & 0.687 & 1.46 & 0.012 & 84 & retrained \\
\multicolumn{7}{l}{\emph{xAI Grok}} \\
\quad grok-3 & grok-4 & 0.779 & 1.45 & 0.027 & 37 & retrained \\
\quad grok-4 & grok-4.20 & 0.557 & 1.38 & 0.000 & 44 & retrained \\
\multicolumn{7}{l}{\emph{Moonshot Kimi K2}} \\
\quad kimi-k2 & kimi-k2.5-think & 0.574 & 1.48 & 0.023 & 44 & retrained \\
\quad kimi-k2.5-think & kimi-k2.6-think & 0.765 & 1.63 & 0.510 & 49 & shared-base \\
\multicolumn{7}{l}{\emph{Mistral large/medium}} \\
\quad mistral-medium-3.1 & mistral-large & 0.420 & 3.19 & 0.000 & 33 & retrained \\
\multicolumn{7}{l}{\emph{Mistral small/open}} \\
\quad mistral-7b & mistral-nemo-12b & 0.189 & 5.80 & 0.018 & 164 & retrained \\
\quad mistral-nemo-12b & mistral-small-24b & 0.324 & 8.03 & 0.066 & 61 & retrained \\
\multicolumn{7}{l}{\emph{Mistral MoE (Mixtral)}} \\
\quad mixtral-8x7b & mixtral-8x22b & 0.209 & 1.78 & 0.000 & 51 & retrained \\
\multicolumn{7}{l}{\emph{Mistral ministral}} \\
\quad ministral-3b & ministral-8b & 0.085 & 2.46 & 0.023 & 264 & retrained \\
\multicolumn{7}{l}{\emph{Amazon Nova}} \\
\quad nova-micro & nova-pro & 0.135 & 3.51 & 0.070 & 114 & retrained \\
\quad nova-pro & nova-premier & 0.327 & 2.24 & 0.008 & 122 & retrained \\
\multicolumn{7}{l}{\emph{Cohere Command}} \\
\quad command-r7b & command-r-plus & 0.026 & 1.30 & 0.000 & 27 & retrained \\
\quad command-r-plus & command-a & 0.232 & 3.50 & 0.053 & 19 & retrained \\
\multicolumn{7}{l}{\emph{Microsoft Phi}} \\
\quad phi-3-mini & phi-4 & 0.000 & 0.00 & 0.000 & 181 & retrained \\
\bottomrule
\end{longtable}
}

%% file: tables/fp_cross_vendor.tex
{\small
\begin{longtable}{p{0.30\textwidth} p{0.30\textwidth} r r r r}
\caption{Complete list of cross-vendor outlier pairs with $\mathrm{HSS} \geq 0.20$ and $\geq 10$ joint-wrong probes on T5--T6.}\label{tab:fp-cross-full} \\
\toprule Model A & Model B & $J$ & lift & HSS & both\_w \\ \midrule \endfirsthead
\toprule Model A & Model B & $J$ & lift & HSS & both\_w \\ \midrule \endhead
ernie-4.5-300b-a47b & llama-3-70b & 0.433 & 3.79 & 0.500 & 12 \\
gpt-4.1-nano & llama-3.1-70b & 0.160 & 1.86 & 0.467 & 15 \\
ernie-4.5-300b-a47b & gpt-4o & 0.302 & 2.21 & 0.462 & 13 \\
gemini-2.0-flash & llama-3.1-70b & 0.278 & 2.35 & 0.462 & 13 \\
claude-opus-4.7 & gpt-5 & 0.534 & 1.46 & 0.417 & 12 \\
claude-opus-4.7 & kimi-k2.6-think & 0.546 & 1.62 & 0.400 & 10 \\
ernie-4.5-300b-a47b & mistral-small-24b & 0.254 & 4.51 & 0.400 & 15 \\
gpt-3.5-turbo & hermes-4-405b & 0.252 & 2.60 & 0.400 & 10 \\
grok-3 & llama-3.1-70b & 0.147 & 1.24 & 0.400 & 10 \\
llama-3-70b & mistral-large & 0.512 & 4.20 & 0.385 & 13 \\
gpt-5 & grok-4 & 0.587 & 1.23 & 0.381 & 21 \\
claude-opus-4.7 & gpt-5-think & 0.524 & 1.42 & 0.364 & 11 \\
gpt-4o-mini & hermes-4-405b & 0.262 & 3.27 & 0.364 & 11 \\
grok-3 & qwen-max & 0.145 & 1.50 & 0.364 & 11 \\
llama-3.1-70b & qwen3-max & 0.202 & 1.77 & 0.364 & 11 \\
llama-4-maverick & qwen-max & 0.209 & 2.25 & 0.364 & 11 \\
ernie-4.5-300b-a47b & qwen-max & 0.293 & 3.89 & 0.357 & 14 \\
gpt-5.5-pro & kimi-k2.6-think & 0.556 & 1.16 & 0.357 & 14 \\
claude-opus-4.8 & deepseek-v4-flash & 0.314 & 1.33 & 0.353 & 17 \\
gemini-2.5-flash & gpt-5-pro & 0.365 & 1.40 & 0.353 & 17 \\
claude-opus-4.7-think & llama-3-70b & 0.350 & 2.05 & 0.333 & 12 \\
gpt-5 & kimi-k2.6-think & 0.573 & 1.33 & 0.333 & 21 \\
gpt-5-pro & kimi-k2.6-think & 0.582 & 1.34 & 0.318 & 22 \\
deepseek-r1-distill-llama-70b-think & llama-3.1-70b & 0.220 & 2.16 & 0.312 & 16 \\
claude-opus-4.7-think & deepseek-v4-flash & 0.525 & 1.35 & 0.308 & 26 \\
claude-opus-4.8 & hy3-preview & 0.310 & 1.33 & 0.308 & 13 \\
gpt-4 & llama-3.1-70b & 0.241 & 1.97 & 0.308 & 13 \\
gpt-4o & kimi-k2 & 0.379 & 1.36 & 0.308 & 13 \\
claude-opus-4.7 & ernie-4.5-300b-a47b & 0.295 & 2.15 & 0.300 & 10 \\
claude-opus-4.8 & gemini-3.1-flash-lite & 0.319 & 1.30 & 0.300 & 10 \\
claude-sonnet-4.5 & gemini-3.1-flash-lite & 0.227 & 1.37 & 0.300 & 10 \\
claude-sonnet-4.5 & gpt-4.1 & 0.247 & 1.49 & 0.300 & 10 \\
claude-sonnet-4.5 & gpt-4o & 0.310 & 2.03 & 0.300 & 10 \\
claude-sonnet-4.5 & kimi-k2 & 0.280 & 1.82 & 0.300 & 10 \\
deepseek-v4-pro & gemini-3.1-flash-lite & 0.262 & 1.26 & 0.300 & 10 \\
gemini-3.1-flash-lite & llama-3.1-70b & 0.161 & 1.31 & 0.300 & 10 \\
hermes-4-405b & qwen3-next-80b-a3b & 0.293 & 3.81 & 0.300 & 10 \\
kimi-k2 & llama-3.1-70b & 0.208 & 1.78 & 0.300 & 10 \\
mistral-large & qwen-max & 0.420 & 5.04 & 0.300 & 10 \\
gpt-4o & llama-4-maverick & 0.459 & 1.68 & 0.294 & 17 \\
claude-sonnet-4.5-think & gpt-4o & 0.417 & 1.62 & 0.286 & 14 \\
deepseek-v4-flash & mistral-large & 0.211 & 1.47 & 0.286 & 14 \\
deepseek-v4-pro & gpt-4o-mini & 0.238 & 2.08 & 0.286 & 14 \\
gpt-4.1 & grok-3-mini-think & 0.350 & 1.45 & 0.286 & 14 \\
kimi-k2 & mistral-large & 0.285 & 2.02 & 0.286 & 21 \\
gemini-2.5-flash & gpt-5-think & 0.370 & 1.41 & 0.278 & 18 \\
claude-haiku-4.5 & gpt-4.1 & 0.069 & 1.34 & 0.273 & 11 \\
claude-opus-4.8 & kimi-k2 & 0.319 & 1.51 & 0.273 & 11 \\
command-r-plus & grok-4 & 0.105 & 1.51 & 0.273 & 11 \\
deepseek-v4-pro & gpt-4.1 & 0.294 & 1.40 & 0.273 & 11 \\
deepseek-v4-pro & o4-mini-think & 0.329 & 1.94 & 0.273 & 11 \\
gemini-3-flash & grok-3-mini-think & 0.277 & 1.12 & 0.273 & 11 \\
glm-4.5-air-think & gpt-5-mini-think & 0.339 & 2.69 & 0.273 & 11 \\
gpt-4.1-nano & hermes-4-405b & 0.150 & 1.96 & 0.273 & 11 \\
gpt-5-think & mistral-small-24b & 0.090 & 1.40 & 0.273 & 11 \\
grok-3 & mistral-large & 0.210 & 1.46 & 0.273 & 11 \\
llama-3.1-70b & nova-pro & 0.258 & 2.87 & 0.273 & 11 \\
claude-opus-4.8 & gemini-2.0-flash & 0.441 & 2.06 & 0.267 & 15 \\
gpt-4o & qwen3-max & 0.420 & 1.52 & 0.261 & 23 \\
kimi-k2.6-think & o3 & 0.573 & 1.31 & 0.261 & 23 \\
glm-5-turbo-think & grok-4 & 0.592 & 1.42 & 0.258 & 31 \\
claude-opus-4.7 & gemini-2.5-flash & 0.428 & 1.85 & 0.250 & 12 \\
claude-opus-4.7 & mixtral-8x22b & 0.373 & 1.81 & 0.250 & 12 \\
claude-opus-4.7 & o3 & 0.496 & 1.36 & 0.250 & 12 \\
claude-opus-4.7-think & qwen3.7-max & 0.552 & 1.53 & 0.250 & 16 \\
claude-opus-4.8 & glm-4.5-air-think & 0.348 & 2.18 & 0.250 & 12 \\
claude-opus-4.8 & grok-3-mini-think & 0.358 & 2.10 & 0.250 & 12 \\
claude-sonnet-4.5 & gpt-4 & 0.303 & 1.88 & 0.250 & 12 \\
deepseek-v4-flash & qwen-max & 0.145 & 1.51 & 0.250 & 16 \\
deepseek-v4-pro & qwen3-next-80b-a3b & 0.291 & 2.68 & 0.250 & 16 \\
ernie-4.5-300b-a47b & grok-3 & 0.197 & 1.40 & 0.250 & 20 \\
ernie-4.5-300b-a47b & llama-4-maverick & 0.244 & 1.89 & 0.250 & 20 \\
gemini-2.0-flash & qwen-max & 0.190 & 2.08 & 0.250 & 16 \\
gpt-5.5-think & kimi-k2.6-think & 0.546 & 1.16 & 0.250 & 16 \\
grok-3 & llama-3-70b & 0.258 & 1.49 & 0.250 & 16 \\
llama-3-70b & mistral-small-24b & 0.228 & 3.90 & 0.250 & 12 \\
glm-5-turbo-think & gpt-5 & 0.550 & 1.33 & 0.241 & 29 \\
gemini-2.0-flash & gpt-4o & 0.487 & 1.77 & 0.240 & 25 \\
gemini-2.5-pro & grok-4 & 0.550 & 1.12 & 0.240 & 25 \\
gpt-4o & grok-3 & 0.400 & 1.21 & 0.238 & 21 \\
gpt-5.3 & grok-4 & 0.522 & 1.25 & 0.238 & 21 \\
claude-opus-4.7 & deepseek-v4-flash & 0.473 & 1.34 & 0.235 & 17 \\
gemini-2.5-flash & gpt-5 & 0.371 & 1.42 & 0.235 & 17 \\
gpt-5.5 & grok-4 & 0.618 & 1.12 & 0.235 & 17 \\
kimi-k2 & llama-3-70b & 0.322 & 1.94 & 0.235 & 17 \\
grok-3 & qwen3-max & 0.530 & 1.44 & 0.232 & 56 \\
claude-haiku-4.5 & ernie-4.5-300b-a47b & 0.162 & 3.61 & 0.231 & 13 \\
claude-haiku-4.5-think & deepseek-v3.2 & 0.190 & 2.29 & 0.231 & 13 \\
claude-opus-4.7 & gemini-2.5-pro-think & 0.498 & 1.34 & 0.231 & 13 \\
deepseek-r1-distill-llama-70b-think & llama-3-70b & 0.326 & 2.37 & 0.231 & 26 \\
gemma-2-27b & gpt-5 & 0.053 & 1.41 & 0.231 & 13 \\
gpt-3.5-turbo & llama-3.1-70b & 0.325 & 2.80 & 0.231 & 13 \\
ling-2.6-flash & seed-1.6-think & 0.151 & 3.22 & 0.231 & 13 \\
llama-3-70b & qwen-max & 0.358 & 4.19 & 0.231 & 13 \\
gemini-3.1-flash-lite & gpt-4o & 0.438 & 1.24 & 0.227 & 22 \\
gpt-5-think & grok-4 & 0.635 & 1.27 & 0.227 & 22 \\
claude-haiku-4.5 & deepseek-v3.2 & 0.122 & 2.35 & 0.222 & 18 \\
claude-opus-4.7 & gemini-2.5-flash-think & 0.515 & 1.57 & 0.222 & 18 \\
claude-opus-4.7 & glm-5 & 0.484 & 1.41 & 0.222 & 18 \\
claude-opus-4.8 & gpt-4o-mini & 0.350 & 2.63 & 0.222 & 18 \\
command-a & deepseek-v4-flash & 0.289 & 1.32 & 0.222 & 36 \\
gemini-2.5-pro-think & kimi-k2.6-think & 0.551 & 1.25 & 0.222 & 27 \\
glm-5.1 & gpt-5 & 0.566 & 1.29 & 0.222 & 36 \\
gpt-4 & llama-3.3-70b & 0.342 & 1.89 & 0.222 & 18 \\
deepseek-v4-flash & gemini-3-flash-think & 0.648 & 1.08 & 0.217 & 23 \\
claude-opus-4.8 & nova-pro & 0.280 & 2.17 & 0.214 & 14 \\
claude-sonnet-4.5 & glm-5-think & 0.265 & 1.63 & 0.214 & 14 \\
claude-sonnet-4.5-think & qwen-max & 0.254 & 2.66 & 0.214 & 14 \\
claude-sonnet-4.6 & deepseek-v3.1 & 0.361 & 1.68 & 0.214 & 14 \\
deepseek-v4-pro & gpt-3.5-turbo & 0.316 & 1.93 & 0.214 & 14 \\
gpt-4-turbo & qwen-max & 0.186 & 1.90 & 0.214 & 14 \\
gpt-4.1 & llama-3.3-70b & 0.267 & 1.44 & 0.214 & 14 \\
gpt-4.1-nano & llama-3-70b & 0.252 & 2.24 & 0.214 & 28 \\
gpt-4o & llama-3.3-70b & 0.371 & 2.13 & 0.214 & 14 \\
grok-3 & o1 & 0.543 & 1.22 & 0.214 & 14 \\
llama-4-maverick & mistral-large & 0.273 & 2.04 & 0.214 & 14 \\
mistral-small-24b & qwen-max & 0.204 & 4.34 & 0.214 & 14 \\
claude-sonnet-4.5-think & gemini-3.1-flash-lite & 0.439 & 1.32 & 0.211 & 19 \\
claude-sonnet-4.5-think & gpt-4 & 0.516 & 1.71 & 0.211 & 19 \\
gemini-2.0-flash & llama-3-70b & 0.338 & 2.13 & 0.208 & 24 \\
gpt-4o-mini & qwen-max & 0.241 & 3.21 & 0.208 & 24 \\
gpt-5-pro & grok-4 & 0.585 & 1.22 & 0.208 & 24 \\
gpt-5-think & kimi-k2.6-think & 0.601 & 1.36 & 0.208 & 24 \\
ernie-4.5-300b-a47b & glm-5.2 & 0.260 & 1.83 & 0.207 & 29 \\
gemini-2.5-flash & glm-5.1 & 0.356 & 1.47 & 0.207 & 29 \\
glm-5.1-think & gpt-5 & 0.576 & 1.25 & 0.207 & 29 \\
gpt-5 & kimi-k2.5-think & 0.564 & 1.21 & 0.205 & 44 \\
claude-3.5-haiku & ernie-4.5-300b-a47b & 0.217 & 2.23 & 0.200 & 10 \\
claude-3.5-haiku & hy3-preview & 0.253 & 1.48 & 0.200 & 10 \\
claude-3.5-haiku & kimi-k2 & 0.330 & 1.98 & 0.200 & 10 \\
claude-3.5-haiku & mistral-large & 0.226 & 2.29 & 0.200 & 10 \\
claude-haiku-4.5 & llama-3-70b & 0.165 & 3.49 & 0.200 & 10 \\
claude-haiku-4.5 & llama-4-maverick & 0.121 & 2.35 & 0.200 & 15 \\
claude-opus-4.7 & gpt-4o & 0.542 & 1.82 & 0.200 & 10 \\
claude-opus-4.7 & gpt-5.5-think & 0.453 & 1.18 & 0.200 & 10 \\
claude-opus-4.7 & qwen3-max & 0.529 & 1.76 & 0.200 & 15 \\
claude-opus-4.7 & qwen3.7-max & 0.524 & 1.57 & 0.200 & 10 \\
claude-sonnet-4-think & glm-5-think & 0.316 & 1.57 & 0.200 & 10 \\
claude-sonnet-4.5 & llama-4-maverick & 0.340 & 2.21 & 0.200 & 10 \\
claude-sonnet-4.5-think & llama-3.1-70b & 0.243 & 2.13 & 0.200 & 10 \\
command-a & gpt-4 & 0.351 & 1.66 & 0.200 & 20 \\
command-r-plus & glm-4.7-think & 0.112 & 1.63 & 0.200 & 15 \\
command-r-plus & seed-1.6-think & 0.198 & 3.12 & 0.200 & 25 \\
deepseek-r1-distill-llama-70b-think & hermes-4-405b & 0.214 & 2.38 & 0.200 & 10 \\
deepseek-v4-flash & llama-3.1-70b & 0.171 & 1.40 & 0.200 & 10 \\
deepseek-v4-pro & gemini-2.0-flash & 0.322 & 1.80 & 0.200 & 15 \\
deepseek-v4-pro & nemotron-3-ultra & 0.305 & 1.56 & 0.200 & 10 \\
gemini-2.5-flash & gpt-5.5 & 0.323 & 1.19 & 0.200 & 15 \\
gemini-2.5-flash & o3 & 0.366 & 1.39 & 0.200 & 25 \\
gemini-2.5-pro & gpt-5 & 0.632 & 1.20 & 0.200 & 25 \\
gemini-2.5-pro-think & gpt-5 & 0.635 & 1.21 & 0.200 & 25 \\
gemini-2.5-pro-think & grok-4 & 0.552 & 1.13 & 0.200 & 25 \\
gemma-3-270m & llama-3.3-70b & 0.041 & 0.51 & 0.200 & 10 \\
glm-4.5-air-think & llama-3.1-70b & 0.314 & 3.00 & 0.200 & 10 \\
glm-5.1-think & gpt-5-think & 0.592 & 1.25 & 0.200 & 30 \\
gpt-4 & qwen-max & 0.196 & 2.03 & 0.200 & 10 \\
gpt-4-turbo & llama-3.3-70b & 0.294 & 1.63 & 0.200 & 20 \\
gpt-5-mini-think & llama-4-maverick & 0.301 & 2.03 & 0.200 & 10 \\
grok-3-mini-think & mimo-v2-pro & 0.366 & 1.94 & 0.200 & 10 \\
hermes-4-405b & nemotron-3-super-120b & 0.185 & 2.09 & 0.200 & 10 \\
hy3-preview & mistral-large & 0.229 & 1.58 & 0.200 & 20 \\
ling-2.6-flash & qwen3.5-122b-a10b-think & 0.138 & 2.99 & 0.200 & 15 \\
llama-3.1-70b & nemotron-3-ultra & 0.211 & 1.76 & 0.200 & 10 \\
llama-3.1-70b & qwen3-235b-a22b-think & 0.119 & 1.82 & 0.200 & 10 \\
\bottomrule \end{longtable}
}

%% file: references.bib
@article{allenzhu2025,
  title={Physics of Language Models: Part 3.3, Knowledge Capacity Scaling Laws},
  author={Allen-Zhu, Zeyuan and Li, Yuanzhi},
  journal={International Conference on Learning Representations},
  year={2025}
}

@article{kandpal2023,
  title={Large Language Models Struggle to Learn Long-Tail Knowledge},
  author={Kandpal, Nikhil and Deng, Haikang and Roberts, Adam and Wallace, Eric and Raffel, Colin},
  journal={International Conference on Machine Learning},
  year={2023}
}

@article{densing2025,
  title={The Densing Law of LLMs},
  author={Huang, Chaojun and others},
  journal={Nature Machine Intelligence},
  year={2025}
}

@article{epochai2024,
  title={Estimating Training Compute of Frontier AI Models},
  author={{Epoch AI}},
  year={2024},
  note={\url{https://epoch.ai/}}
}

@article{cai2025substitution,
  title={Model Substitution Auditing for Language Model APIs},
  author={Cai, Dingfan and others},
  year={2025}
}

@article{llmmap2025,
  title={LLMmap: Fingerprinting For Large Language Models},
  author={Pasquini, Dario and Tsingenopoulos, Ilias and Rossi, Mauro},
  journal={USENIX Security Symposium},
  year={2025}
}

@article{overshadow2025,
  title={The Law of Knowledge Overshadowing in Large Language Models},
  author={Zhang, Yongqi and others},
  journal={Association for Computational Linguistics},
  year={2025}
}

@article{hong2025,
  title={The Rise of Parameter Specialization for Knowledge Storage in Large Language Models},
  author={Hong, Jiaxi and others},
  journal={Neural Information Processing Systems},
  year={2025}
}

@article{chang2024,
  title={How Do Large Language Models Acquire Factual Knowledge During Pretraining?},
  author={Chang, Hoyeon and Park, Jinho and Ye, Seonghyeon and others},
  journal={Neural Information Processing Systems},
  year={2024}
}

@article{tirumala2022,
  title={Memorization Without Overfitting: Analyzing the Training Dynamics of Large Language Models},
  author={Tirumala, Kushal and Markosyan, Aram and Zettlemoyer, Luke and Aghajanyan, Armen},
  journal={Neural Information Processing Systems},
  year={2022}
}

@article{kaplan2020scaling,
  title={Scaling Laws for Neural Language Models},
  author={Kaplan, Jared and McCandlish, Sam and Henighan, Tom and Brown, Tom B. and Chess, Benjamin and Child, Rewon and Gray, Scott and Radford, Alec and Wu, Jeffrey and Amodei, Dario},
  journal={arXiv preprint arXiv:2001.08361},
  year={2020}
}

@inproceedings{hoffmann2022training,
  title={Training Compute-Optimal Large Language Models},
  author={Hoffmann, Jordan and Borgeaud, Sebastian and Mensch, Arthur and Buchatskaya, Elena and Cai, Trevor and Rutherford, Eliza and de Las Casas, Diego and Hendricks, Lisa Anne and Welbl, Johannes and Clark, Aidan and others},
  booktitle={Advances in Neural Information Processing Systems},
  volume={35},
  year={2022}
}

@inproceedings{clark2022unified,
  title={Unified Scaling Laws for Routed Language Models},
  author={Clark, Aidan and de Las Casas, Diego and Guy, Aurelia and Mensch, Arthur and Paganini, Michela and Hoffmann, Jordan and Damoc, Bogdan and Hechtman, Blake A. and Cai, Trevor and Borgeaud, Sebastian and others},
  booktitle={International Conference on Machine Learning},
  year={2022}
}

@inproceedings{geva2021transformer,
  title={Transformer Feed-Forward Layers Are Key-Value Memories},
  author={Geva, Mor and Schuster, Roei and Berant, Jonathan and Levy, Omer},
  booktitle={Proceedings of the 2021 Conference on Empirical Methods in Natural Language Processing},
  pages={5484--5495},
  year={2021}
}

@inproceedings{meng2022locating,
  title={Locating and Editing Factual Associations in {GPT}},
  author={Meng, Kevin and Bau, David and Andonian, Alex and Belinkov, Yonatan},
  booktitle={Advances in Neural Information Processing Systems},
  volume={35},
  year={2022}
}

@inproceedings{dai2022knowledge,
  title={Knowledge Neurons in Pretrained Transformers},
  author={Dai, Damai and Dong, Li and Hao, Yaru and Sui, Zhifang and Chang, Baobao and Wei, Furu},
  booktitle={Proceedings of the 60th Annual Meeting of the Association for Computational Linguistics},
  pages={8493--8502},
  year={2022}
}

@inproceedings{petroni2019language,
  title={Language Models as Knowledge Bases?},
  author={Petroni, Fabio and Rockt{\"a}schel, Tim and Riedel, Sebastian and Lewis, Patrick and Bakhtin, Anton and Wu, Yuxiang and Miller, Alexander H.},
  booktitle={Proceedings of the 2019 Conference on Empirical Methods in Natural Language Processing},
  pages={2463--2473},
  year={2019}
}

@inproceedings{roberts2020much,
  title={How Much Knowledge Can You Pack Into the Parameters of a Language Model?},
  author={Roberts, Adam and Raffel, Colin and Shazeer, Noam},
  booktitle={Proceedings of the 2020 Conference on Empirical Methods in Natural Language Processing},
  pages={5418--5426},
  year={2020}
}

@inproceedings{hendrycks2021measuring,
  title={Measuring Massive Multitask Language Understanding},
  author={Hendrycks, Dan and Burns, Collin and Basart, Steven and Zou, Andy and Mazeika, Mantas and Song, Dawn and Steinhardt, Jacob},
  booktitle={International Conference on Learning Representations},
  year={2021}
}

@inproceedings{wang2024mmlupro,
  title={{MMLU-Pro}: A More Robust and Challenging Multi-Task Language Understanding Benchmark},
  author={Wang, Yubo and Ma, Xueguang and Zhang, Ge and Ni, Yuansheng and Chandra, Abhranil and Guo, Shiguang and Ren, Weiming and Arulraj, Aaran and He, Xuan and Jiang, Ziyan and Li, Tianle and Ku, Max and Wang, Kai and Zhuang, Alex and Fan, Rongqi and Yue, Xiang and Chen, Wenhu},
  booktitle={Advances in Neural Information Processing Systems},
  year={2024}
}

@inproceedings{rein2023gpqa,
  title={{GPQA}: A Graduate-Level {Google}-Proof {Q\&A} Benchmark},
  author={Rein, David and Hou, Betty Li and Stickland, Asa Cooper and Petty, Jackson and Pang, Richard Yuanzhe and Dirani, Julien and Michael, Julian and Bowman, Samuel R.},
  booktitle={Conference on Language Modeling},
  year={2024}
}

@misc{wei2024simpleqa,
  title={Measuring Short-Form Factuality in Large Language Models},
  author={Wei, Jason and Karina, Nguyen and Chung, Hyung Won and Jiao, Yunxin Joy and Papay, Spencer and Glaese, Amelia and Schulman, John and Fedus, William},
  howpublished={OpenAI},
  year={2024},
  note={\url{https://openai.com/index/introducing-simpleqa/}}
}

@misc{cobbe2021gsm8k,
  title={Training Verifiers to Solve Math Word Problems},
  author={Cobbe, Karl and Kosaraju, Vineet and Bavarian, Mohammad and Chen, Mark and Jun, Heewoo and Kaiser, Lukasz and Plappert, Matthias and Tworek, Jerry and Hilton, Jacob and Nakano, Reiichiro and Hesse, Christopher and Schulman, John},
  year={2021},
  eprint={2110.14168},
  archivePrefix={arXiv},
  primaryClass={cs.LG}
}

@misc{phan2025hle,
  title={Humanity's Last Exam},
  author={Phan, Long and Gatti, Alice and Han, Ziwen and Li, Nathaniel and Hu, Josephina and Zhang, Hugh and others},
  year={2025},
  eprint={2501.14249},
  archivePrefix={arXiv},
  primaryClass={cs.CL}
}

@inproceedings{joshi2017triviaqa,
  title={{TriviaQA}: A Large Scale Distantly Supervised Challenge Dataset for Reading Comprehension},
  author={Joshi, Mandar and Choi, Eunsol and Weld, Daniel and Zettlemoyer, Luke},
  booktitle={Proceedings of the 55th Annual Meeting of the Association for Computational Linguistics},
  pages={1601--1611},
  year={2017}
}

@article{kwiatkowski2019natural,
  title={Natural Questions: A Benchmark for Question Answering Research},
  author={Kwiatkowski, Tom and Palomaki, Jennimaria and Redfield, Olivia and Collins, Michael and Parikh, Ankur and Alberti, Chris and Epstein, Danielle and Polosukhin, Illia and Devlin, Jacob and Lee, Kenton and others},
  journal={Transactions of the Association for Computational Linguistics},
  volume={7},
  pages={453--466},
  year={2019}
}

@inproceedings{shazeer2017outrageously,
  title={Outrageously Large Neural Networks: The Sparsely-Gated Mixture-of-Experts Layer},
  author={Shazeer, Noam and Mirhoseini, Azalia and Maziarz, Krzysztof and Davis, Andy and Le, Quoc V. and Hinton, Geoffrey E. and Dean, Jeff},
  booktitle={International Conference on Learning Representations},
  year={2017}
}

@article{fedus2022switch,
  title={Switch Transformers: Scaling to Trillion Parameter Models with Simple and Efficient Sparsity},
  author={Fedus, William and Zoph, Barret and Shazeer, Noam},
  journal={Journal of Machine Learning Research},
  volume={23},
  number={120},
  pages={1--39},
  year={2022}
}

@inproceedings{lewis2020retrieval,
  title={Retrieval-Augmented Generation for Knowledge-Intensive {NLP} Tasks},
  author={Lewis, Patrick and Perez, Ethan and Piktus, Aleksandra and Petroni, Fabio and Karpukhin, Vladimir and Goyal, Naman and K{\"u}ttler, Heinrich and Lewis, Mike and Yih, Wen-tau and Rockt{\"a}schel, Tim and Riedel, Sebastian and Kiela, Douwe},
  booktitle={Advances in Neural Information Processing Systems},
  volume={33},
  pages={9459--9474},
  year={2020}
}

@inproceedings{ouyang2022training,
  title={Training Language Models to Follow Instructions with Human Feedback},
  author={Ouyang, Long and Wu, Jeffrey and Jiang, Xu and Almeida, Diogo and Wainwright, Carroll and Mishkin, Pamela and Zhang, Chong and Agarwal, Sandhini and Slama, Katarina and Ray, Alex and others},
  booktitle={Advances in Neural Information Processing Systems},
  volume={35},
  year={2022}
}

@inproceedings{christiano2017deep,
  title={Deep Reinforcement Learning from Human Preferences},
  author={Christiano, Paul F. and Leike, Jan and Brown, Tom and Martic, Miljan and Legg, Shane and Amodei, Dario},
  booktitle={Advances in Neural Information Processing Systems},
  volume={30},
  year={2017}
}

@article{bai2022constitutional,
  title={Constitutional {AI}: Harmlessness from {AI} Feedback},
  author={Bai, Yuntao and Kadavath, Saurav and Kundu, Sandipan and Askell, Amanda and Kernion, Jackson and Jones, Andy and Chen, Anna and Goldie, Anna and Mirhoseini, Azalia and McKinnon, Cameron and others},
  journal={arXiv preprint arXiv:2212.08073},
  year={2022}
}

@book{zipf1949human,
  title={Human Behavior and the Principle of Least Effort},
  author={Zipf, George Kingsley},
  publisher={Addison-Wesley Press},
  year={1949}
}

@article{piantadosi2014zipf,
  title={Zipf's Word Frequency Law in Natural Language: A Critical Review and Future Directions},
  author={Piantadosi, Steven T.},
  journal={Psychonomic Bulletin \& Review},
  volume={21},
  number={5},
  pages={1112--1130},
  year={2014}
}

@article{shannon1948mathematical,
  title={A Mathematical Theory of Communication},
  author={Shannon, Claude E.},
  journal={The Bell System Technical Journal},
  volume={27},
  number={3},
  pages={379--423},
  year={1948}
}

@inproceedings{wei2022chain,
  title={Chain-of-Thought Prompting Elicits Reasoning in Large Language Models},
  author={Wei, Jason and Wang, Xuezhi and Schuurmans, Dale and Bosma, Maarten and Ichter, Brian and Xia, Fei and Chi, Ed and Le, Quoc V. and Zhou, Denny},
  booktitle={Advances in Neural Information Processing Systems},
  volume={35},
  pages={24824--24837},
  year={2022}
}

@article{liang2023holistic,
  title={Holistic Evaluation of Language Models},
  author={Liang, Percy and Bommasani, Rishi and Lee, Tony and Tsipras, Dimitris and Soylu, Dilara and Yasunaga, Michihiro and Zhang, Yian and Narayanan, Deepak and Wu, Yuhuai and Kumar, Ananya and others},
  journal={Transactions on Machine Learning Research},
  year={2023}
}

@inproceedings{mallen2023trust,
  title={When Not to Trust Language Models: Investigating Effectiveness of Parametric and Non-Parametric Memories},
  author={Mallen, Alex and Asai, Akari and Zhong, Victor and Das, Rajarshi and Khashabi, Daniel and Hajishirzi, Hannaneh},
  booktitle={Proceedings of the 61st Annual Meeting of the Association for Computational Linguistics},
  pages={9802--9822},
  year={2023}
}

@inproceedings{carlini2021extracting,
  title={Extracting Training Data from Large Language Models},
  author={Carlini, Nicholas and Tram{\`e}r, Florian and Wallace, Eric and Jagielski, Matthew and Herbert-Voss, Ariel and Lee, Katherine and Roberts, Adam and Brown, Tom and Song, Dawn and Erlingsson, {\'U}lfar and Oprea, Alina and Raffel, Colin},
  booktitle={30th USENIX Security Symposium},
  pages={2633--2650},
  year={2021}
}

@inproceedings{carlini2023quantifying,
  title={Quantifying Memorization Across Neural Language Models},
  author={Carlini, Nicholas and Ippolito, Daphne and Jagielski, Matthew and Lee, Katherine and Tram{\`e}r, Florian and Zhang, Chiyuan},
  booktitle={International Conference on Learning Representations},
  year={2023}
}

@article{touvron2023llama,
  title={{LLaMA}: Open and Efficient Foundation Language Models},
  author={Touvron, Hugo and Lavril, Thibaut and Izacard, Gautier and Martinet, Xavier and Lachaux, Marie-Anne and Lacroix, Timoth{\'e}e and Rozi{\`e}re, Baptiste and Goyal, Naman and Hambro, Eric and Azhar, Faisal and others},
  journal={arXiv preprint arXiv:2302.13971},
  year={2023}
}

@article{touvron2023llama2,
  title={Llama 2: Open Foundation and Fine-Tuned Chat Models},
  author={Touvron, Hugo and Martin, Louis and Stone, Kevin and Albert, Peter and Almahairi, Amjad and Babaei, Yasmine and Bashlykov, Nikolay and Batra, Soumya and Bhargava, Prajjwal and Bhosale, Shruti and others},
  journal={arXiv preprint arXiv:2307.09288},
  year={2023}
}

@article{yang2024qwen2,
  title={Qwen2 Technical Report},
  author={Yang, An and Yang, Baosong and Hui, Binyuan and Zheng, Bo and Yu, Bowen and Zhou, Chang and Li, Chengpeng and Li, Chengyuan and Liu, Dayiheng and Huang, Fei and others},
  journal={arXiv preprint arXiv:2407.10671},
  year={2024}
}

@article{gemma2024,
  title={Gemma: Open Models Based on Gemini Research and Technology},
  author={{Gemma Team} and Mesnard, Thomas and Hardin, Cassidy and Dadashi, Robert and Bhupatiraju, Surya and Pathak, Shreya and Sifre, Laurent and Rivi{\`e}re, Morgane and Kale, Mihir Sanjay and Love, Juliette and others},
  journal={arXiv preprint arXiv:2403.08295},
  year={2024}
}

@article{deepseek2024v2,
  title={{DeepSeek-V2}: A Strong, Economical, and Efficient Mixture-of-Experts Language Model},
  author={{DeepSeek-AI}},
  journal={arXiv preprint arXiv:2405.04434},
  year={2024}
}

@article{deepseek2025r1,
  title={{DeepSeek-R1}: Incentivizing Reasoning Capability in {LLMs} via Reinforcement Learning},
  author={{DeepSeek-AI}},
  journal={arXiv preprint arXiv:2501.12948},
  year={2025}
}

@article{openai2023gpt4,
  title={{GPT-4} Technical Report},
  author={{OpenAI}},
  journal={arXiv preprint arXiv:2303.08774},
  year={2023}
}

@article{vrandecic2014wikidata,
  title={Wikidata: A Free Collaborative Knowledgebase},
  author={Vrande{\v{c}}i{\'c}, Denny and Kr{\"o}tzsch, Markus},
  journal={Communications of the ACM},
  volume={57},
  number={10},
  pages={78--85},
  year={2014}
}

@inproceedings{brown2020language,
  title={Language Models are Few-Shot Learners},
  author={Brown, Tom and Mann, Benjamin and Ryder, Nick and Subbiah, Melanie and Kaplan, Jared D. and Dhariwal, Prafulla and Neelakantan, Arvind and Shyam, Pranav and Sastry, Girish and Askell, Amanda and others},
  booktitle={Advances in Neural Information Processing Systems},
  volume={33},
  pages={1877--1901},
  year={2020}
}

@inproceedings{groeneveld2024olmo,
  title={{OLMo}: Accelerating the Science of Language Models},
  author={Groeneveld, Dirk and Beltagy, Iz and Walsh, Pete and Bhagia, Akshita and Kinney, Rodney and Tafjord, Oyvind and Jha, Ananya Harsh and Ivison, Hamish and Magnusson, Ian and Wang, Yizhong and others},
  booktitle={Proceedings of the 62nd Annual Meeting of the Association for Computational Linguistics},
  year={2024}
}

@article{morris2025memorize,
  title={How Much Do Language Models Memorize?},
  author={Morris, John X. and Sitawarin, Chawin and Guo, Chuan and Kokhlikyan, Narine and Suh, G. Edward and Rush, Alexander M. and Chaudhuri, Kamalika and Mahloujifar, Saeed},
  journal={arXiv preprint arXiv:2505.24832},
  year={2025}
}

@article{pan2025compression,
  title={Understanding {LLM} Behaviors via Compression: Data Generation, Knowledge Acquisition and Scaling Laws},
  author={Pan, Zhixuan and Wang, Shaowen and Li, Jian},
  journal={Neural Information Processing Systems},
  year={2025}
}

@inproceedings{lu2024factmemorization,
  title={Scaling Laws for Fact Memorization of Large Language Models},
  author={Lu, Xingyu and Li, Xiaonan and Cheng, Qinyuan and Ding, Kai and Huang, Xuanjing and Qiu, Xipeng},
  booktitle={Findings of the Association for Computational Linguistics: EMNLP 2024},
  year={2024}
}

@article{gao2025modelequality,
  title={Model Equality Testing: Which Model Is This {API} Serving?},
  author={Gao, Irena and Liang, Percy and Guestrin, Carlos},
  journal={International Conference on Learning Representations},
  year={2025}
}

@article{ludziejewski2025moe,
  title={Joint {MoE} Scaling Laws: Mixture of Experts Can Be Memory Efficient},
  author={Ludziejewski, Jan and Pi{\'o}ro, Maciej and Krajewski, Jakub and Stefaniak, Maciej and Krutu{\l}, Micha{\l} and Ma{\l}a{\'s}nicki, Jan and Cygan, Marek and Sankowski, Piotr and Adamczewski, Kamil and Mi{\l}o{\'s}, Piotr and Jaszczur, Sebastian},
  journal={International Conference on Machine Learning},
  year={2025}
}

@article{zhao2025moescaling,
  title={Towards a Comprehensive Scaling Law of Mixture-of-Experts},
  author={Zhao, Guoliang and Fu, Yuhan and Li, Shuaipeng and Sun, Xingwu and Xie, Ruobing and Wang, An and Han, Weidong and Yang, Zhen and Sun, Weixuan and Zhang, Yudong and Xu, Cheng-zhong and Wang, Di and Jiang, Jie},
  journal={arXiv preprint arXiv:2509.23678},
  year={2025}
}

@article{tsai2025rofl,
  title={{RoFL}: Robust Fingerprinting of Language Models},
  author={Tsai, Yun-Yun and Guo, Chuan and Yang, Junfeng and van der Maaten, Laurens},
  journal={arXiv preprint arXiv:2505.12682},
  year={2025}
}

@article{nasery2025fingerprinting,
  title={Scalable Fingerprinting of Large Language Models},
  author={Nasery, Anshul and Hayase, Jonathan and Brooks, Creston and Sheng, Peiyao and Tyagi, Himanshu and Viswanath, Pramod and Oh, Sewoong},
  journal={arXiv preprint arXiv:2502.07760},
  year={2025}
}

@article{shi2025kddistillation,
  title={Knowledge Distillation Detection for Open-weights Models},
  author={Shi, Qin and Zheng, Amber Yijia and Song, Qifan and Yeh, Raymond A.},
  journal={Neural Information Processing Systems},
  year={2025}
}

@article{li2025experts,
  title={Leave It to the Experts: Detecting Knowledge Distillation via {MoE} Expert Signatures},
  author={Li, Pingzhi and Huang, Morris Yu-Chao and Tan, Zhen and Song, Qingquan and Peng, Jie and Zou, Kai and Cheng, Yu and Xu, Kaidi and Chen, Tianlong},
  journal={arXiv preprint arXiv:2510.16968},
  year={2025}
}

@article{badhe2026longtail,
  title={Long-Tail Knowledge in Large Language Models: Taxonomy, Mechanisms, Interventions and Implications},
  author={Badhe, Sanket and Shah, Deep and Kathrotia, Nehal},
  journal={arXiv preprint arXiv:2602.16201},
  year={2026}
}

@article{chen2025factoids,
  title={Continual Memorization of Factoids in Language Models},
  author={Chen, Howard and Geng, Jiayi and Bhaskar, Adithya and Friedman, Dan and Chen, Danqi},
  journal={Neural Information Processing Systems},
  year={2025}
}

@misc{pineai2026whispercoding,
  title        = {{Pine AI}: The Most Natural Human-Computer Interface Is Your Voice},
  author       = {{Pine AI}},
  year         = {2026},
  howpublished = {Blog post},
  url          = {https://www.19pine.ai/blog/pine-ai-the-most-natural-human-computer-interface-is-your-voice},
  note         = {Accessed 2026-06-28}
}

@misc{sturb2026sanity,
  title        = {Sanity-checking Incompressible Knowledge Probes},
  author       = {Sturb and {LawrenceC}},
  year         = {2026},
  howpublished = {LessWrong},
  url          = {https://www.lesswrong.com/posts/veFMEzDDyWaer2Sms/sanity-checking-incompressible-knowledge-probes},
  note         = {Accessed 2026-07-05}
}
